\documentclass{article}

\usepackage{microtype}
\usepackage{graphicx}
\usepackage{subfigure}
\usepackage{wrapfig}
\usepackage{booktabs} 
\usepackage{multirow}
\usepackage[listings,skins,breakable]{tcolorbox}

\usepackage[utf8]{inputenc} 
\usepackage[T1]{fontenc}    
\usepackage{inconsolata}    

\usepackage{hyperref}

\usepackage{url}            
\usepackage{booktabs}       
\usepackage{amsfonts}       
\usepackage{nicefrac}       
\usepackage{microtype}      
\usepackage{xcolor}         

\usepackage[accepted]{icml2024}

\usepackage{times}
\usepackage{latexsym}

\usepackage{amsmath}
\usepackage{amssymb}
\usepackage{mathtools}
\usepackage{amsthm}
\usepackage{algorithm}      
\usepackage{algorithmic}    
\usepackage{xcolor}
\usepackage{bbding}         
\usepackage[capitalize,noabbrev]{cleveref}

\theoremstyle{plain}

\theoremstyle{definition}

\theoremstyle{remark}

\usepackage[textsize=tiny]{todonotes}

\icmltitlerunning{FuseGen: PLM Fusion for Data-generation based Zero-shot Learning}

\begin{document}

\twocolumn[
\icmltitle{FuseGen: PLM Fusion for Data-generation based Zero-shot Learning}




\begin{icmlauthorlist}
\icmlauthor{Tianyuan Zou}{air}
\icmlauthor{Yang Liu}{air,shanghai-ai}
\icmlauthor{Peng Li}{air,shanghai-ai}
\icmlauthor{Jianqing Zhang}{air,shangjiao}
\icmlauthor{Jingjing Liu}{air}
\icmlauthor{Ya-Qin Zhang}{air}
\end{icmlauthorlist}

\icmlaffiliation{air}{Institute for AI Industry Research (AIR), Tsinghua University, Beijing, China}
\icmlaffiliation{shangjiao}{Shanghai Jiao Tong University}
\icmlaffiliation{shanghai-ai}{Shanghai Artificial Intelligence Laboratory}

\icmlcorrespondingauthor{Yang Liu}{liuy03@air.tsinghua.edu.cn}
\icmlcorrespondingauthor{Peng Li}{lipeng@air.tsinghua.edu.cn}

\icmlkeywords{Machine Learning, ICML}

\vskip 0.3in
]



\printAffiliationsAndNotice{} 

\begin{abstract}

Data generation-based zero-shot learning, although effective in training Small Task-specific Models (STMs) via synthetic datasets generated by Pre-trained Language Models (PLMs), is often limited by the low quality of such synthetic datasets. Previous solutions have primarily focused on single PLM settings, where synthetic datasets are typically restricted to specific sub-spaces and often deviate from real-world distributions, leading to severe distribution bias. To mitigate such bias, we propose FuseGen, a novel data generation-based zero-shot learning framework that 
introduces a new criteria for subset selection from synthetic datasets via utilizing multiple PLMs and trained STMs. The chosen subset provides in-context feedback to each PLM, enhancing dataset quality through iterative data generation. 
Trained STMs are then used for sample re-weighting as well, further improving data quality. Extensive experiments across diverse tasks demonstrate that FuseGen substantially outperforms existing methods, highly effective in boosting STM performance in a PLM-agnostic way. Code is provided in \url{https://github.com/LindaLydia/FuseGen}.

\end{abstract}

\section{Introduction} \label{sec:introduction}
Despite the prevalence of powerful Pre-trained Language Models (PLMs)~\cite{achiam2023gpt,team2023gemini,devlin2019bert} such as GPT-4, Small Task-specific Models (STMs) are indispensable due to their compact size and efficiency, especially for resource-constrained environments~\cite{bommasani2021opportunities}. To compensate for the scarcity of high-quality training data, synthetic data generated by PLMs has been widely applied for STM training~\cite{ye2022zerogen, wang2023openchat}.
In particular, \textit{data-generation based zero-shot learning}~\cite{ye2022zerogen,meng2022generating,gao2023self,ye2022progen} trains STM using the dataset synthesized by one PLM through task-related label-descriptive prompts, 
requiring only the task name (\textit{e.g.} movie review sentiment analysis) and label categories (\textit{e.g.} positive/negative). 
This zero-shot trained STM is significantly smaller than the original PLM with comparable performance~\cite{ye2022zerogen}, thus is particularly advantageous for domains with limited computational resources (\textit{e.g.} on mobile devices) or strict data privacy constraints (\textit{e.g.} in finance applications).

\begin{figure*}
    \centering
    \subfigure[Llama-2 ZeroGen $K=1$ (84.23)]{
         \begin{minipage}[t]{0.31\linewidth}
         \centering
         \includegraphics[width=1\linewidth]{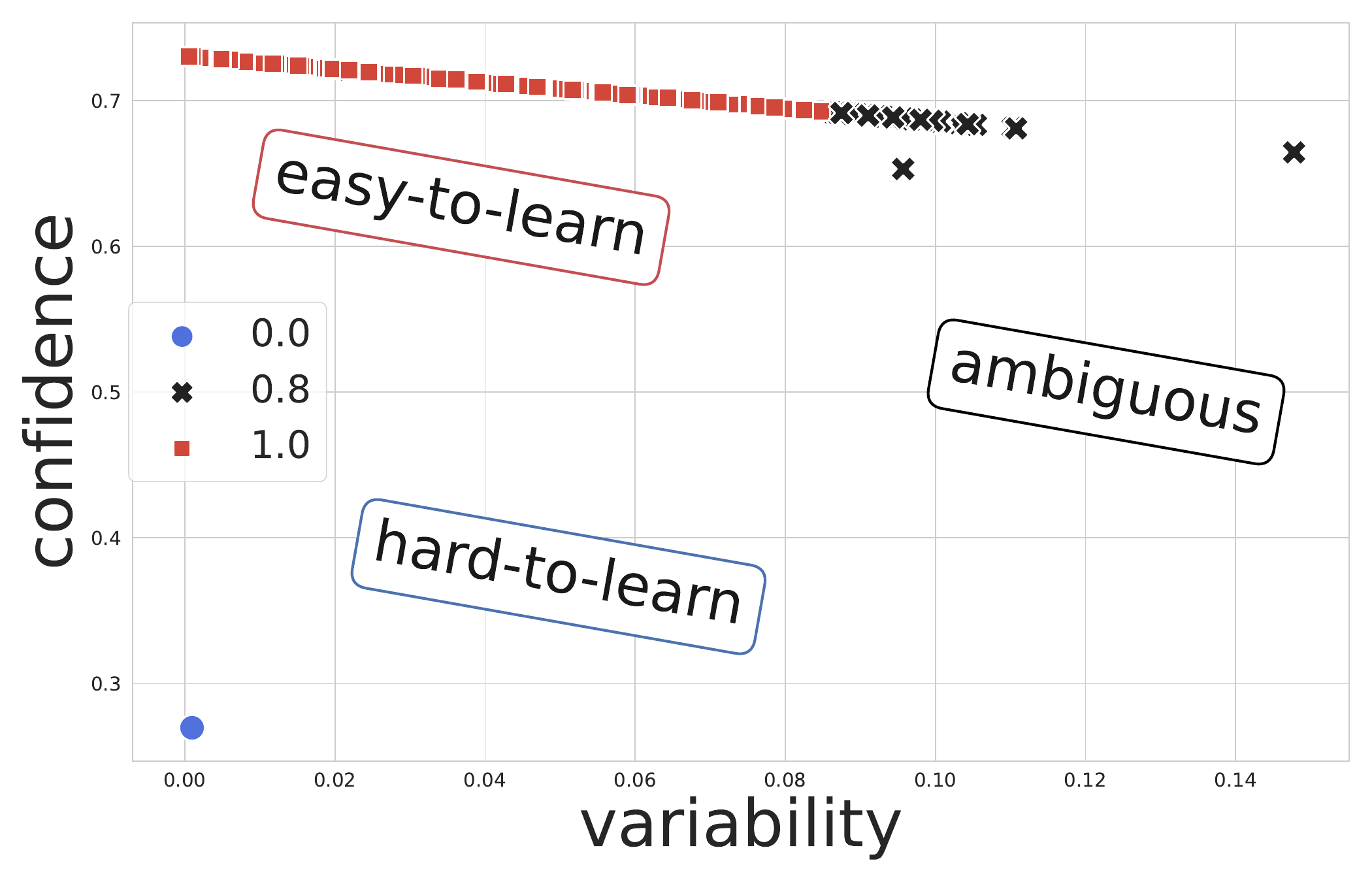}
         \label{subfig:dataset_cartography_original_llama_graphonly}
         \end{minipage}
     }
     \subfigure[Llama-2 ProGen $K=1$ (84.24)]{
         \begin{minipage}[t]{0.31\linewidth}
         \centering
         \includegraphics[width=1\linewidth]{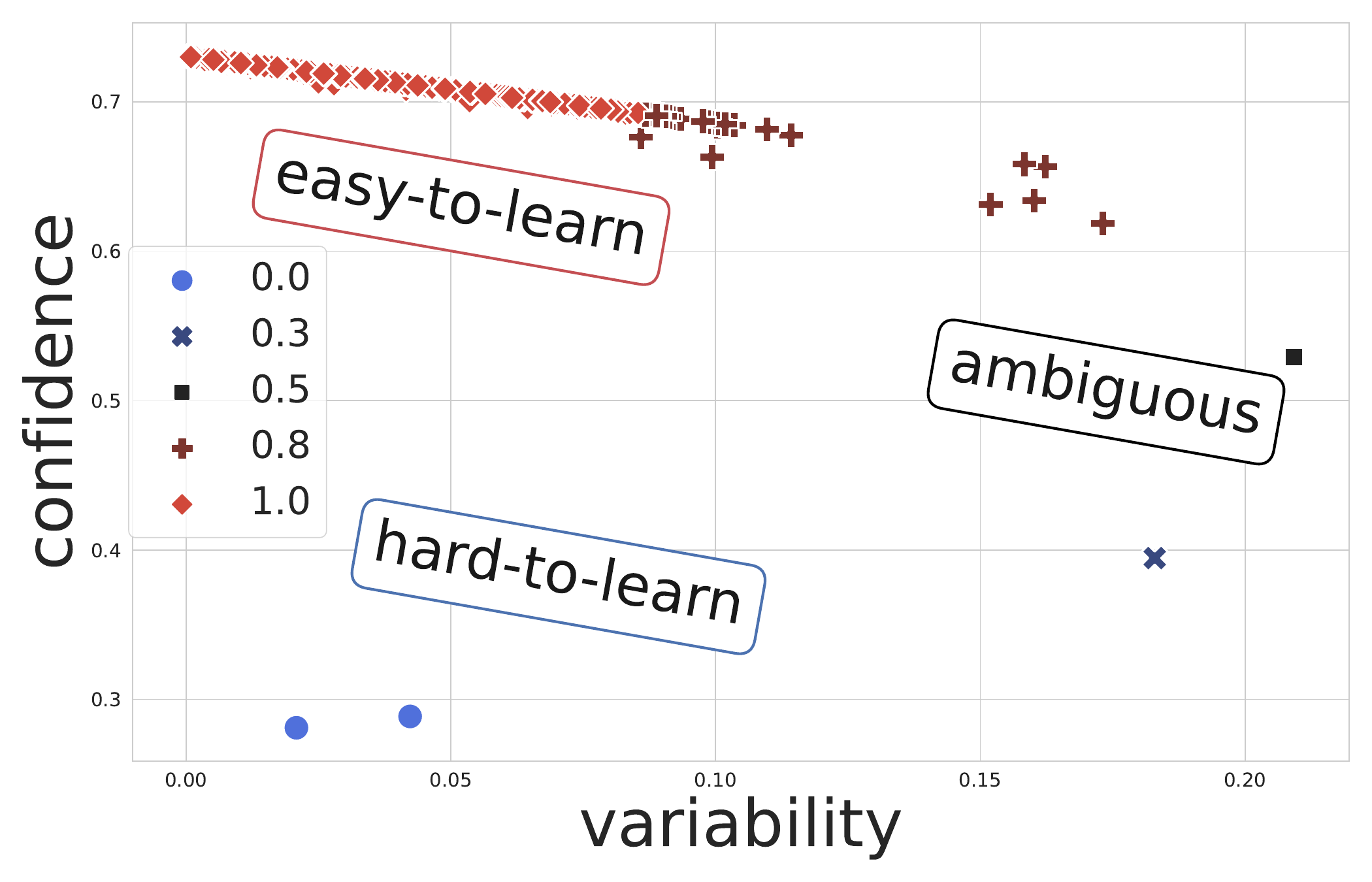}
         \label{subfig:dataset_cartography_single_progen_llama_graphonly}
         \end{minipage}
     }
     \subfigure[Llama-2 Ours $K=6$ (86.60)]{
         \begin{minipage}[t]{0.31\linewidth}
         \centering
         \includegraphics[width=1\linewidth]{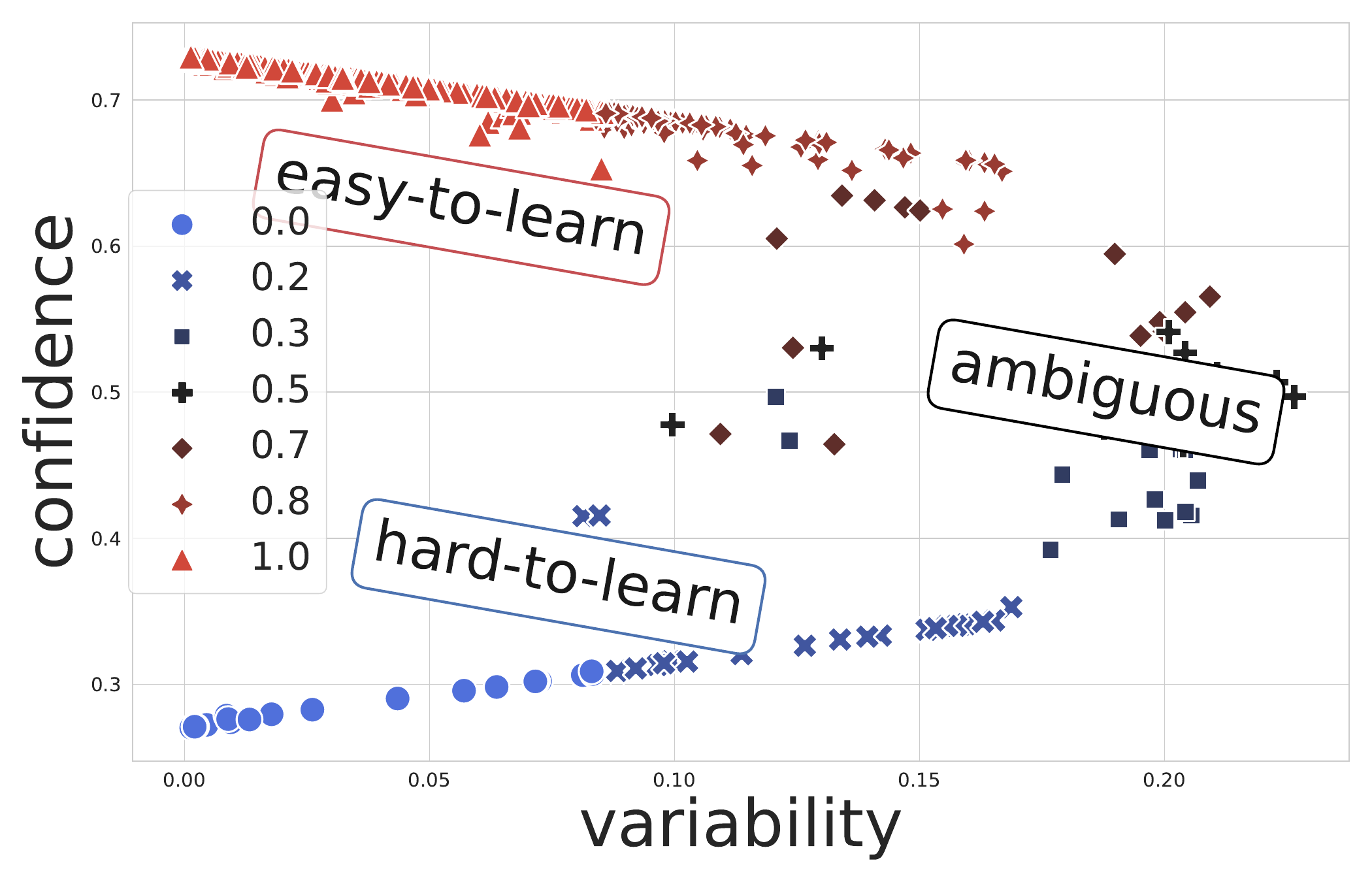}
         \label{subfig:dataset_cartography_fusegen_llama_graphonly}
         \end{minipage}
     }
    \subfigure[Flan-T5 ZeroGen $K=1$ (88.18)]{
         \begin{minipage}[t]{0.31\linewidth}
         \centering
         \includegraphics[width=1\linewidth]{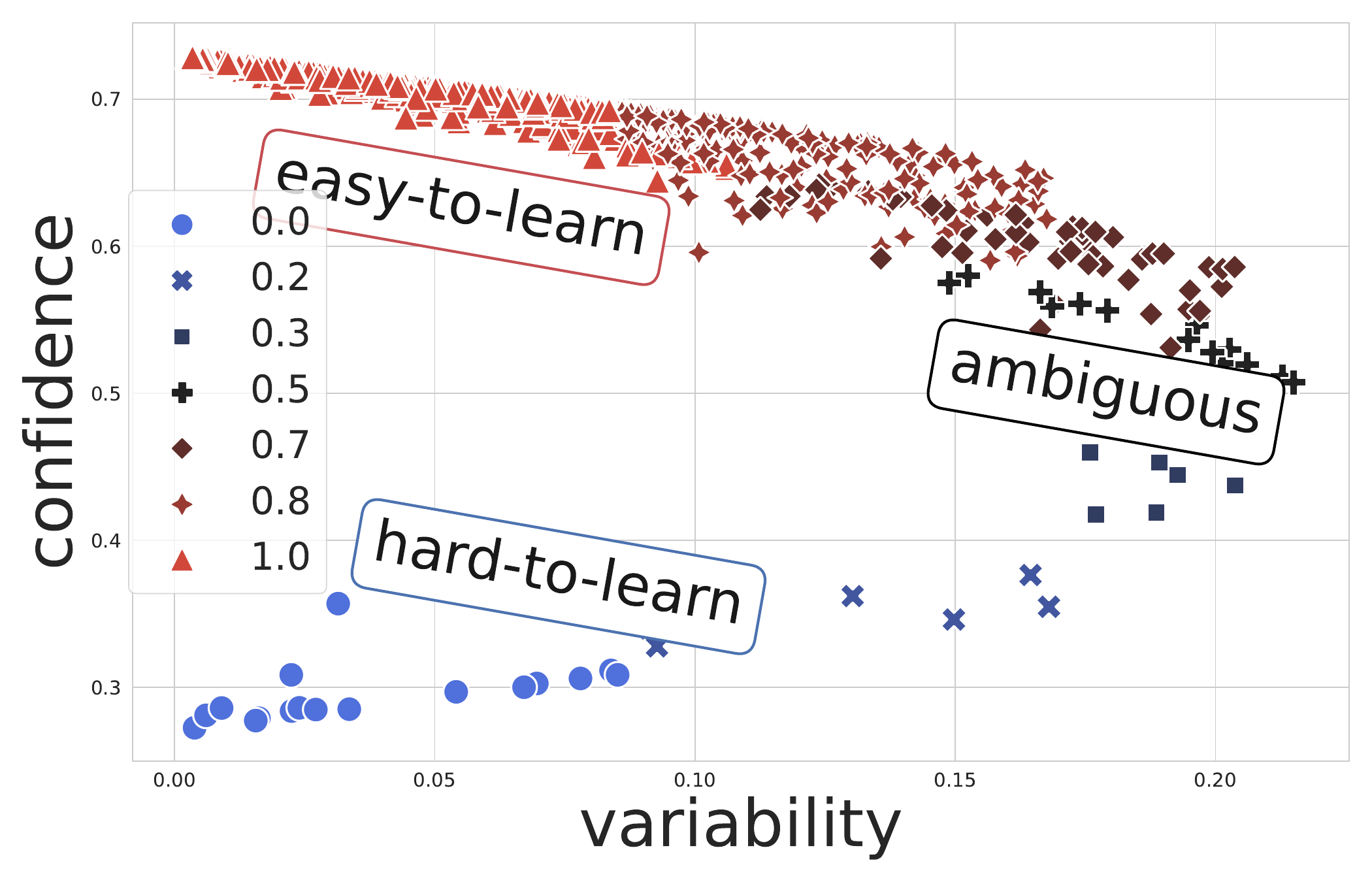}
         \label{subfig:dataset_cartography_original_flant5_graphonly}
         \end{minipage}
     }
     \subfigure[Flan-T5 ProGen $K=1$ (85.80)]{
         \begin{minipage}[t]{0.31\linewidth}
         \centering
         \includegraphics[width=1\linewidth]{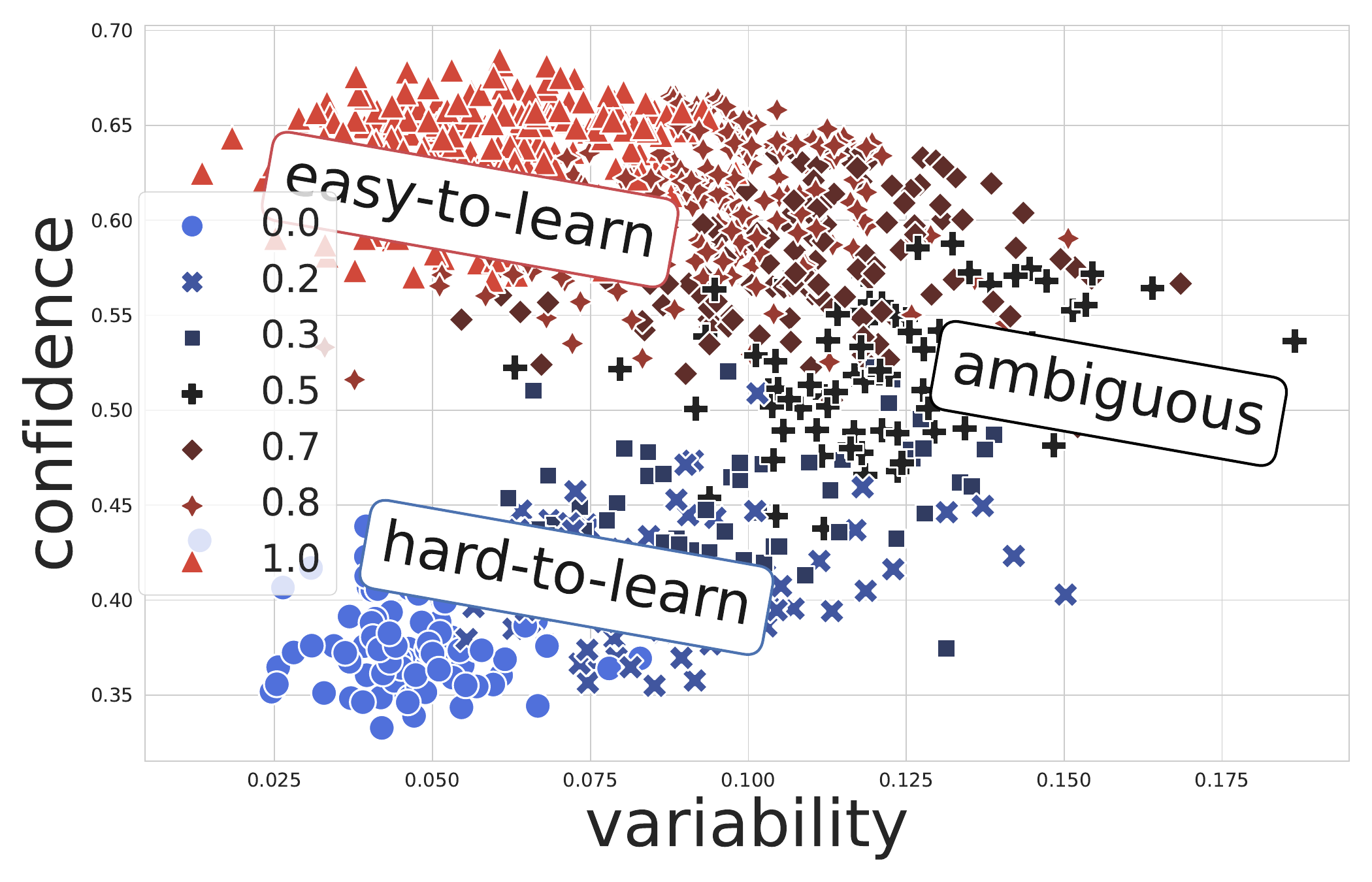}
         \label{subfig:dataset_cartography_single_progen_flant5_graphonly}
         \end{minipage}
     }
     \subfigure[Flan-T5 Ours $K=6$ (88.73)]{
         \begin{minipage}[t]{0.31\linewidth}
         \centering
         \includegraphics[width=1\linewidth]{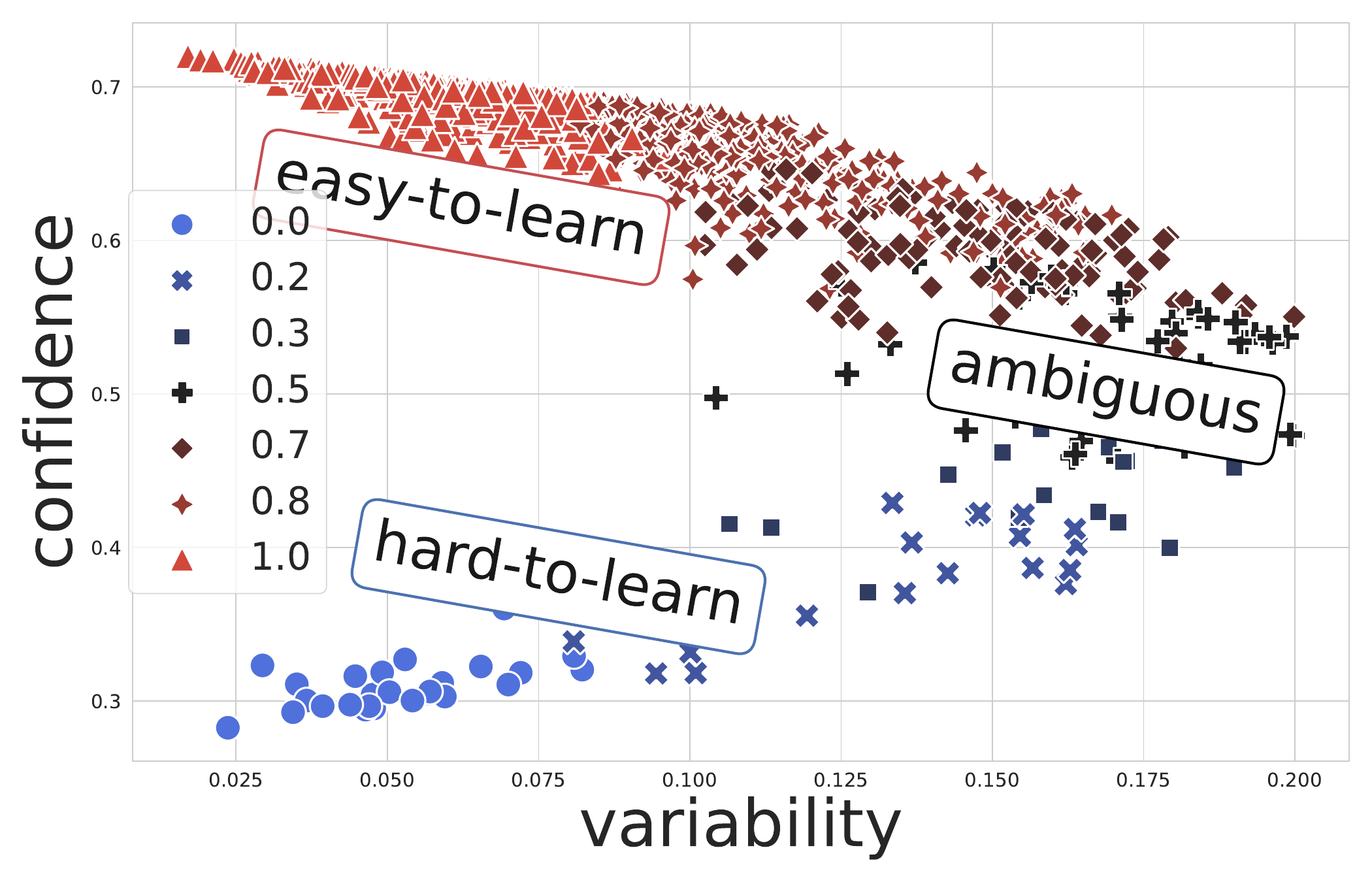}
         \label{subfig:dataset_cartography_fusegen_flant5_graphonly}
         \end{minipage}
     }
     \vspace{-0.5em}
     \caption{Synthetic dataset cartography~\cite{swayamdipta2020dataset} using $1,000$ samples generated by Llama-2 and Flan-T5 for movie review semantic analysis. ZeroGen~\cite{ye2022zerogen} uses zero-shot prompt for generation, while ProGen~\cite{ye2022progen} and FuseGen (Ours) use few-shot prompt with feedback, with ProGen relying on a single PLM and FuseGen leveraging multiple PLMs. $K$ is the number of PLMs. Numbers within parentheses are the results of STM trained with Self-boosting Weight Adjustment (see \cref{subsec:methodology_data_quality_improvement}) and evaluated over IMDb~\cite{maas2011learning_imdb} dataset. Results for more PLMs are provided in \cref{fig:appendix_dataset_cartography} in \cref{subsec:appendix_dataset_cartography}.} 
    \label{fig:dataset_cartography_llama_flant5}
\end{figure*}

However, the long-standing low-quality issue of synthetic data  impedes the practical application of STMs to a wider range~\cite{gao2023self,ye2022progen}. Previous works on improving synthetic data quality mainly focus on enhancing data diversity~\cite{fan2018hierarchical,holtzman2020the,su2022contrastive,yu2024large}, reducing redundancy~\cite{bolon2013review,deng2023mutual}, and implementing data-importance-guided in-context feedback ~\cite{ye2022progen} or sample re-weighting~\cite{gao2023self}. 
Despite notable advancements, they primarily rely on  one single PLM as source, inevitably overlooking the inherent distribution biases of synthetic datasets.

\begin{figure}[!tb]
    \centering
    \includegraphics[width=0.99\columnwidth]{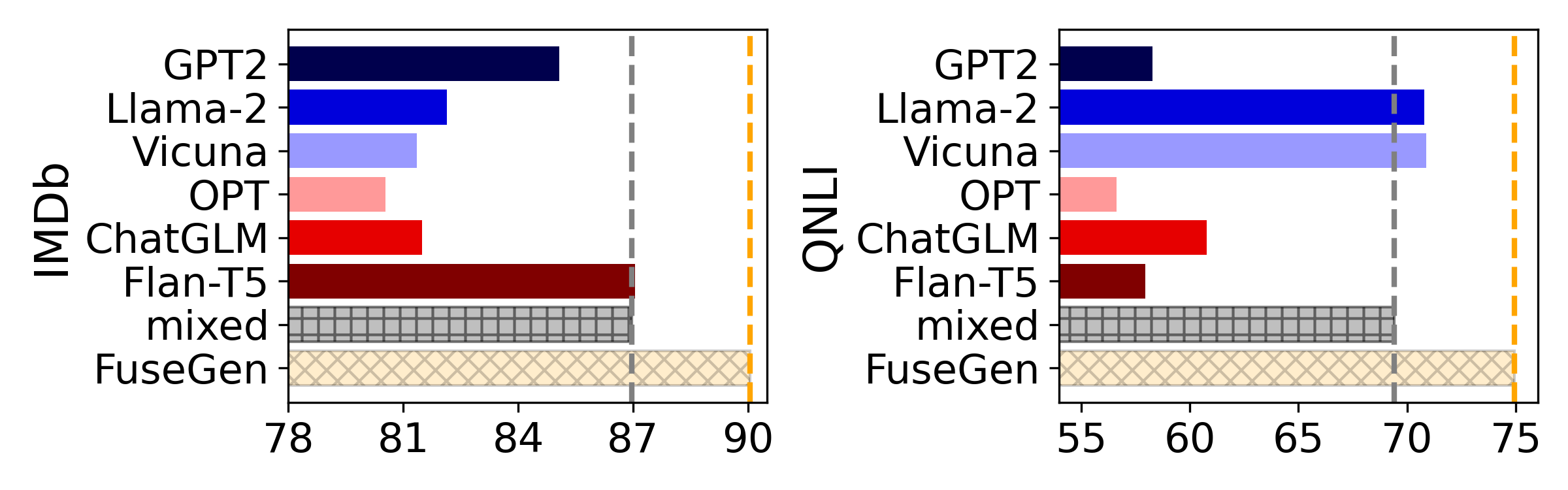}
    \caption{Performance of STM trained using $6,000$ synthetic data samples generated by various PLMs. 
    ``mixed'' uses a dataset comprising $6,000$ total samples given by the $6$ listed PLMs ($1,000$ samples per PLM). ``FuseGen'' (Ours) uses the $6$ listed PLMs and $6,000$ samples.} 
    \label{fig:mixing_ineffective}
\vspace{-1em}
\end{figure}

To thoroughly investigate these biases and their impact on STM performance, we conduct two pilot studies. 
As illustrated in \cref{fig:dataset_cartography_llama_flant5}, we use the dataset cartography approach~\cite{swayamdipta2020dataset} to plot the cartography of synthetic datasets given by different PLMs. Dataset samples are categorized into \textit{easy-to-learn} (marked in red), \textit{ambiguous} (marked in black) and \textit{hard-to-learn} (marked in blue) based on their confidence and variability, defined as the mean and standard deviation of model probabilities for their labels across training epochs. Since \textit{easy-to-learn} samples aid convergence and \textit{ambiguous} samples are vital for boosting performance~\cite{swayamdipta2020dataset}, an ideal dataset should predominantly contain diverse \textit{easy-to-learn} and \textit{ambiguous} samples, with fewer \textit{hard-to-learn} samples which are often mislabeled~\cite{swayamdipta2020dataset}. This composition of diverse samples promises better STM performance. 
In a second study, we provide the comparison between STMs trained with different datasets that vary in sources and generation methods, as illustrated in \cref{fig:mixing_ineffective}.

These visualization analyses reveal three key observations: 
$(1)$ Synthetic datasets from different PLMs exhibit significant distribution biases. For example, \cref{subfig:dataset_cartography_original_llama_graphonly,subfig:dataset_cartography_original_flant5_graphonly} show that the zero-shot synthetic dataset produced by Llama-2~\cite{touvron2023llama2} primarily includes \textit{easy-to-learn} samples, whereas that of Flan-T5~\cite{chung2022scaling} contains a more balanced mixture of all $3$ categories. 
$(2)$ 
Distribution biases are difficult to overcome by only relying on a single PLM. ProGen~\cite{ye2022progen}, an advanced single-PLM generation method, only slightly improves the ratio of \textit{easy-to-learn} and \textit{ambiguous} samples (\cref{subfig:dataset_cartography_single_progen_llama_graphonly}), while adversely increases the proportion of \textit{hard-to-learn} samples in some cases (\cref{subfig:dataset_cartography_single_progen_flant5_graphonly}). 
$(3)$ Simply mixing samples from multiple PLMs is ineffective.
As demonstrated in \cref{fig:mixing_ineffective}, plainly combining data generated by multiple PLMs improves STM performance compared to most single-PLM cases, but is still worse than the best single PLM. 

To tackle these challenges, we propose FuseGen, a smart data generation-based zero-shot learning framework that mitigates inherent dataset distribution bias by harnessing the diversity of a PLM cluster. 
In FuseGen, given a specific task and its label categories, synthetic datasets are initially generated by various PLMs in a zero-shot manner, which are then used to train their respective STMs.
To alleviate distribution bias, FuseGen selects superior samples generated by multiple PLMs as shared in-context feedback, and prompts each PLM to accumulate higher-quality data iteratively. 
To select relevant in-context samples, FuseGen pivots on an efficient cross-model criteria 
that considers both dataset composition and individual sample importance. 
To mitigate the negative impact of poor-quality samples, FuseGen further uses a self-boosting method to dynamically adjust sample weights to optimize STM in training.
As demonstrated in \cref{subfig:dataset_cartography_fusegen_llama_graphonly,subfig:dataset_cartography_fusegen_flant5_graphonly,fig:mixing_ineffective}, with these novel techniques,  FuseGen effectively reduces distribution biases and achieves better STM performance than state-of-the-art methods. 

Our contributions can be summarized as follows:

\begin{itemize}
    \item We introduce a novel data-generation based zero-shot learning framework, FuseGen, which collaboratively leverages multiple PLMs to generate higher-quality synthetic dataset without incurring any additional queries to PLMs themselves. Further, FuseGen neither requires access to nor fine-tunes the parameters of PLMs.

    \item We propose a novel cross-model criteria for selecting in-context samples, which then serves as generation feedback, and a self-boosting method for improving STM performance.

    \item Extensive evaluations on $8$ NLI and NLU tasks with $6$ open-source and $2$ closed-source PLMs demonstrate the consistent superiority of FuseGen over single-PLM methods. This PLM-agnostic nature eliminates the reliance on specific PLMs for downstream tasks.
\end{itemize}

\begin{figure*}[!tb]
    \begin{center}
    \centerline{\includegraphics[width=0.998\linewidth]{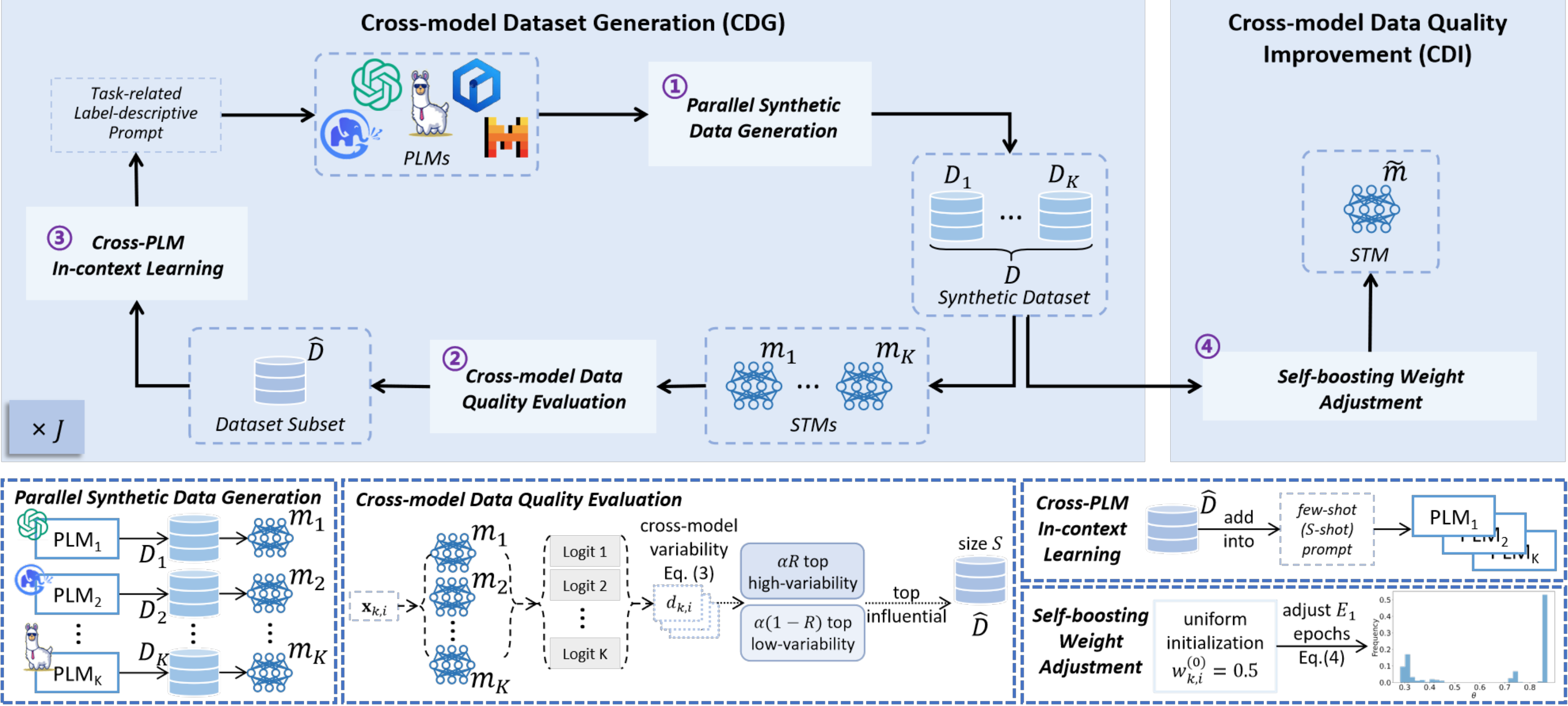}}
    \caption{Illustrated Workflow of FuseGen with two components: Cross-model Data Generation (CDG) and Cross-model Data Quality Improvement (CDI). CDG iteratively executes parallel synthetic data generation, cross-model data quality evaluation and cross-PLM in-context learning. CDI implements self-boosting weight adjustment for sample-reweighted training of STM.}
    \label{fig:workflow}
  \end{center}
\vspace{-2em}
\end{figure*}

\section{Related Work}

\textbf{Data-generation based Zero-shot Learning.} 
A recent line of research focuses on 
exploiting the data generation capabilities of PLMs~\cite{ye2022zerogen,meng2022generating,ye2022progen,gao2023self} to generate
synthetic data for training a target model~\cite{meng2022generating,ye2022zerogen,ye2022progen,gao2023self}. The dataset is generated by prompting PLM with task and label descriptions. 
A critical challenge for this approach is that generated datasets often contain low-quality samples. 
Recent attempts to address this include techniques to enhance dataset diversity (\textit{e.g.} Top-k sampling~\cite{fan2018hierarchical}, nucleus sampling~\cite{holtzman2020the}, diversely attributed prompts~\cite{yu2024large}, and contrastive search decoding~\cite{su2022contrastive}). Additionally, feature selection~\cite{bolon2013review} helps eliminate redundant information within the dataset. Finally, methods like progressive generation with in-context feedback~\cite{ye2022progen} and sample re-weighting~\cite{ye2022progen} focus on identifying and amplifying the influence of high-quality samples.
Despite significant progress, existing studies often overlook the inherent data distribution bias in synthetic datasets generated by a single PLM. In contrast, our work explores avoiding this bias by leveraging diverse multiple PLMs.

\textbf{Fusion of PLMs.}
Recent studies suggest that it is possible to combine the capabilities of multiple PLMs to obtain a model with stronger performance~\cite{wan2024knowledge,wan2024fusechat,li2024more}. 
Existing PLM knowledge-fusion techniques can be grouped into \textit{training-time fusion} and \textit{test-time fusion}~\cite{mavromatis2024pack}. 
\textit{Training-time fusion} methods~\cite{wan2024knowledge,wan2024fusechat} fuse PLMs' token-level predictions produced during training time to fine-tune a target PLM, requiring abundant computational resources.
\textit{Test-time fusion} methods do not fine-tune PLMs, but utilize methods such as logits averaging~\cite{mavromatis2024pack} and majority voting ~\cite{li2024more} to fuse the knowledge of PLMs at test time. In addition, interactions and collaborations among PLM agents~\cite{liu2023dynamic,du2023improving} have been investigated. 

All these works demonstrate that collaboration among diverse PLMs helps. However, all existing works require direct access to training samples, which means they are not applicable to the setting of data generation-based zero-shot learning, the problem we aim to solve.  

\section{FuseGen} 
\label{sec:preliminaries_and_motivation} 
\subsection{Preliminaries}
In \textit{data-generation based zero-shot learning}~\cite{ye2022zerogen,gao2023self} with \textit{a single PLM},   
given a downstream task like text classification,
a PLM $\mathcal{P}$ with parameter $\Phi_{\mathcal{P}}$ first generates a synthetic dataset $\mathcal{D}=\{(\mathbf{x}_{i},y_{i})\}_{i=1}^{N}$ of size $N$. This is accomplished by using a proper task-related label-descriptive prompt $\mathcal{T}(\cdot)$ (examples are provided in \cref{subsec:appendix_prompt}) as follows:
\begin{equation} \label{eq:generation}
  \mathbf{x}_{i} \sim \mathcal{P}(\cdot|\mathcal{T}(y_{i}),\Phi_{\mathcal{P}}) \, .
\end{equation}
$\mathcal{D}$ is then used to train an STM $m$ with the following training objective:
\begin{equation} \label{eq:stm_train_loss}
    \mathcal{L} = \sum_{i=1}^{N} \ell(m(\mathbf{x}_{i}),y_{i}),
\end{equation}
where $\ell$ is a common loss function, \textit{e.g.} cross-entropy loss.

\subsection{FuseGen Architecture Overview} 
\label{sec:methodology} 

Different from previous works, we focus on \textit{multi-PLM setting} and propose FuseGen. The FuseGen workflow is illustrated in  \cref{fig:workflow}. 
In a nutshell, FuseGen consists of two main components: Cross-model Dataset Generation (CDG) (\cref{subsec:methodology_data_generation}) and Cross-model Data Quality Improvement (CDI) (\cref{subsec:methodology_data_quality_improvement}). For CDG, given a fixed number of samples to generate in total, PLMs progressively generate datasets for multiple rounds, each round using an improved subset of samples generated from previous rounds as in-context examples. This is realized in three steps: $(1)$ \textit{Parallel Synthetic Data Generation}: each PLM generates its own dataset and trains a respective STM. $(2)$ \textit{Cross-model Data Quality Evaluation}: the quality of generated samples is evaluated using a cross-PLM criteria to select a desirable subset. $(3)$ \textit{Cross-PLM In-context Learning}: the cross-PLM subsets are used as in-context examples to prompt PLMs to generate new datasets. Step $(1)$ is then repeated. After the required number of samples is reached, we perform CDI 
which re-weights samples with a self-boosting strategy.
 \cref{alg:algorithm_full_functions} provides an overview of the above steps, with each function detailed in \cref{sec:appendix_algorithm}.

\begin{algorithm}[tb]
\small
\caption{FuseGen} 
\label{alg:algorithm_full_functions}
\textbf{Input:}\\ \quad
$K$ PLMs, empty synthetic dataset $\{\mathcal{D}_k\leftarrow\emptyset\}_{k=1}^K$, target number of synthetic samples $N$ for each PLM, sample selection hyper-parameter $\alpha,R,S$,
number of feedback steps $J$ taken to obtain in total $N$ synthetic samples, 
random initialized STM $m_{(0)}$, 
test dataset of downstream task $\mathcal{A}$, 
initialized sample weights ${\left\{\{w_{k,i}^{(0)}\}_{i=1}^{N}\right\}}_{k=1}^K$, 
learning rate $\eta$, 
number of weight adjustment epochs $E_1$, 
number of STM training epochs $E_2$.\\
\textbf{Output:} STM $\tilde{m}$ that obtains the effectively aggregated knowledge from $K$ PLMs.

\begin{algorithmic}[1]
    \STATE Initialize in-context feedback samples $\hat{\mathcal{D}} \leftarrow \emptyset$.
    \FOR{$j=0$ {\bfseries to} $J$}
        \FOR{$k=1$ {\bfseries to} $K$ {\bfseries in parallel}}
            \STATE $\mathcal{D}_k \leftarrow$ \verb|S_AccumulativeSynDataGeneration(|$\mathcal{D}_k$, $\hat{\mathcal{D}}$, $N$, $J$, $j$\verb|)|.
            \STATE $m_k \leftarrow$ \verb|S_STMTraining(|$\mathcal{D}_k$, $m_{(0)}$, $E_2$\verb|)|.
        \ENDFOR
        \STATE $\tilde{m} \leftarrow$ \verb|S_STMTraining(|$\cup_{k=1}^{K}\mathcal{D}_k$, $m_{(0)}$, $E_2$\verb|)|.
        \STATE $\hat{\mathcal{D}} \leftarrow$ \verb|C_SampleSelection(|$\cup_{k=1}^{K}\mathcal{D}_k$, $\{m_k\}_{k=1}^K$, $\tilde{m}$, $\alpha$, $R$, $S$\verb|)|.
    \ENDFOR
    \STATE $\tilde{m} \leftarrow$ \verb|S_WeightAdjustSTMTraining(|$\cup_{k=1}^{K}\mathcal{D}_k$, $m_{(0)}$, $\cup_{k=1}^{K}\left\{\{w_{k,i}^{(0)}\}_{i=1}^{N}\right\}$, $E_1$, $E_2$\verb|)|.
\end{algorithmic}
\end{algorithm}

\subsection{Cross-model Dataset Generation} \label{subsec:methodology_data_generation}
In FuseGen, each PLM iteratively generates a total of $N$ samples across $J+1$ rounds, incorporating feedback from STMs after each of the first $J$ rounds. In each round, a total of $\frac{N}{J+1}$ samples are generated using the accumulated knowledge of multiple PLMs from previous rounds as feedback. Specifically, the following steps are taken: 

\textbf{Parallel Synthetic Dataset Generation.}
In each round, each of $K$ PLMs (denoted as $\{\mathcal{P}_k\}_{k=1}^K$)  generates a synthetic dataset $\mathcal{D}_k=\{(\mathbf{x}_{k,i},y_{k,i})\}_{i=1}^{\frac{N}{J+1}}$ of size $\frac{N}{J+1}$ in parallel with the same task-related label-descriptive prompt $\mathcal{T}(\cdot)$ as described in \cref{sec:preliminaries_and_motivation}. 
Each dataset is then used to train a separate STM $m_k$ following~\cref{eq:stm_train_loss}. This step produces $K$ separate STMs and $K$ synthetic datasets.

\textbf{Cross-model Data Quality Evaluation.} \label{subsec:method_sample_selection}
In this step, we aim to select a desirable subset from
$\mathcal{D}=\bigcup_{k=1}^K \mathcal{D}_k$ to guide data generation.
To accomplish this goal, we utilize the knowledge of trained STMs at hand and develop a simple yet efficient criteria for data-quality evaluation.

As discussed in \cref{sec:introduction} 
, \textit{easy-to-learn} samples of low-variability and \textit{ambiguous} samples of high-variability are both vital for constructing a desirable dataset, 
valuable for training convergence and model generalization ability, respectively.
Inspired by this
, we first use cross-model variability $d_{k,i}$ to categorize each sample,
defined as: 
\begin{equation}
\small
    d_{k,i}=\mathrm{STD}(p_{1,k,i}[y_{k,i}],...,p_{k',k,i}[y_{k,i}],...,p_{K,k,i}[y_{k,i}])    
\end{equation}
where $p_{k',k,i}[y_{k,i}]$ denotes STM model $m_{k'}$'s predicted probability of synthetic label $y_{k,i}$ on that sample $\mathbf{x}_{k,i}$, and $\mathrm{STD}$ represents standard deviation\footnote{Different from \citet{swayamdipta2020dataset}, we do not include confidence (\textit{i.e.} mean of predicted probability in our criteria, as the synthetic label is not used for in-context samples (see \cref{subsec:appendix_prompt} for in-context sample examples).}. 
To prompt the generation of a dataset that includes both low-variability (low $d_{k,i}$) and high-variability (high $d_{k,i}$) data, we  
select a small number of candidates (of size $R \ll N$) comprised of $\alpha R$ top high-variability and $(1-\alpha)R$ top low-variability samples, where $\alpha$ is a hyper-parameter that controls the percentage of high-variability samples.
The goal here is to efficiently select a smaller and more manageable subset from a large set of candidates.
The selected subset can then be processed by more computationally intensive ranking. To further identify samples that are vital for training, we train an STM $\tilde{m}$ using $\mathcal{D}$ 
and leverage the noise-resistant influence function proposed in ProGen~\cite{ye2022progen} to select the top-$S$ influential samples from the $R$ candidate samples ($S < R$). 
Our results validate that these selected samples 
originate from various PLMs (See \cref{subsec:appendix_sample_origination}.)

\textbf{Cross-PLM In-context Learning.} \label{subsec:method_cross_plm_progen}
After selecting $S$ in-context samples (denoted as $\hat{\mathcal{D}}$), we add them to the original prompt $\mathcal{T}(\cdot)$, resulting in $\mathcal{T}(\hat{\textbf{x}}_{1}, \dots,\hat{\textbf{x}}_{S};\cdot)$ (see examples in \cref{subsec:appendix_prompt}). We then send the feedback prompt to each PLM to generate $\frac{N}{J+1}$ new samples following $\mathbf{x}_{k,i} \sim \mathcal{P}_k(\cdot|\mathcal{T}(\hat{\textbf{x}}_{1}, \dots,\hat{\textbf{x}}_{S};y_{k,i}),\Phi_{\mathcal{P}_k})$, where $\Phi_{\mathcal{P}_k}$ denotes the parameter of $\mathcal{P}_k$. In this way, PLMs can learn from each other and generate datasets with improved quality.

\subsection{Cross-model Data Quality Improvement} \label{subsec:methodology_data_quality_improvement}
After CDG process that improves overall data distribution, we perform one last step of re-weighting samples by their quality, determined by a \textbf{Self-boosting Weight Adjustment (SWA)} approach.

As \textit{hard-to-learn} samples (refer to \cref{subfig:dataset_cartography_fusegen_llama_graphonly,subfig:dataset_cartography_fusegen_flant5_graphonly}) and low-quality samples (\textit{e.g.} meaningless or irrelevant) still exist post-CDG, we down-weight these samples in each training round of the final STM
$\tilde{m}$. Specifically, a weight $w_{k,i}$ (uniformly initialized as $0.5$) is assigned to each sample in $\mathcal{D}=\{{\{(\mathbf{x}_{k,i},y_{k,i})\}_{i=1}^{N}\}}_{k=1}^K$. 
At the $e_1$-th weight-adjustment round of $\tilde{m}$, we update $w_{k,i}$ using the following boosting strategy inspired by TrAdaBoost~\cite{dai2007boosting}:
\label{subsec:methodology_weight_decay}
\begin{equation} \label{eq:weight_adjust_onlywrong_onlyself_loss}
\begin{split}
     w_{k,i}^{(e_1+1)} &= w_{k,i}^{(e_1)} \beta^{-\text{error}_{k,i}(1-{\text{correct}_{k,i}})}, \\ \, k&=1,\dots,K,\,\, i=1,\dots,N \, ,
\end{split}
\end{equation}
where $\beta=\frac{1}{1+\sqrt{\frac{2 \ln{(NK)}}{E_1}}}>0$ is a constant value for weight adjustment, $E_1$ is the number of total epochs for weight adjustment, $\text{error}_{k,i}=1-p_{k,i}[y_{k,i}]$ is the prediction error of $\tilde{m}$ on data sample $\mathbf{x}_{k,i}$, and $\text{correct}_{k,i}=1$ if $\tilde{m}$ predicts sample $\mathbf{x}_{k,i}$ correctly, otherwise $\text{correct}_{k,i}=0$. Normalization is applied afterwards to guarantee that 
$\sum_{k=1}^K\sum_{i=1}^{N} w_{k,i}^{(e_1)} = 0.5 NK$. After normalization, $w_{k,i}$ for correctly inferred samples increases while that for wrongly inferred samples decreases. 
A new STM is trained from scratch with the new weights after each adjustment step. Training details are provided in \cref{alg:algorithm_full_functions,alg:algorithm_functions}. 
With SWA, the training objective for $\tilde{m}$ using all synthetic data $\mathcal{D}$ is given by:
\begin{equation} \label{eq:weighted_loss}
    \mathcal{L}=\sum_{k=1}^{K}\sum_{i=1}^N w_{k,i}\cdot\ell(\tilde{m}(\mathbf{x}_{k,i}),y_{k,i}) \, . 
\end{equation}

Unlike SunGen~\cite{gao2023self}, which utilizes a self-guided sample re-weighting method with bi-level SGD optimization to enhance its STM performance, 
our SWA achieves comparable STM performance without requiring this computationally expensive optimization step (see \cref{sec:experiments,subsec:appendix_ablation_more_datasets}). This translates to a significantly smaller computational cost.

\section{Experiments}\label{sec:experiments}

\begin{table*}[!tb]
    \centering
    \small
    \resizebox{0.98\linewidth}{!}{
    \begin{tabular}{l||cccccc||cccccc}
    \toprule
        ~ & \multicolumn{6}{c||}{IMDb} & \multicolumn{6}{c}{SST-2} \\
        \cmidrule(){2-13}
        ~ & $\tilde{m}_{G}$ & $\tilde{m}_{L}$ & $\tilde{m}_{V}$ & $\tilde{m}_{O}$ & $\tilde{m}_{C}$ & $\tilde{m}_{F}$ & $\tilde{m}_{G}$ & $\tilde{m}_{L}$ & $\tilde{m}_{V}$ & $\tilde{m}_{O}$ & $\tilde{m}_{C}$ & $\tilde{m}_{F}$ \\
    \midrule
        ZeroGen $^\spadesuit$ & 85.07 & 82.14 & 81.36 & 80.54 & 81.49 & 87.06 & 80.99 & 79.47 & 82.33 & 82.00 & 86.49 & 81.88 \\
        SunGen $^\spadesuit$ & 86.94 & 86.59 & 84.93 & 85.21 & 84.76 & \underline{89.79} & 83.45 & 84.30 & 84.04 & 83.49 & \underline{87.18} & 83.53 \\
        ProGen $^\spadesuit$ & 85.68 & 84.33 & 82.14 & 85.57 & 87.41 & 88.00 & 83.60 & 79.53 & 82.53 & 82.78 & 86.64 & 83.17 \\
        \hline
        \\[-1em]
        FuseGen (Ours) & \multicolumn{6}{c||}{\textbf{90.06}} & \multicolumn{6}{c}{\textbf{87.51}} \\
    \midrule
    \midrule
        ~ & \multicolumn{6}{c||}{Yelp} & \multicolumn{6}{c}{QNLI} \\
        \cmidrule(){2-13}
        ~ & $\tilde{m}_{G}$ & $\tilde{m}_{L}$ & $\tilde{m}_{V}$ & $\tilde{m}_{O}$ & $\tilde{m}_{C}$ & $\tilde{m}_{F}$ & $\tilde{m}_{G}$ & $\tilde{m}_{L}$ & $\tilde{m}_{V}$ & $\tilde{m}_{O}$ & $\tilde{m}_{C}$ & $\tilde{m}_{F}$ \\
    \midrule
        ZeroGen $^\spadesuit$ & 89.73 & 89.74 & 85.67 & 87.13 & 82.00 & 92.41 & 58.30 & 70.79 & 70.88 & 56.64 & 60.77 & 57.95 \\
        SunGen $^\spadesuit$ & 91.85 & 89.30 & 89.06 & 91.22 & 88.86 & \underline{93.13} & 62.26 & 74.20 & \underline{74.35} & 57.50 & 65.64 & 58.21 \\
        ProGen $^\spadesuit$ & 91.26 & 89.82 & 88.55 & 89.00 & 88.81 & 91.71 & 58.38 & 69.56 & 70.29 & 57.46 & 61.08 & 69.44 \\
        \hline
        \\[-1em]
        FuseGen (Ours) & \multicolumn{6}{c||}{\textbf{93.47}} & \multicolumn{6}{c}{\textbf{74.92}} \\
    \midrule
    \midrule
        ~ & \multicolumn{6}{c||}{MNLI-matched} & \multicolumn{6}{c}{MNLI-mismatched} \\
        \cmidrule(){2-13}
        ~ & $\tilde{m}_{G}$ & $\tilde{m}_{L}$ & $\tilde{m}_{V}$ & $\tilde{m}_{O}$ & $\tilde{m}_{C}$ & $\tilde{m}_{F}$ & $\tilde{m}_{G}$ & $\tilde{m}_{L}$ & $\tilde{m}_{V}$ & $\tilde{m}_{O}$ & $\tilde{m}_{C}$ & $\tilde{m}_{F}$ \\
    \midrule
        ZeroGen $^\spadesuit$ & 41.99 & 48.52 & 45.87 & 36.16 & 32.65 & 47.37 & 46.38 & 50.04 & 48.10 & 36.74 & 33.00 & 49.95 \\
        SunGen $^\spadesuit$ & 44.66 & \underline{49.43} & 46.27 & 37.44 & 32.71 & 49.04 & 47.45 & \underline{51.67} & 48.63 & 38.35 & 33.02 & 51.66 \\
        ProGen $^\spadesuit$ &  43.35 & 48.69 & 47.50 & 36.79 & 32.81 & 48.56 & 46.57 & 50.57 & 49.65 & 40.27 & 33.01 & 50.24 \\
        \hline
        \\[-1em]
        FuseGen (Ours) & \multicolumn{6}{c||}{\textbf{49.76}} & \multicolumn{6}{c}{\textbf{51.70}} \\
    \midrule
    \midrule
        ~ & \multicolumn{6}{c||}{AgNews} & \multicolumn{6}{c}{MarkedNews} \\
        \cmidrule(){2-13}
        ~ & $\tilde{m}_{G}$ & $\tilde{m}_{L}$ & $\tilde{m}_{V}$ & $\tilde{m}_{O}$ & $\tilde{m}_{C}$ & $\tilde{m}_{F}$ & $\tilde{m}_{G}$ & $\tilde{m}_{L}$ & $\tilde{m}_{V}$ & $\tilde{m}_{O}$ & $\tilde{m}_{C}$ & $\tilde{m}_{F}$ \\
    \midrule
        ZeroGen $^\spadesuit$ & 77.86 & 83.40 & 81.25 & 84.81 & 83.17 & 81.87 & 77.16 & 74.49 & 74.10 & 77.80 & 80.33 & 76.12 \\
        SunGen $^\spadesuit$ & 80.94 & 84.44 & 82.50 & \underline{85.68} & 84.12 & 85.57 & 78.01 & 76.75 & 76.39 & 78.15 & 82.16 & 77.85 \\
        ProGen $^\spadesuit$ & 78.68 & 83.93 & 81.46 & 85.66 & 84.74 & 84.59 & 77.17 & 76.51 & 76.14 & 77.93 & \underline{82.70} & 78.75 \\
        \hline
        \\[-1em]
        FuseGen (Ours) & \multicolumn{6}{c||}{\textbf{86.89}} & \multicolumn{6}{c}{\textbf{83.85}} \\
    \bottomrule
    \end{tabular}
    }
\caption{Comparison of FuseGen and baselines with $K=6$. Methods marked by $^\spadesuit$ are single-PLM methods. $\tilde{m}_{G}$, $\tilde{m}_{L}$, $\tilde{m}_{V}$, $\tilde{m}_{O}$, $\tilde{m}_{C}$, $\tilde{m}_{F}$ represents the final STM performance with single PLM GPT-2, Llama-2, Vicuna, OPT, ChatGLM3 and Flan-T5, respectively. 
Best result is marked as \textbf{bold}, and the second best is marked with \underline{underline}.} 
\label{tab:6k_comparison}
\end{table*}

\subsection{Experimental Settings}

\textbf{Models.} In our experiments, we evaluate on $6$ open-source PLMs: GPT-2-xl (GPT-2)~\cite{radford2019language}, Llama-2-7b-chat-hf (Llama-2)~\cite{touvron2023llama2}, Vicuna-7b-1.5v (Vicuna)~\cite{vicuna2023}, OPT-6.7b (OPT)~\cite{zhang2022opt}, ChatGLM3-6b-base (ChatGLM3)~\cite{du2022glm} and Flan-T5-xl (Flan-T5)~\cite{chung2022scaling}. 2 closed-source PLMs are also used for generating synthetic datasets: GPT-3.5-turbo-instruct (GPT-3.5)~\cite{openai2021gpt3-5} and GPT-4-turbo-preview (GPT-4)~\cite{openai2023gpt4}. For the choice of STM, we use
bert-base-uncased (BERT), a pre-trained model, to perform downstream classification tasks. 
The trained STM is evaluated over a real-world human-annotated dataset (test dataset) 
$\mathcal{A}$ that is never used during training.

\textbf{Datasets.}
We select 7 well-developed datasets to evaluate our framework: $1)$ IMDb~\cite{maas2011learning_imdb} and SST-2~\cite{socher2013recursive_sst2,wang2019glue} for movie review semantic analysis task, $2)$ Yelp-polarity~\cite{zhang2015character_yelp} for restaurant review semantic analysis task, $3)$ AgNews~\cite{zhang2015character_agnews} for news category classification task, $4)$ QNLI~\cite{wang2019glue} for question-information entailment classification task, $5)$ MNLI (both matched and mismatched)~\cite{adina2018abroad_mnli} for sentence-pair relation classification task.
To test the effectiveness of FuseGen on unseen tasks, we further create a new dataset named MarkedNews from AgNews.
MarkedNews categorizes articles containing the symbol ``\$'' as ``Money with \$ included'', and all other articles retain their original AgNews categories. 
This creates  a new $5$-class classification task: ``World'', ``Sports'', ``Business'', ``Technology'', and ``Money with \$ included''.
We adopt the original test dataset as $\mathcal{A}$ except for QNLI and MNLI, where ground-truth labels are unavailable. In these cases, we use the validation sets instead. 
The experiments
run on A100-80G.

\textbf{Baselines.} 
We compare our framework with several existing data-generation based zero-shot learning methods, including $1)$ ZeroGen~\cite{ye2022zerogen} which directly trains an STM using the generated synthetic data, $2)$ SunGen~\cite{gao2023self} which recovers a robust synthetic dataset through sample-level weight optimization, and $3)$ ProGen~\cite{ye2022progen} which progressively generates data using self-given in-context feedback through prompt. To ensure a fair comparison, all methods generate the same number of samples. In other words, each single-PLM method produces a total of $N \times K$ samples. 

\textbf{Implementation Details.} 
Unless otherwise stated, the following setting is applied: $N=1,000$ synthetic data samples generated by each PLM are used for FuseGen; the BERT models (STMs) are trained with Adam optimizer with a learning rate of $2\times10^{-5}$ and training epochs ($E_2$) of $3$.
When training STMs, weight adjustment is performed for $30$ iterations ($E_1=30$). 
Each experiment is repeated $3$ times using different random seeds, and averaged accuracy
is reported. 
$\alpha=0.5,R=40,S=8$ is used to select in-context samples for constructing feedback prompt, except for QNLI and MNLI 
datasets, where $R=20,S=4$ is used in order to fit the maximum input length of each PLM. $J=4$ is used for iterative generation (both FuseGen and ProGen). For SunGen, $50$ samples are used for sample-weight backward gradient estimation. 

\subsection{Main Results} \label{subsec:main_results}

\cref{tab:6k_comparison} summarizes the main results of our FuseGen framework and compared baseline methods. 
To ensure comprehensive evaluation, each single-PLM baseline method is evaluated using samples generated from each of the PLMs.

\textbf{Open-source PLMs.}
\cref{tab:6k_comparison} shows that FuseGen consistently outperforms all baselines using the same number of generated samples.
(\textit{i.e.} each PLM generates $6,000$ samples for training $\tilde{m}_{k}$ for baselines), demonstrating the superior data quality of FuseGen. 
Our method achieves up to $1.2\%$ increase in STM performance over the best-performing single-PLM baseline, which exploits the optimal PLM for each task. SunGen performs consistently well among single-PLM baselines, but the ideal PLM varies by task. However, in  zero-shot setting, where no task-specific samples are available, pre-selecting a PLM for optimal training performance is impractical. FuseGen is free from such pre-selection.

\textbf{Unseen Tasks.}
Evaluation results for FuseGen and baselines over our new dataset MarkedNews are shown in \cref{tab:6k_comparison}, with synthetic data generation prompts detailed in \cref{subsec:appendix_prompt}.
FuseGen outperforms all baselines consistently, demonstrating its ability to enhance downstream STM performance 
even when PLMs lack prior knowledge of the unseen classification task. 
\begin{table}[!tb]
\small
\centering
    \resizebox{0.79\linewidth}{!}{
    \begin{tabular}{l||cc}
    \toprule
        ~ & $\tilde{m}_{GPT-3.5}$ & $\tilde{m}_{GPT-4}$ \\
    \midrule
        ZeroGen $^\spadesuit$ & 51.66 & 49.48 \\
        SunGen $^\spadesuit$ & 52.92 & \underline{55.82} \\
        ProGen $^\spadesuit$ & 52.50 & 55.76 \\
        \hline
        \\[-1em]
        FuseGen (Ours) & \multicolumn{2}{c}{\textbf{56.56}} \\
    \bottomrule
    \end{tabular}
    }
\caption{Comparison of FuseGen and baseline methods on closed-source PLMs with QNLI dataset and $K=2$. 
}
\label{tab:2bert_qnli_close_sourced}
\end{table}
\textbf{Closed-source PLMs.} We also conduct experiments on the fusion of two popular closed-source models (GPT-3.5 and GPT-4) using QNLI dataset with $K=2$. Results in \cref{tab:2bert_qnli_close_sourced} (each $\tilde{m}_{k}$ is trained with $2,000$ samples) demonstrate the superior performance of FuseGen compared to baselines. 

FuseGen's consistent superiority across diverse tasks and models underscores its PLM-agnostic nature. This eliminates the need
of relying on specific models for downstream tasks, making it a more flexible and efficient solution.

\subsection{Ablation Study}

\subsubsection{Multi-PLM v.s. Single-PLM} \label{subsubsec:mult_and_single}


\begin{figure}
    \centering
    \includegraphics[width=0.99\columnwidth]{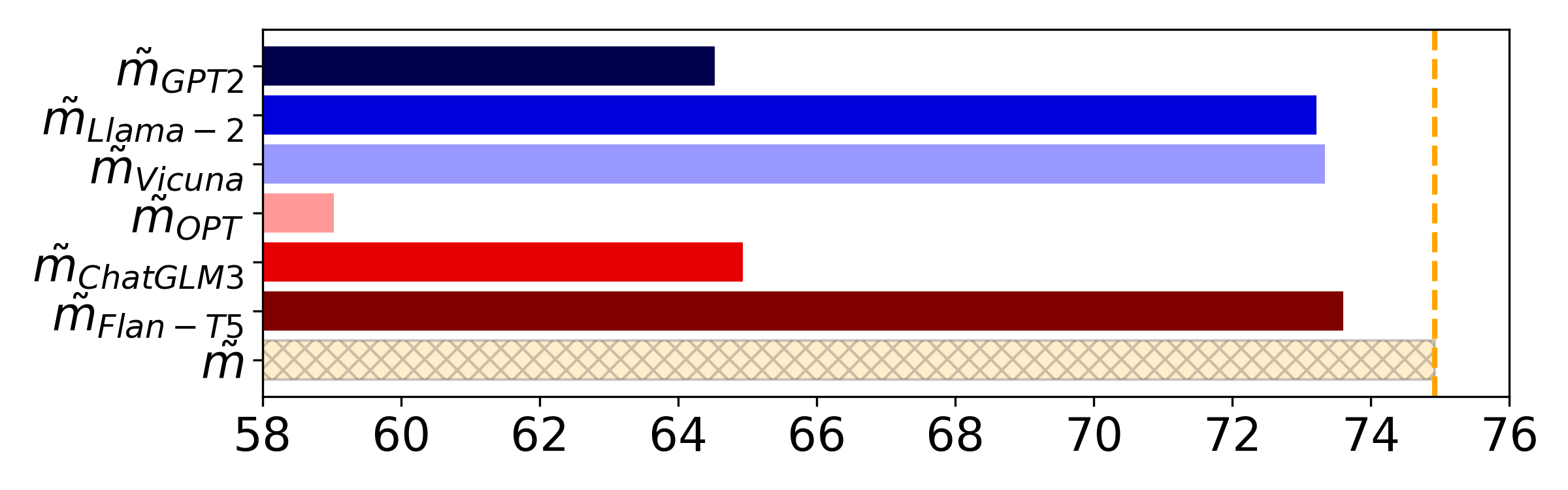}
    \vspace{-1em}
    \caption{Comparison of FuseGen between using multi-PLM (last bar) and single-PLM with QNLI dataset.}
    \label{fig:singlePLM_and_multiPLM}
\end{figure}

We evaluate the impact of multi-PLM fusion by comparing FuseGen between using multi-PLM ($K=6$) and single-PLM ($K=1$). Results are provided in \cref{fig:singlePLM_and_multiPLM}. Since cross-model variability evaluation in CDG can not be performed for $K=1$, random selection is applied here to select $R$ candidate samples, whereas CDI is applied to both cases.
\cref{fig:singlePLM_and_multiPLM} shows that \textit{multi-PLM collaboration is vital for further improving the quality of synthetic dataset, yielding better STM performance than 
relying on single-PLM}.
Detailed results on more datasets are provided in \cref{tab:singlePLM_and_multiPLM} in \cref{subsec:appendix_k_is_1_more_datasets}.

\subsubsection{In-context Sample Selection} \label{subsubsec:sample_selection_results}

\begin{table}[!tb]
    \centering
    \resizebox{1\linewidth}{!}{
    \begin{tabular}{cc|c||cccccc|c}
    \toprule
        \multicolumn{2}{c|}{Variability} & \multirow{2}{*}{\shortstack{Influ-\\ence}} & \multirow{2}{*}{$m_{G}$} & \multirow{2}{*}{$m_{L}$} & \multirow{2}{*}{$m_{V}$} & \multirow{2}{*}{$m_{O}$} & \multirow{2}{*}{$m_{C}$} & \multirow{2}{*}{$m_{F}$} & \multirow{2}{*}{$\tilde{m}$} \\
        \cline{1-2}
        Low & High & ~ & ~ & ~ & ~ & ~ & ~ & ~ & ~ \\
    \midrule
        \multicolumn{2}{c|}{Rand.} & \XSolidBrush & 52.47 & 67.48 & 65.90 & 50.52 & \underline{56.68} & 67.66 & 72.89 \\
        \Checkmark & \XSolidBrush & \XSolidBrush & 53.77 & 66.18 & 61.33 & 50.96 & 53.37 & 66.13 & 73.76 \\
        \XSolidBrush & \Checkmark & \XSolidBrush & 54.98 & 65.48 & 60.76 & 49.79 & 54.28 & 65.47 & 73.81 \\
        \Checkmark & \Checkmark & \XSolidBrush & \underline{58.59} & \underline{70.85} & 66.31 & 50.38 & 55.23 & 67.83 & 74.14 \\
        \multicolumn{2}{c|}{Rand.} & \Checkmark & 54.25 & 70.44 & \underline{70.74} & \underline{51.19} & \underline{56.68} & 68.84 & 74.07 \\
        \Checkmark & \XSolidBrush & \Checkmark & 54.00 & 70.07 & 67.75 & 51.12 & 55.70 & 66.49 & 74.08 \\
        \XSolidBrush & \Checkmark & \Checkmark & 54.85 & 66.47 & 64.46 & 50.08 & 56.50 & \underline{70.50} & \underline{74.16} \\
        \hline
        \\[-1em]
        \Checkmark & \Checkmark & \Checkmark & \multirow{2}{*}{\textbf{59.68}} & \multirow{2}{*}{\textbf{71.48}} & \multirow{2}{*}{\textbf{72.37}} & \multirow{2}{*}{\textbf{52.37}} & \multirow{2}{*}{\textbf{57.33}} & \multirow{2}{*}{\textbf{72.12}} & \multirow{2}{*}{\textbf{74.92}} \\
        \multicolumn{3}{c||}{FuseGen (Ours)} & ~ & ~ & ~ & ~ & ~ & ~ & ~ \\
    \bottomrule
    \end{tabular}
    }
    \caption{Comparison of different in-context sample selection methods with QNLI as test dataset. ``Variability'' is cross-model variability, and ``Rand.'' stands for random sampling for in-context sample candidate selection. $m_{G}$, $m_{L}$, $m_{V}$, $m_{O}$, $m_{C}$, $m_{F}$ each represents $m_{GPT-2}$, $m_{Llama-2}$, $m_{Vicuna}$, $m_{OPT}$, $m_{ChatGLM3}$, $m_{Flan-T5}$ and $\tilde{m}$ is the final STM trained using $\mathcal{D}$. Best result is marked as \textbf{bold} and the second best marked with \underline{underline} for each STM (each column).} 
    \label{tab:incontext_sample_selection_ablation}
\end{table}

In-context sample selection is a critical component of the FuseGen framework, as it influences the quality of feedback from STMs to PLMs, which in turn affects the generation quality of PLMs. In this section, we compare various in-context sample selection strategies, including random selection, high-variability and low-variability selection. The latter two exclusively select top-$R$ high-variability or low-variability samples, respectively. We also evaluate each strategy with and without fine-grained influence-based selection. The results are shown in \cref{tab:incontext_sample_selection_ablation}. 
We also report the performance of each $m_{k}$ trained with SWA using the corresponding $\mathcal{D}_{k}$ during the FuseGen process in \cref{tab:incontext_sample_selection_ablation}.
Our in-context sample selection strategy surpasses other alternatives consistently, not just in the final STM performance, but also for each intermediate small model $m_{k}$ produced during FuseGen. This 
underscores \textit{the efficacy of our selection approach} and \textit{FuseGen's ability to produce higher-quality datasets for all PLMs involved}.


\subsubsection{Effectiveness of SWA and CDG} \label{subsubsec:ablation_swa}

\begin{table}[!tb]
    \centering
    \resizebox{1\linewidth}{!}{
    \begin{tabular}{l||cccccc|c}
    \toprule
        ~ & $m_{G}$ & $m_{L}$ & $m_{V}$ & $m_{O}$ & $m_{C}$ & $m_{F}$ & $\tilde{m}$ \\
    \midrule
        FuseGen (Ours) & 59.68 & 71.48 & 72.37 & 52.37 & 57.33 & 72.12 & 74.92 \\
        \hline
        \\[-1em]
        w/o SWA & 56.72 & 69.99 & 70.94 & 51.98 & 56.39 & 68.65 & 73.41 \\
        \hline
        \\[-1em]
        w/o CDG \& SWA & 51.24 & 65.81 & 70.61 & 50.83 & 53.01 & 55.73 & 69.41 \\
        \hline
        \\[-1em]
        SDG+mixed & 52.13 & 69.22 & 70.11 & 51.79 & 54.87 & 68.58 & 70.20 \\
    \bottomrule
    \end{tabular}
    }
\caption{Comparison between FuseGen and its ablations using $N=1,000$ with QNLI as test dataset. $m_{G}$, $m_{L}$, $m_{V}$, $m_{O}$, $m_{C}$, $m_{F}$ each represents $m_{GPT-2}$, $m_{Llama-2}$, $m_{Vicuna}$, $m_{OPT}$, $m_{ChatGLM3}$, $m_{Flan-T5}$, while $\tilde{m}$ is the final STM trained using the  dataset $\mathcal{D}$.} 
\label{tab:ablation_results}
\end{table}

As FuseGen consists of $2$ components, CDG and CDI (mainly achieved by SWA), we perform ablation study by removing SWA and CDG step by step from FuseGen, resulting in $2$ ablations: ``w/o SWA'' and ``w/o CDG \& SWA''. Note when both CDI and CDG are removed, datasets are generated from multiple PLMs using zero-shot prompt and naively combined (the "mixed" case in \cref{fig:mixing_ineffective}).
We further add ablation ``SDG+mixed'' (also without SWA) which naively combines datasets given by multiple PLMs using self-guided data generation (SDG) for in-context feedback (same as $K=1$ in \cref{subsubsec:mult_and_single}).
Results are summarized in \cref{tab:ablation_results} and \cref{tab:ablation_results_appendix} in \cref{subsec:appendix_ablation_more_datasets}. 
From \cref{tab:ablation_results}, we observe a $1.51\%$ drop in $\tilde{m}$ performance when removing SWA, and another $5.51\%$ drop 
when further removing CDG, 
demonstrating that \textit{SWA is effective in boosting knowledge transfer from synthetic dataset to STM} and \textit{CDG is effective in fusing the knowledge of multiple PLMs}. Also, CDG (``w/o SWA'') outperforms ``SDG+mixed'' by a huge margin ($3.21\%$), verifying the superiority of collaborative feedback over self-guided feedback.

As SunGen~\cite{gao2023self} also re-weights samples to boost STM performance, we further compare the performance of SWA with SunGen (using 50 samples for estimating gradients of sample weights), with results shown in \cref{tab:time_cost_sungen_swd}.
We observe that, SunGen's computational cost is two orders-of-magnitude higher than SWA when re-weighting $1,000$ to $6,000$ samples, yet delivers comparable performance.
This underscores the effectiveness and efficiency of SWA, making our framework much more computationally effective.

\begin{table}[!tb]
\centering
    \resizebox{1.\linewidth}{!}{
    \begin{tabular}{c|c||c|cccccc}
    \toprule
        \multicolumn{2}{c||}{~} &  time [s] & $\tilde{m}_{G}$ & $\tilde{m}_{L}$ & $\tilde{m}_{V}$ & $\tilde{m}_{O}$ & $\tilde{m}_{C}$ & $\tilde{m}_{F}$ \\
    \midrule
        \multirow{2}{*}{$1,000$} & SunGen & \hphantom{0}43.3 & \textbf{57.46} & \textbf{72.01} & 72.14 & 50.71 & \textbf{55.45} & 57.31 \\ 
        ~ & SWA & \hphantom{0}\hphantom{0}0.1 & 56.95 & 71.13 & \textbf{72.21} & \textbf{51.96} & 55.12 & \textbf{57.43} \\ 
        \hline
        \\[-1em]
        \multirow{2}{*}{$6,000$} & SunGen & 240.8 & 62.26 & 74.20 & \textbf{74.35} & 57.50 & \textbf{65.64} & 58.21 \\ 
        ~ & SWA & \hphantom{0}\hphantom{0}0.5 & \textbf{62.59} & \textbf{74.58} & \textbf{74.35} & \textbf{58.42} & 64.81 & \textbf{58.47} \\ 
    \bottomrule
    \end{tabular}
    }
\caption{Comparison on running time for each weight adjustment epoch and STM performance between SunGen and SWA with QNLI as test dataset. Best result is marked as \textbf{bold}.}
\label{tab:time_cost_sungen_swd}
\end{table}

\begin{figure}[!tb]
    \centering
    \subfigure[Effect of $\alpha$]{
         \begin{minipage}[t]{0.3\linewidth}
         \centering
         \includegraphics[width=1\linewidth]{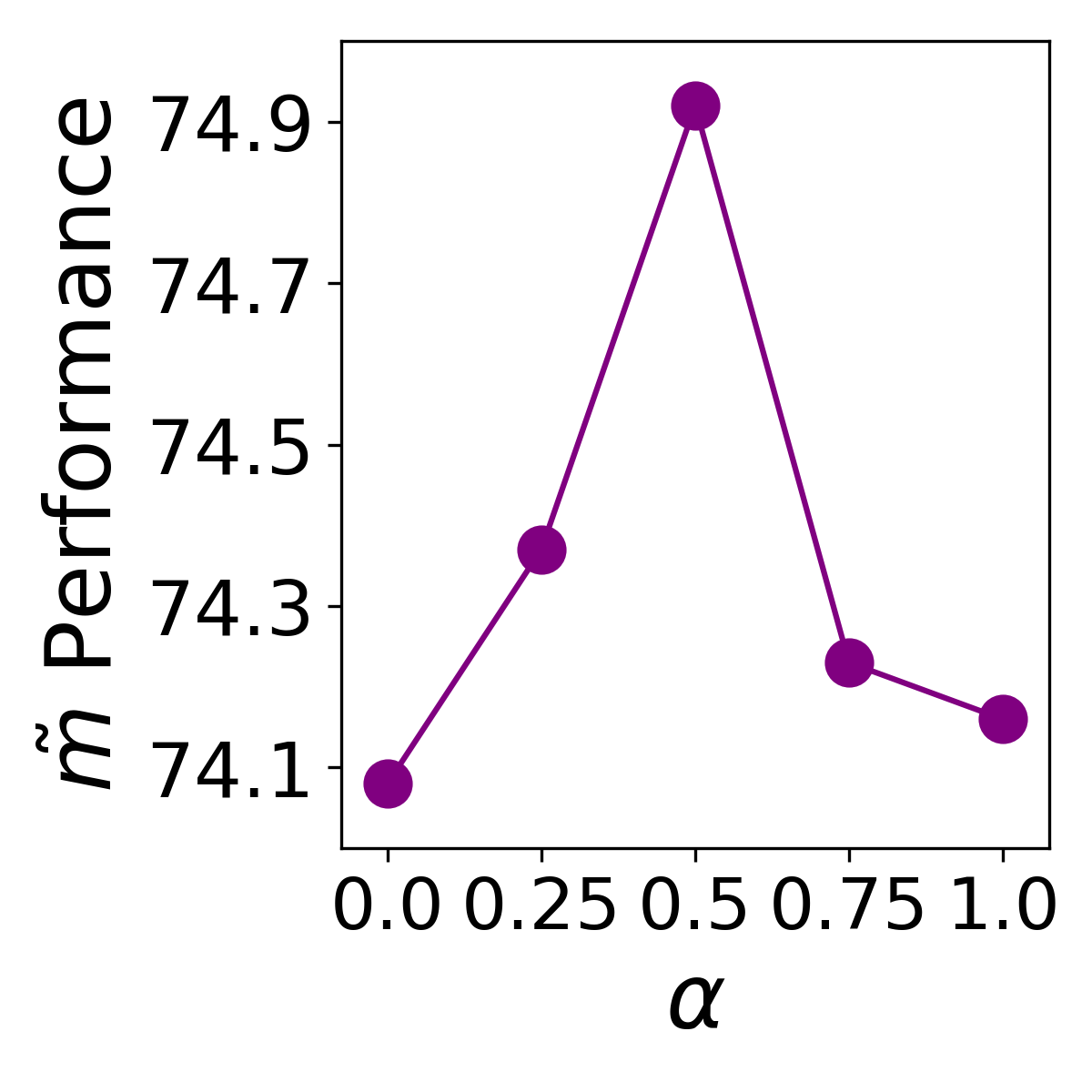}
         \vspace{-1em}
         \label{fig:varying_alpha_onlyFuse}
         \end{minipage}
     }
     \vspace{-0.3em}
     \subfigure[Effect of $N$]{
         \begin{minipage}[t]{0.3\linewidth}
         \centering
         \includegraphics[width=1\linewidth]{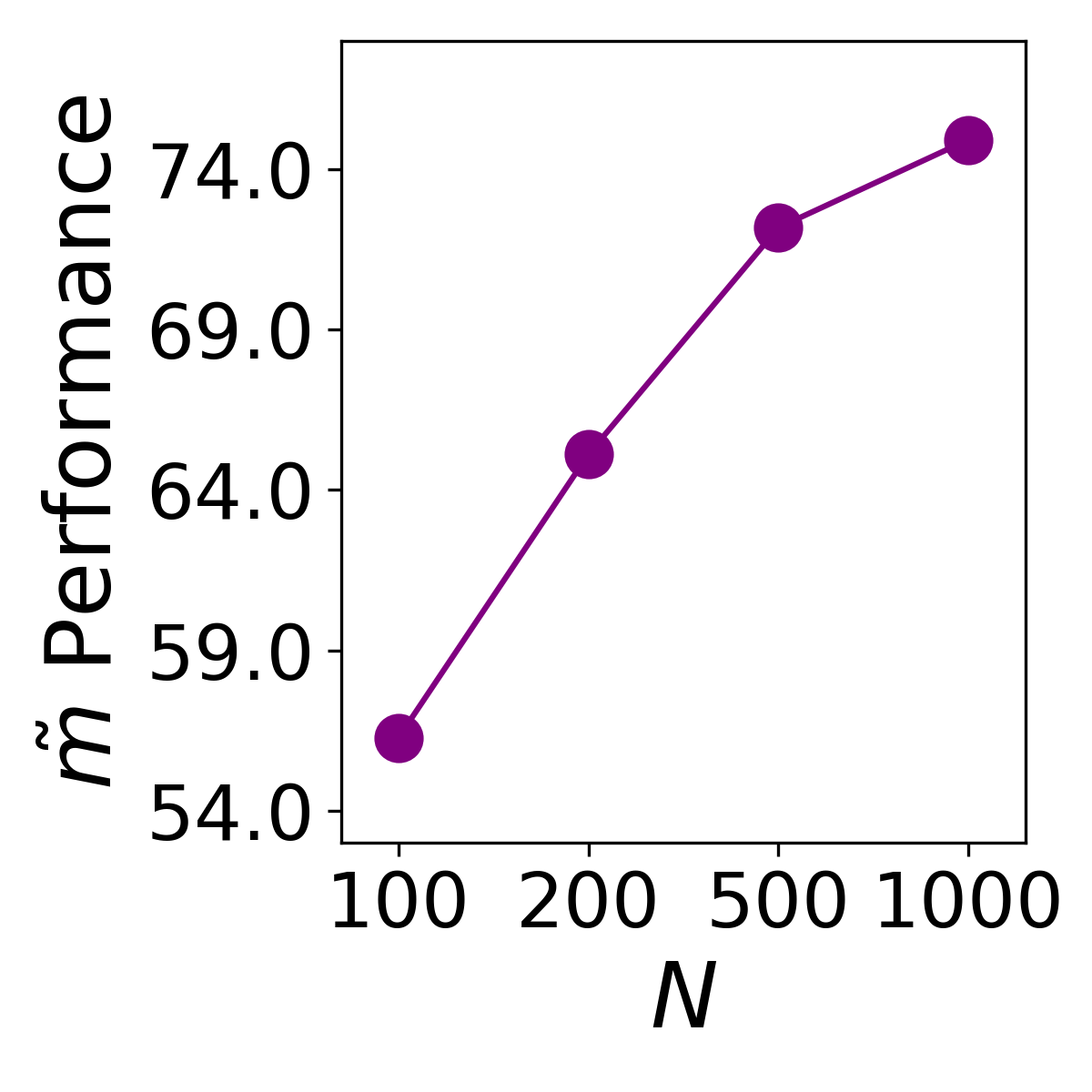}
         \vspace{-1em}
         \label{fig:varying_N_onlyFuse}
         \end{minipage}
     }
     \vspace{-0.3em}
     \subfigure[Effect of $J$]{
         \begin{minipage}[t]{0.3\linewidth}
         \centering
         \includegraphics[width=1\linewidth]{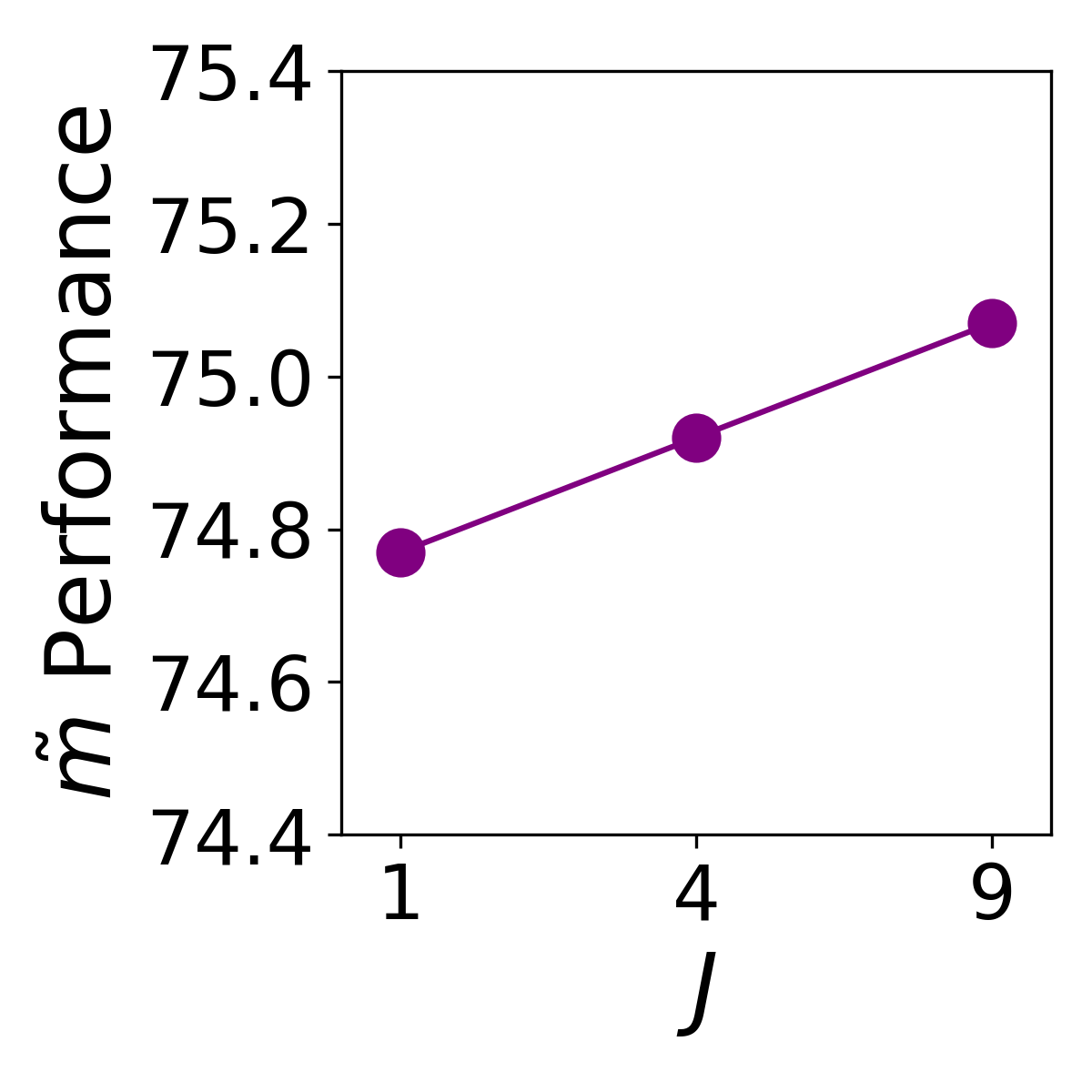}
         \vspace{-1em}
         \label{fig:varying_J_onlyFuse}
         \end{minipage}
     }
     \vspace{-0.3em}
     \caption{Ablation results on different hyper-parameters used for FuseGen with QNLI as test dataset.}
    \label{fig:results_ablations_onlyFuse}
\vspace{-1em}    
\end{figure}

\subsubsection{Effect of Hyper-parameters} 

We further study the impact of hyper-parameters $\alpha$ (ratio of high-variability samples within the $R$ in-context sample candidates), $N$ (sample generation budget), and $J$ (feedback times) of FuseGen with $K=6$ in \cref{fig:results_ablations_onlyFuse}. Detailed results with each $m_{k}$ are included in
\cref{tab:varying_alpha,tab:varying_N,tab:varying_J} 
in \cref{subsec:appendix_ablation_results}.



\textbf{Effect of $\alpha$.} 
\cref{fig:varying_alpha_onlyFuse} shows that, too many or too few high-variability samples in the candidate set both hurt the synthetic dataset quality, resulting in lower STM performance, whereas a balanced mix ($\alpha =0.5$) yields the highest STM results. 




\textbf{Effect of $N$.} 
\cref{fig:varying_N_onlyFuse} demonstrates that STM performance  improves with the increase of $N$. Additionally, the performance improvement rate decelerates at larger values of $N$. 



\textbf{Effect of $J$.} 
From \cref{fig:varying_J_onlyFuse}, we observe that increasing $J$ results in a slight but consistent improvement in performance, likely due to the fact that more precise guidance is given to PLMs by a more frequent feedback during the process.

\section{Conclusion}

We propose a novel data-generation based zero-shot learning framework FuseGen that harnesses the collaborative capability of multiple PLMs to improve synthetic data generation of PLMs. 
We first integrate multiple PLMs to alleviate distribution bias of synthetic datasets through cross-PLM in-context samples selection, for constructing better feedback recursively. To further improve the quality of the generated synthetic dataset and boost STM performance, we employ a self-boosting weight adjustment strategy to down-weight low-quality samples. Extensive experiments and ablation studies on various NLI and NLU tasks demonstrate that FuseGen is highly  
effective, query-efficient and PLM-agnostic without the reliance on specific PLMs for downstream tasks, making it a more flexible and resource-efficient solution.  

\section{Limitations}
This work sheds lights on the possibility of multi-PLM collaboration in the field of zero-shot learning. 
However, it does not delve deeply into the interrelationships between pairs of PLMs. A more thorough investigation could yield insightful conclusions regarding which PLMs are most complementary to one another. Meanwhile, aside from seeding the same feedback to all PLMs, more personalized feedback can be constructed to better suit the inherit distribution bias of each PLM, which may further boost STM performances. 
\nocite{langley00}

\bibliography{example_paper}
\bibliographystyle{icml2024}

\clearpage
\newpage
\appendix
\label{appendix}

\section{Prompts Used in Experiments}
\subsection{Task-related Label-descriptive Prompts} \label{subsec:appendix_prompt}
We present the prompts used for synthetic dataset generation in \cref{tab:appendix_prompt}. 
For information-question entailment analysis task (QNLI) and sentence pair relation analysis task (MNLI), we leverage the open-source Wikipedia-short (\url{https://github.com/yumeng5/SuperGen/tree/main/pretrain_corpus}) dataset, which contains short Wikipedia sequences ($5$ to $30$ words) extracted from sentences in Wikipedia. We use these sentences as the information source for the prompts. In other words, each occurrence of \emph{<information>} or \emph{<sentence1>} within the prompt is replaced with a randomly-chosen Wikipedia-short sequence before feeding it to PLMs. 

\begin{table*}[!htb]
\centering
    \resizebox{0.998\linewidth}{!}{
    \begin{tabular}{c|c|l|c}
    \toprule
        Dataset (task) & type & prompt & label \\
    \midrule
        \multirow{2}[2]{*}{\shortstack{IMDb and SST2\\(semantic analysis\\of movie review)}} & zero-shot & {\shortstack[l]{``The movie review in \textit{\textbf{positive/negative}} sentiment for a movie is: ''}} & \textit{\textbf{positive/negative}} \\
        \cline{2-4}
        \\[-1em]
        ~ & few-shot & \shortstack[l]{``The movie review is: \textit{<sample\_1>}\\The movie review is: \textit{<sample\_2>}\\...\\The movie review is: \textit{<sample\_S>}\\ \\The movie review in \textit{\textbf{positive/negative}} sentiment which is diverse \\in the expression compared to the above given samples is: ''} & \textit{\textbf{positive/negative}} \\
        \hline
        \\[-1em]
        \multirow{2}[2]{*}{\shortstack{Yelp\\(semantic analysis\\of restaurant review)}} & zero-shot & ``The restaurant review in \textit{\textbf{positive/negative}} sentiment is:'' & \textit{\textbf{positive/negative}} \\
        \cline{2-4}
        \\[-1em]
        ~ & few-shot & \shortstack[l]{``The restaurant review is: \textit{<sample\_1>}\\The restaurant review is: \textit{<sample\_2>}\\...\\The restaurant review is: \textit{<sample\_S>}\\ \\The new restaurant review in \textit{\textbf{positive/negative}} sentiment which is diverse in\\the expression compared to the above given samples is: ''} & \textit{\textbf{positive/negative}} \\
        \hline
        \\[-1em]
        \multirow{2}[2]{*}{\shortstack{QNLI\\(information-question\\entailment analysis)}} & zero-shot & \shortstack[l]{``Information: \textit{<information>} \\Question (answer \textit{\textbf{in/not in}} above information): ''} & \textit{\textbf{entailment/not\_entailment}} \\
        \cline{2-4}
        \\[-1em]
        ~ & few-shot & \shortstack[l]{``The Information-Question pair is: \textit{<sample\_1>}\\The Information-Question pair is: \textit{<sample\_2>}\\...\\The Information-Question pair is: \textit{<sample\_S>}\\ \\The new Information-Question pair which is diverse in the expression \\compared to the above given samples is: Information: \textit{<information>} \\Question (answer \textit{\textbf{in/not in}} above information): ''} & \textit{\textbf{entailment/not\_entailment}} \\
        \hline
        \\[-1em]
        \multirow{2}[2]{*}{\shortstack{MNLI (matched\\and mismatched)\\(sentence pair\\relation analysis)}} & zero-shot & \shortstack[l]{``<sentence1> \textit{\textbf{In other words,}} /\\<sentence1> \textit{\textbf{Furthermore,}} /\\\textit{\textbf{There is a rumor that}} <sentence1> \textit{\textbf{However, the truth is:}} ''} & \shortstack{\textit{\textbf{entailment/}}\\\textit{\textbf{neutral}}/\\\textit{\textbf{contradiction}}} \\
        \cline{2-4}
        \\[-1em]
        ~ & few-shot & \shortstack[l]{``The sentence pair is: \textit{<sample\_1>}\\The sentence pair is: \textit{<sample\_2>}\\...\\The sentence pair is: \textit{<sample\_S>}\\ \\The new sentence pair which is diverse in the expression \\compared to the above given samples is: <sentence1> \textit{\textbf{In other words,}} /\\<sentence1> \textit{\textbf{Furthermore,}} /\\\textit{\textbf{There is a rumor that}} <sentence1> \textit{\textbf{However, the truth is:}} ''} & \shortstack{\textit{\textbf{entailment/}}\\\textit{\textbf{neutral}}/\\\textit{\textbf{contradiction}}} \\
        \hline
        \\[-1em]
        \multirow{2}[2]{*}{\shortstack{AgNews\\(news articles\\classification)}} & zero-shot & \shortstack[l]{``The news articles is in the category of \textit{\textbf{World/Sports/Business/}}\textit{\textbf{Technology}}: ''} & \shortstack{\textit{\textbf{World/Sports/}}\\\textit{\textbf{Business/Technology}}} \\
        \cline{2-4}
        \\[-1em]
        ~ & few-shot & \shortstack[l]{``The news article is: \textit{<sample\_1>}\\The news article is: \textit{<sample\_2>}\\...\\The news article is: \textit{<sample\_S>}\\ \\The new news article in the category of \textit{\textbf{World/Sports/Business/Technology}}\\which is diverse in the expression compared to the above given samples is: ''} & \shortstack{\textit{\textbf{World/Sports/}}\\\textit{\textbf{Business/Technology}}} \\ 
        \hline
        \\[-1em]
        \multirow{2}[2]{*}{\shortstack{MarkedNews\\(self-defined news\\articles classification)}} & zero-shot & \shortstack[l]{``A news article in the category of \textit{\textbf{World that does not include `\$'/Sports that}}\\\textit{\textbf{does not include `\$'/Business that does not include `\$'/Technology that does}}\\\textit{\textbf{not include `\$'/Money with `\$' included}}: ''} & \shortstack{\textit{\textbf{World/Sports/}}\\\textit{\textbf{Business/Technology/}}\\\textit{\textbf{Money with \$ included}}} \\
        \cline{2-4}
        \\[-1em]
        ~ & few-shot & \shortstack[l]{``The news article is: \textit{<sample\_1>}\\The news article is: \textit{<sample\_2>}\\...\\The news article is: \textit{<sample\_S>}\\ \\The new news article in the category of \textit{\textbf{World that does not include `\$'/}}\\\textit{\textbf{Sports that does not include `\$'/Business that does not include `\$'/}}\\\textit{\textbf{Technology that does not include `\$'/Money with `\$' included}} which is\\diverse in the expression compared to the above given samples is: ''} & \shortstack{\textit{\textbf{World/Sports/}}\\\textit{\textbf{Business/Technology/}}\\\textit{\textbf{Money with \$ included}}} \\
    \bottomrule
    \end{tabular}
    }
\caption{Prompt used for synthetic dataset generation.}
\label{tab:appendix_prompt}
\end{table*}

Below we also provide $2$ examples of the few-shot prompts used in FuseGen 
. We need to clarify that, label information is not included in the in-context samples. 

\begin{tcolorbox}[colback=white, colframe=black, coltitle=white, colbacktitle=black, fonttitle=\bfseries, title=Few-shot prompt for movie review semantic analysis, rounded corners=southwest, boxrule=0.5mm, arc=2mm, width=\linewidth, breakable]
\small
    The movie review is: This is an excellent romantic comedy that relies more on wit and character than on silly, typical formula. A lot of people I know walked away from this movie disappointed, but I found it an enjoyable experience. I also don't understand why Hollywood thinks that 'quirkiness' is more important than story, or why they can't seem to create movies in which the plot is interesting and makes sense.\\
    The movie review is: There's a lot of talent wasted here. Haggis overuses his themes and is unable to let his characters go in this soapy melodrama.\\
    The movie review is: The movie is not fast paced and some of the drama was a bit too much for me, but I did like it.\\
    The movie review is: There is a certain helplessness in allowing ourselves to be tricked by the tricky cuts that grace the first half of the film. It allows us to suspend our disbelief and see what we want to see. It's not a movie I'd love to watch again, but it is one I'm glad I got to see.\\
    The movie review is: I will be the first to admit that the animation is crude in some parts. What I liked about the movie is that it had a very fun story line and I loved the songs.
    The movie review is: There's no reason you shouldn't enjoy this semi-tangential off-shoot of a popular video game; it's a fun, goofy movie that doesn't rely on the whole 'cinematic universe' concept\\
    The movie review is: engaging and entertaining, with excellent performances from David Niven and Barbara Stanwyck. 2.Sheila is stunning in the movie, a lady obsessed with the detective, especially when working in an area with limited light. 3.The climax is shocking - but it's entirely appropriate, as the plot's terrible.\\
    The movie review is: Many don't like the hero, and still others were glad they saw it and it was good. With that said, there are some surprising plot holes, inconsistencies and potential points of plot-holes that also need to be addressed before anyone can put their money into the film. If anyone was wondering how people like things and don't like other people like things, this movie is a great example.\\
    \\
    The new movie review in negative sentiment which is diverse in the expression compared to the above given samples is: 
\end{tcolorbox}

\begin{tcolorbox}[colback=white, colframe=black, coltitle=white, colbacktitle=black, fonttitle=\bfseries, title=Few-shot prompt for information-question entailment analysis, rounded corners=southwest, boxrule=0.5mm, arc=2mm, width=\linewidth, breakable]
\small
    The Information-Question pair is: Soon after, the account began to go viral, attracting the attention of reddit streams, content aggregators, art critics, and Renoir$\textbackslash$u2019s own descendants.[SEP]and Renoir's own accounts suggests that they met in early November 1881 when the baron stopped at their boardinghouse. ''Below a quadriga in the Louvre courtyard, Henri left his easel with his model and ran up the stairway to Duret with the idea of showing him what he had accomplished.`` (from Renoir's biography by Fr?\\
    The Information-Question pair is: She made her American debut in 1910, with the New York Symphony Orchestra, under conductor Walter Damrosch.[SEP]If this photo were to depict a specific moment in history, or an individual's life, which historical period or individual would it most closely resemble?\\
    The Information-Question pair is: The Fall Line is an American true crime podcast that covers lesser-known cases of murder and disappearance from minority communities in Georgia.[SEP]The founder is the founder. If the owner owns the club, is it the 'Alamo' of crime blogs (or is it an 'evil bar')?\\
    The Information-Question pair is: She was a Member of the Supreme Council of the Uzbek SSR.[SEP]Who was the head of the Uzbek SSR during her time on the Supreme Council?\\
    \\
    The new Information-Question pair which is diverse in the expression compared to the above given samples is: Information: ``<information>''\\
    Question (answer not in above information): 
\end{tcolorbox}

\section{Detailed Algorithms} \label{sec:appendix_algorithm}
 We provide the detailed algorithms for each function used in \cref{alg:algorithm_full_functions} here in \cref{alg:algorithm_functions}.

\begin{algorithm}[!htb]
\small
\caption{Functions used in \cref{alg:algorithm_full_functions} for FuseGen}
\label{alg:algorithm_functions}
    \textbf{function} \verb|S_AccumulativeSynDataGeneration(|$\mathcal{D}_k$, $\hat{\mathcal{D}}$, $N$, $J$, $j$\verb|)|:
    
\begin{algorithmic}[]
    \IF{$j=0$}
        \STATE Use zero-shot prompt as working prompt $\mathcal{T}$.
    \ELSE
        \STATE Use $\hat{\mathcal{D}}$ to create few-shot prompt as working prompt $\mathcal{T}$.
    \ENDIF
    \STATE Generate $\frac{N}{J+1}$ samples using $\mathcal{T}$ and add them to $\mathcal{D}_k$.
    \STATE \textbf{return} $\mathcal{D}_k$.

    \STATE
\end{algorithmic}

    \textbf{function} \verb|S_STMTraining(|$\mathcal{D}$, $m_{(0)}$, $E_2$\verb|)|:
    
\begin{algorithmic}[]
    \STATE Initialize a trainable STM ${m} \leftarrow m_{(0)}$ and train ${m}$ using $\mathcal{D}_k$ for $E_2$ epochs with \cref{eq:stm_train_loss}.
    \STATE \textbf{return} $m$.

    \STATE
\end{algorithmic}

    \textbf{function} \verb|C_SampleSelection(|$\mathcal{D}$, $\{m_k\}_{k=1}^K$, $\tilde{m}$, $\alpha$, $R$, $S$\verb|)|:
    
\begin{algorithmic}[]
    \STATE Reset $\hat{\mathcal{D}} \leftarrow \emptyset$.
    \FOR{$k'=1$ {\bfseries to} $K$} 
        \FOR{Each sample $(\mathbf{x}_{k,i},y_{k,i})$ in $\mathcal{D}$}
            \STATE Obtain the prediction vector $p_{k',k,i}=m_{k'}(\mathbf{x}_{k,i}) \in \mathbb{R}^{C}$ and predicted label-position probability $p_{k',k,i}[y_{k,i}] \in \mathbb{R}^{1}$.
            \STATE Calculate disagreement score $d_{k,i}=\mathrm{STD}(p_{1,k,i}[y_{k,i}],...,p_{k',k,i}[y_{k,i}],...,p_{K,k,i}[y_{k,i}])$.
        \ENDFOR
    \ENDFOR
    \STATE Sort all the samples within $\mathcal{D}$ and add the top-$(1-\alpha)R$ samples with the lowest score and top-$\alpha R$ samples with the highest samples into $\hat{\mathcal{D}}$.
    \STATE Calculate the influence score of each sample in $\hat{\mathcal{D}}$ with $\tilde{m}$ using Eq.(3) in \citet{ye2022progen}.
    \STATE $\hat{\mathcal{D}} \leftarrow \{$top-$S$ samples with the highest influence score$\}$.
    \STATE \textbf{return} $\hat{\mathcal{D}}$.

    \STATE
\end{algorithmic}

     \textbf{function} \verb|S_WeightAdjustSTMTraining(|$\mathcal{D}$, $m_{(0)}$, $\{w_{i}^{(0)}\}_{i=1}^{N}$, $E_1$, $E_2$\verb|)|:
     
\begin{algorithmic}[]
    \FOR{$e_{1}=0$ {\bfseries to} $E_1-1$}
        \STATE Initialize a trainable STM ${m} \leftarrow m_{(0)}$ and train ${m}$ using $\mathcal{D}$ for $E_2$ epochs with weighted loss using $\{w_{i}^{(e_1)}\}_{i=1}^{N}$ and \cref{eq:weighted_loss}.
        \STATE Adjust sample-level weight $w_{i}^{(e_1+1)}\leftarrow w_{i}^{(e_1)}$ with ${m}$ using \cref{eq:weight_adjust_onlywrong_onlyself_loss} for each sample $(\mathbf{x}_{i},y_{i}),\,i=1,\dots,N$.
    \ENDFOR
    \STATE \textbf{return} $m$.

\end{algorithmic}

\end{algorithm}

\section{Additional Experimental Results} \label{sec:appendix_experiments}

\subsection{Dataset Cartography of More Synthetic Datasets} \label{subsec:appendix_dataset_cartography} 

Dataset cartography~\cite{swayamdipta2020dataset} approach characterizes each sample by its confidence and variability, which are defined as the mean and standard deviation of the model probability of its related label across all training epochs. For example, if the model correctly predict a sample's label across training epochs, it will have high confidence and low variability. These samples are regarded as \textit{easy-to-learn} samples 
, whereas those with low variability yet low confidence are identified as \textit{hard-to-learn} samples. Conversely, samples with high variability are deemed \textit{ambiguous}. 

\begin{figure*}
\vspace{-3em}
    \centering
    \subfigure[GPT-2 ZeroGen K=1 (84.83)]{
         \begin{minipage}[t]{0.26\linewidth}
         \centering
         \includegraphics[width=1\linewidth]{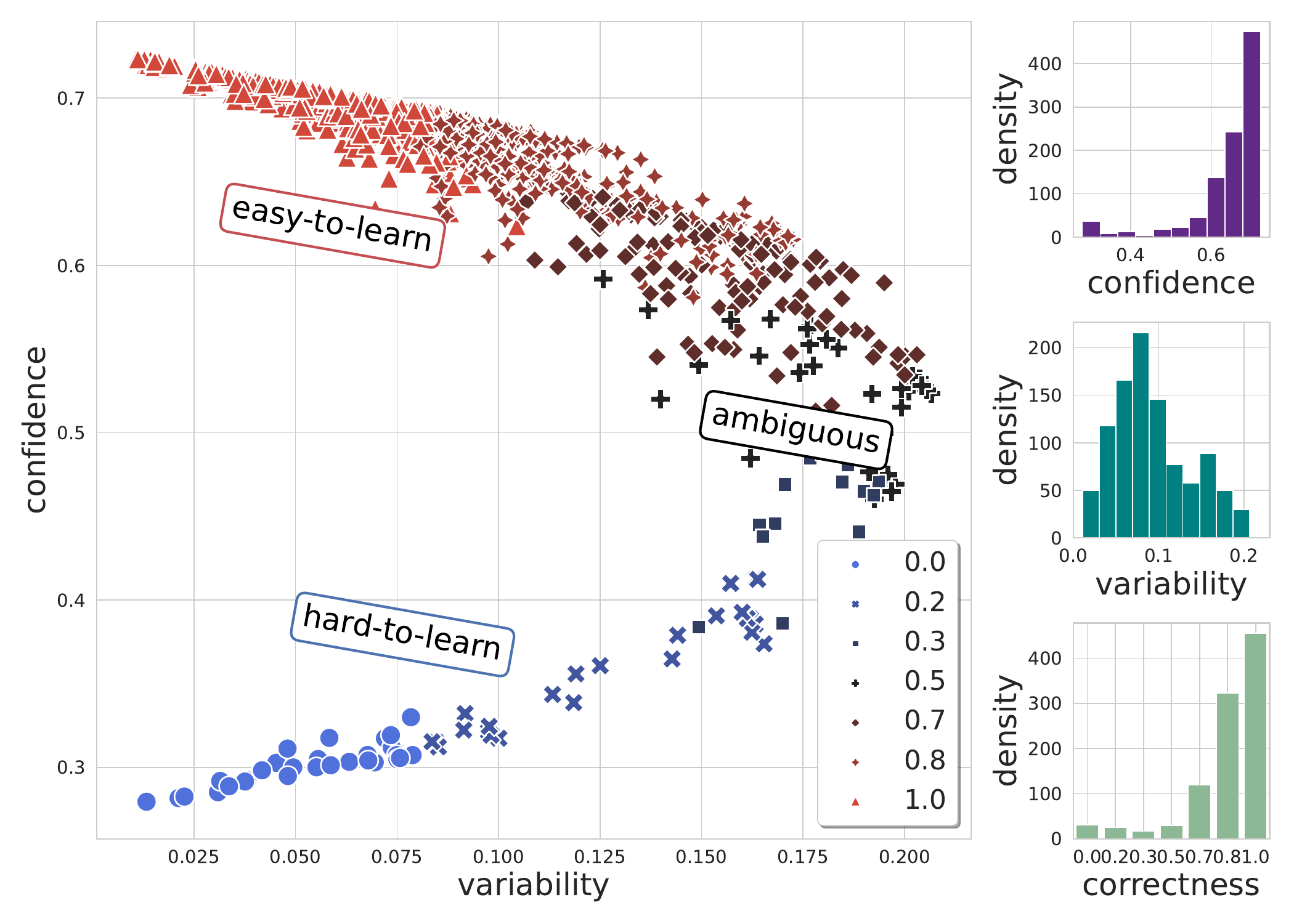}
         \vspace{-2em}
         \label{subfig:dataset_cartography_original_gpt}
         \end{minipage}
     }
     \vspace{-0.5em}
     \subfigure[GPT-2 ProGen K=1 (85.74)]{
         \begin{minipage}[t]{0.26\linewidth}
         \centering
         \includegraphics[width=1\linewidth]{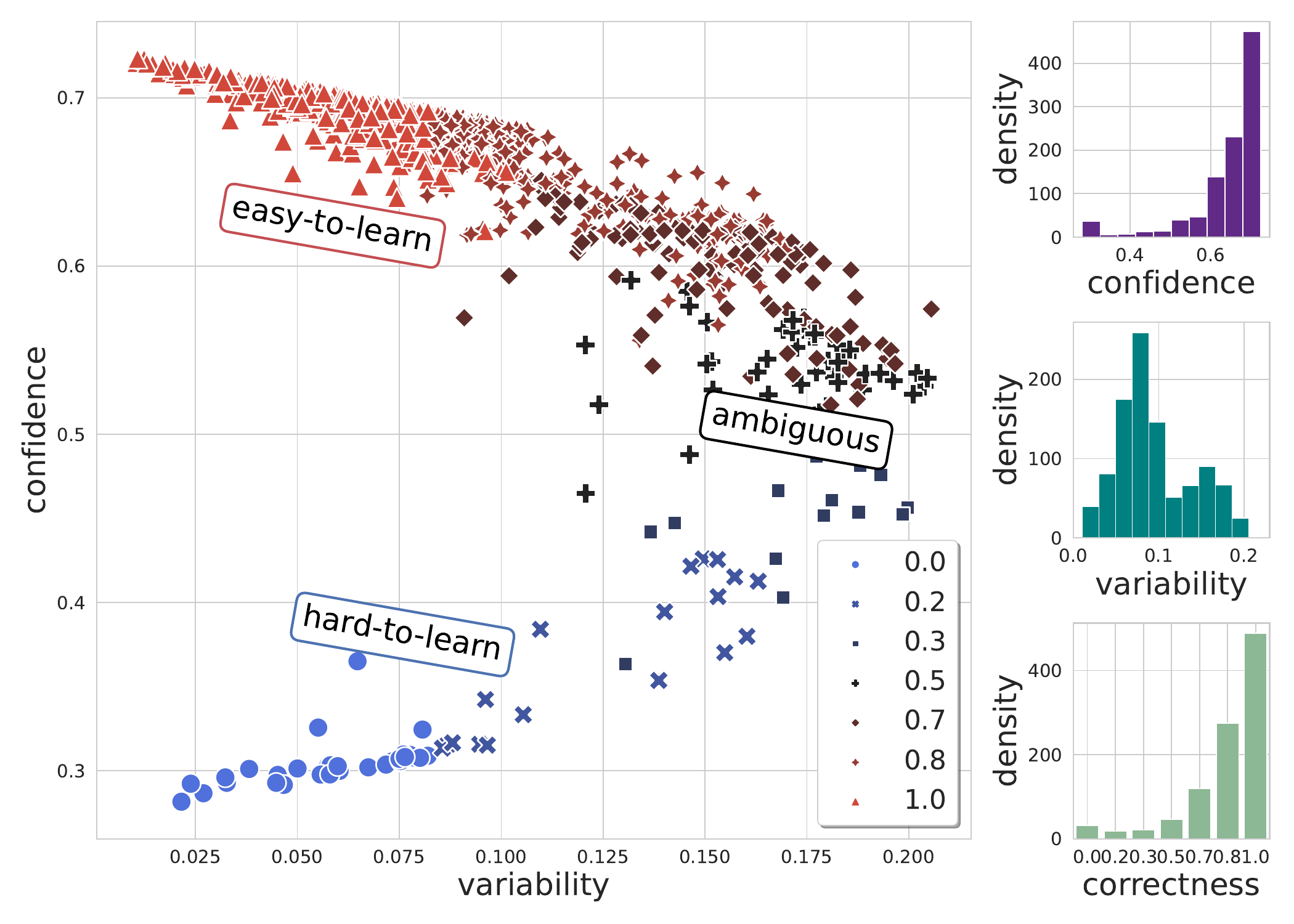}
         \vspace{-2em}
         \label{subfig:dataset_cartography_single_progen_gpt}
         \end{minipage}
     }
     \vspace{-0.5em}
     \subfigure[GPT-2 Ours K=6 (87.85)]{
         \begin{minipage}[t]{0.26\linewidth}
         \centering
         \includegraphics[width=1\linewidth]{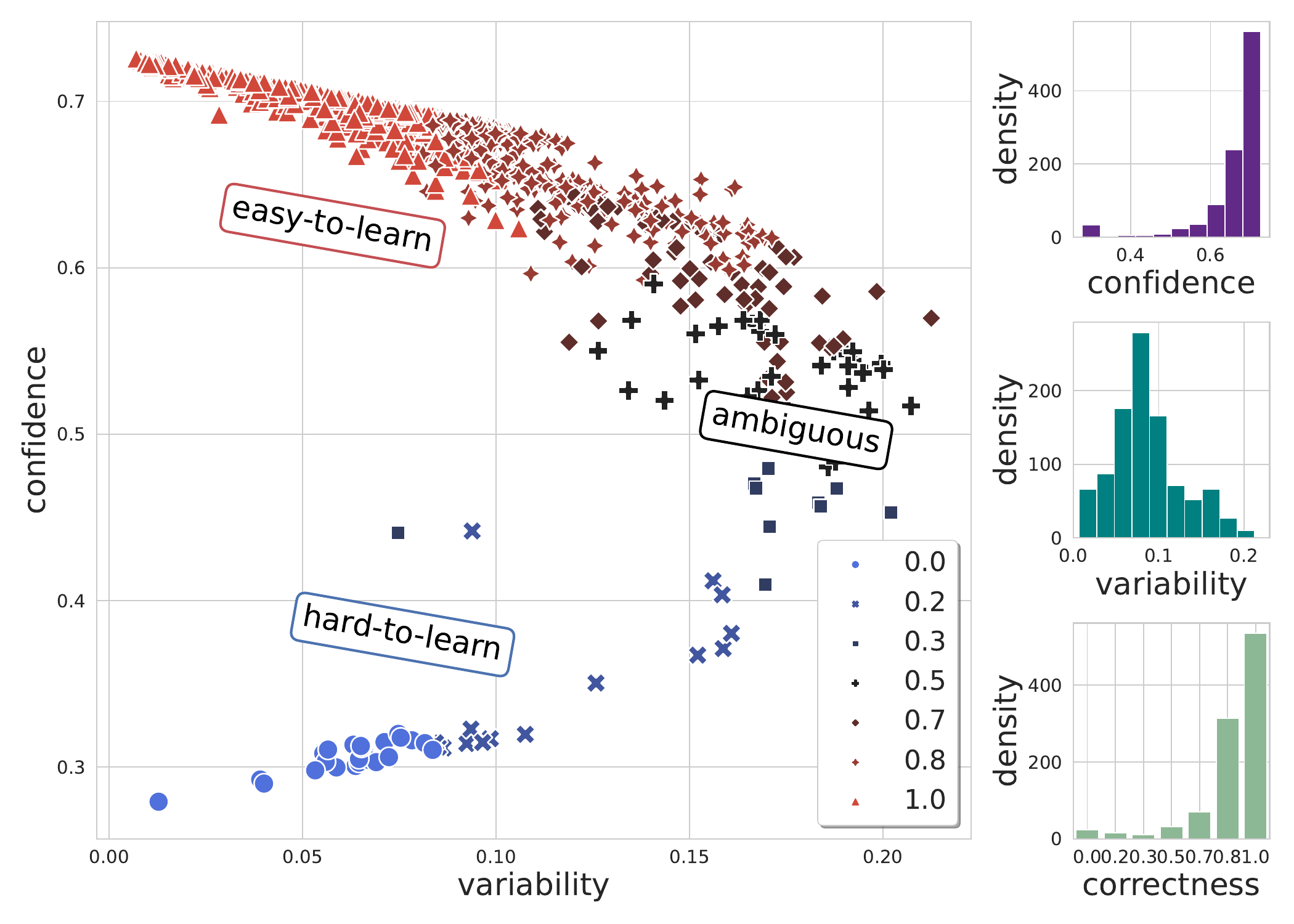}
         \vspace{-2em}
         \label{subfig:dataset_cartography_fusegen_gpt}
         \end{minipage}
     }
     \vspace{-0.5em}
    \subfigure[Llama-2 ZeroGen K=1 (84.23)]{
         \begin{minipage}[t]{0.26\linewidth}
         \centering
         \includegraphics[width=1\linewidth]{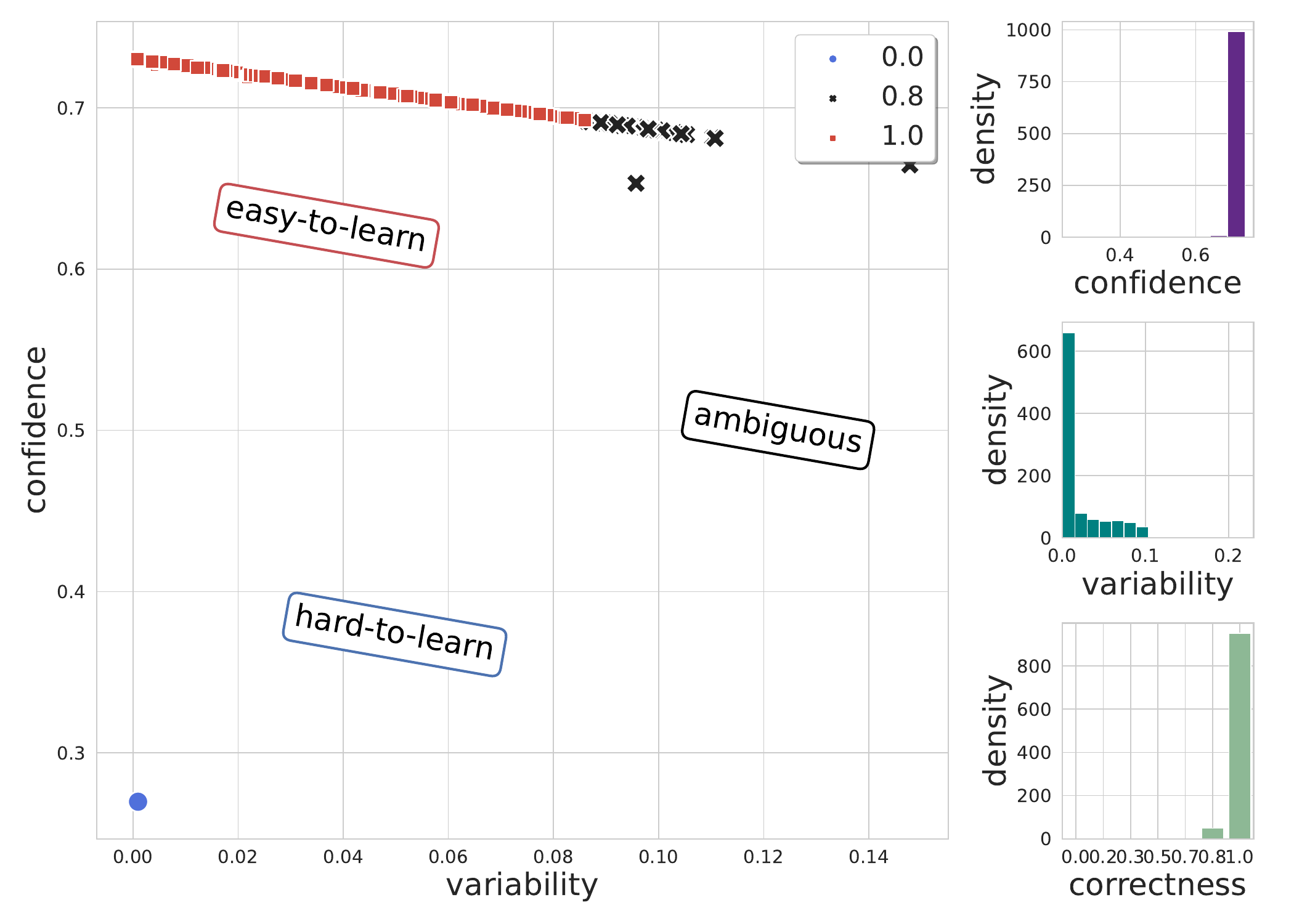}
         \vspace{-2em}
         \label{subfig:dataset_cartography_original_llama}
         \end{minipage}
     }
     \vspace{-0.2em}
     \subfigure[Llama-2 ProGen K=1 (84.24)]{
         \begin{minipage}[t]{0.26\linewidth}
         \centering
         \includegraphics[width=1\linewidth]{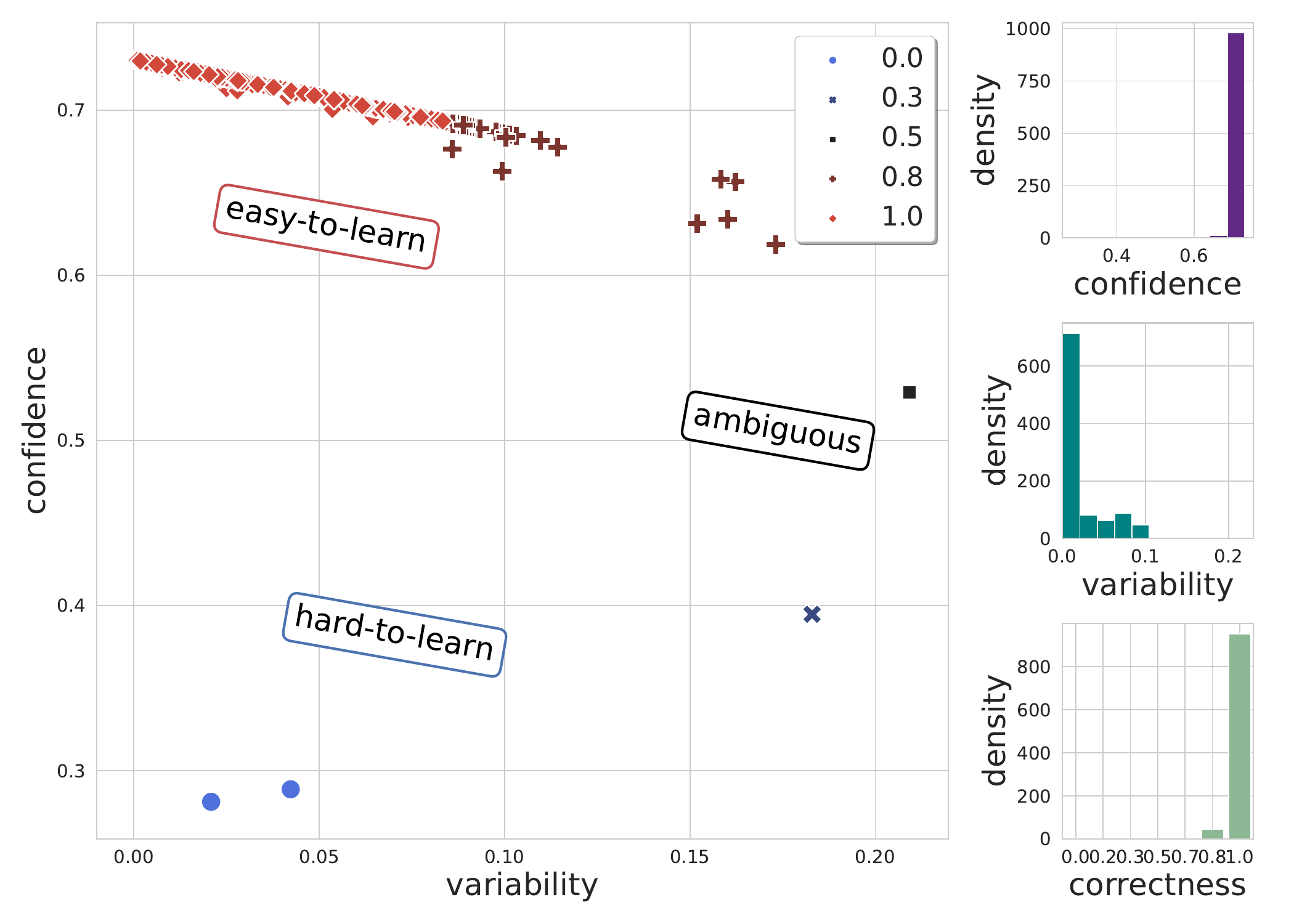}
         \vspace{-2em}
         \label{subfig:dataset_cartography_single_progen_llama}
         \end{minipage}
     }
     \vspace{-0.2em}
     \subfigure[Llama-2 Ours K=6 (86.60)]{
         \begin{minipage}[t]{0.26\linewidth}
         \centering
         \includegraphics[width=1\linewidth]{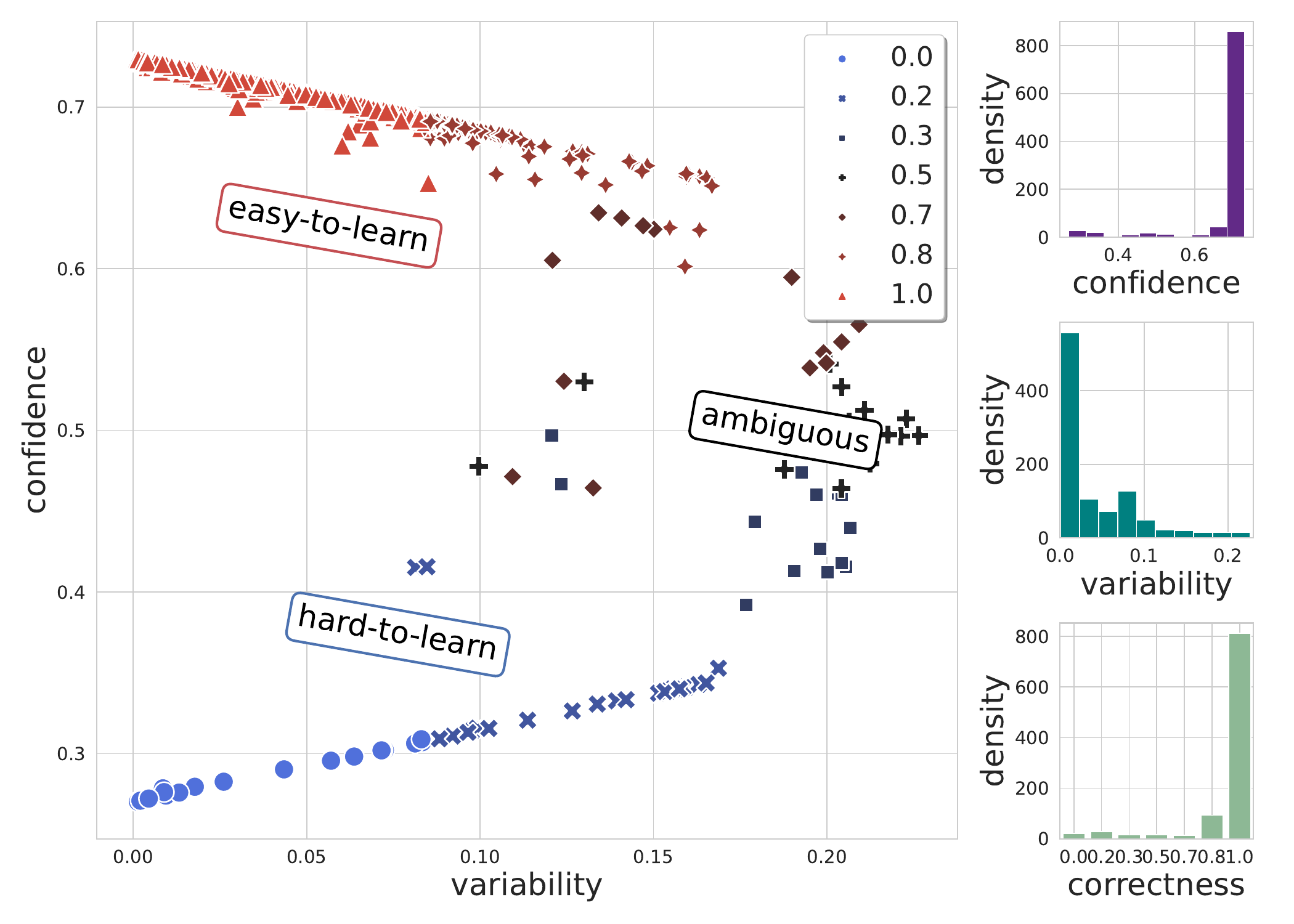}
         \vspace{-2em}
         \label{subfig:dataset_cartography_fusegen_llama}
         \end{minipage}
     }
     \vspace{-0.2em}
    \subfigure[Vicuna ZeroGen K=1 (82.37)]{
         \begin{minipage}[t]{0.26\linewidth}
         \centering
         \includegraphics[width=1\linewidth]{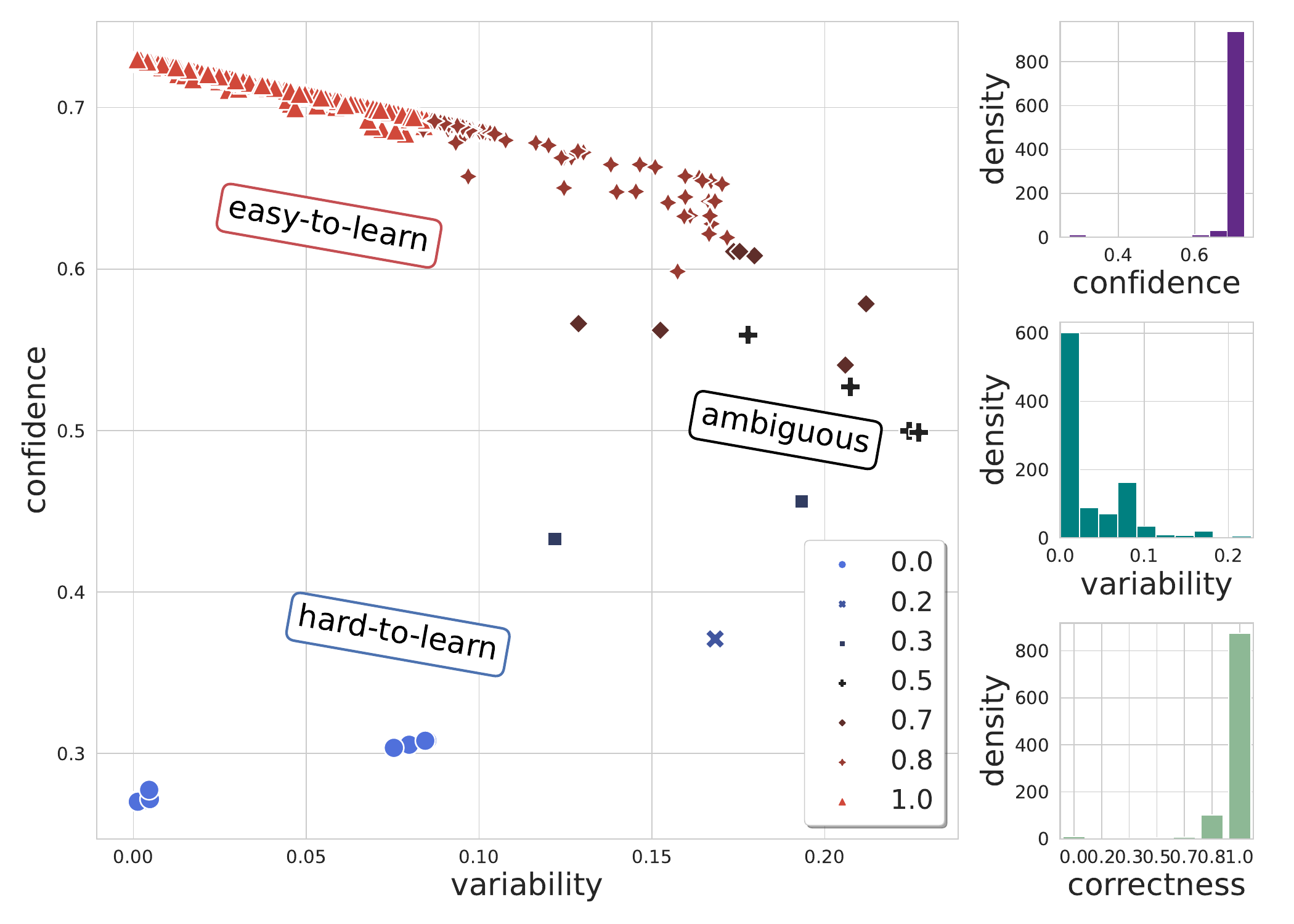}
         \vspace{-2em}
         \label{subfig:dataset_cartography_original_vicuna}
         \end{minipage}
     }
     \vspace{-0.3em}
     \subfigure[Vicuna ProGen K=1 (83.60)]{
         \begin{minipage}[t]{0.26\linewidth}
         \centering
         \includegraphics[width=1\linewidth]{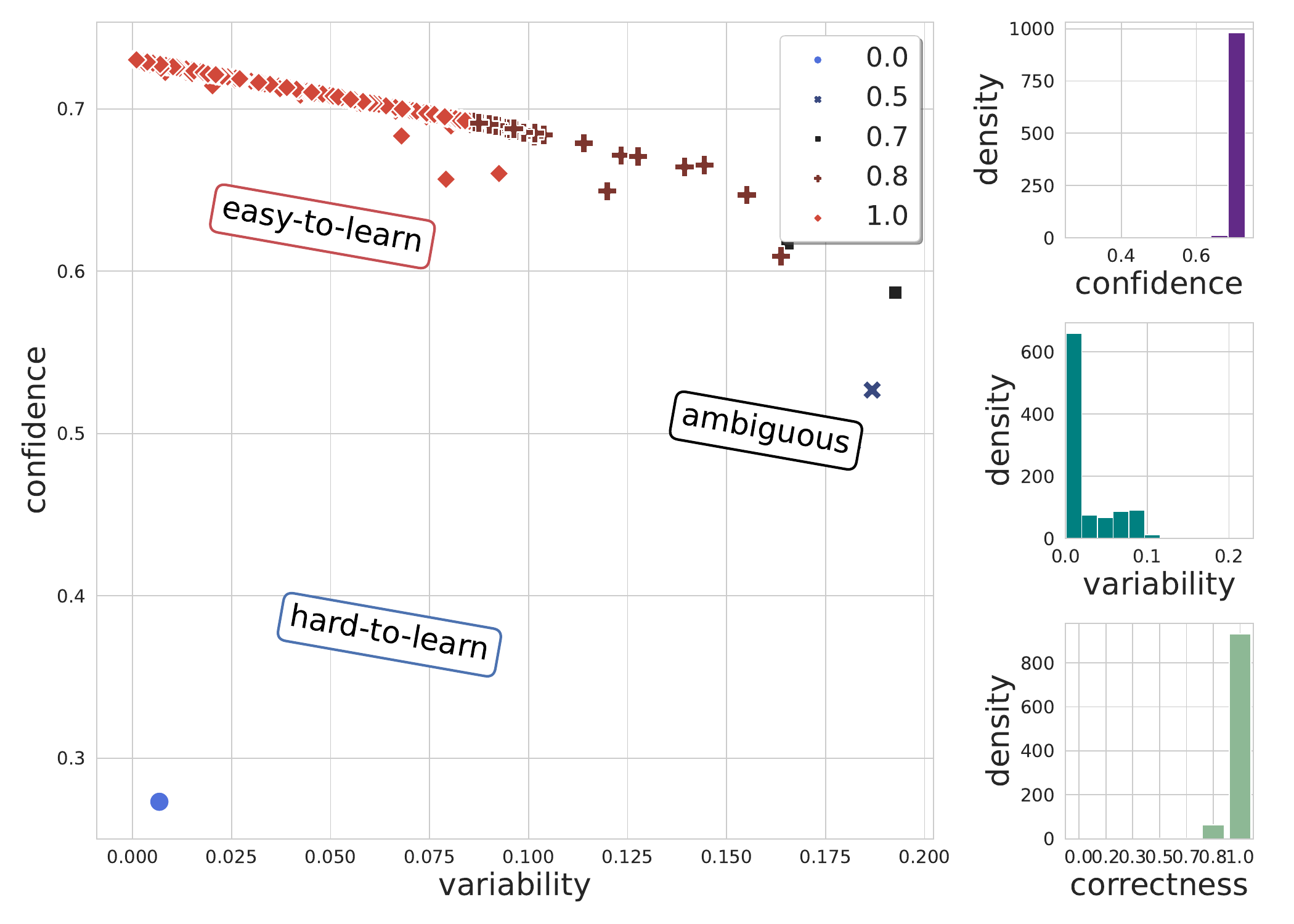}
         \vspace{-2em}
         \label{subfig:dataset_cartography_single_progen_vicuna}
         \end{minipage}
     }
     \vspace{-0.3em}
     \subfigure[Vicuna Ours K=6 (87.50)]{
         \begin{minipage}[t]{0.26\linewidth}
         \centering
         \includegraphics[width=1\linewidth]{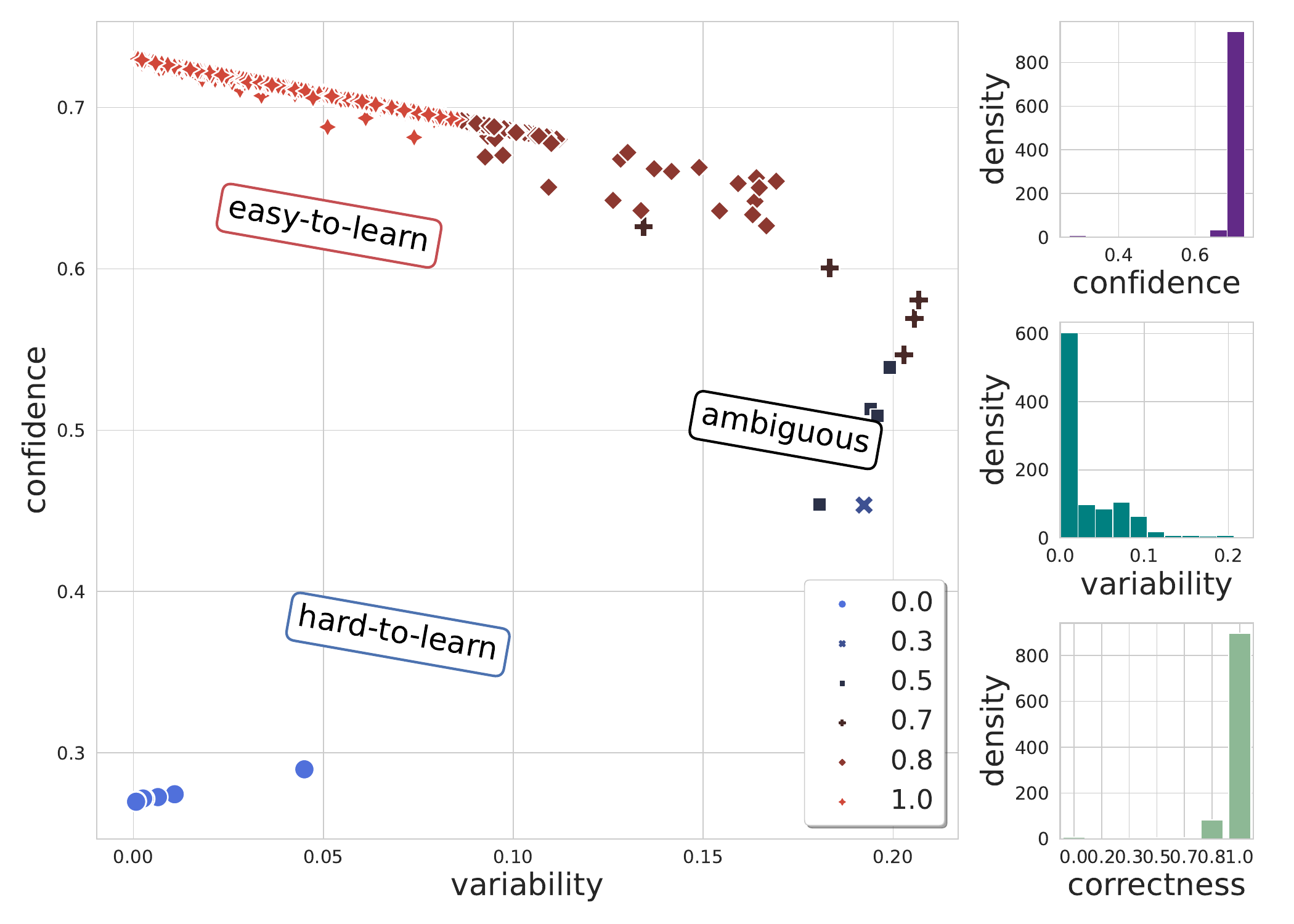}
         \vspace{-2em}
         \label{subfig:dataset_cartography_fusegen_vicuna}
         \end{minipage}
     }
     \vspace{-0.3em}
    \subfigure[OPT ZeroGen K=1 (84.97)]{
         \begin{minipage}[t]{0.26\linewidth}
         \centering
         \includegraphics[width=1\linewidth]{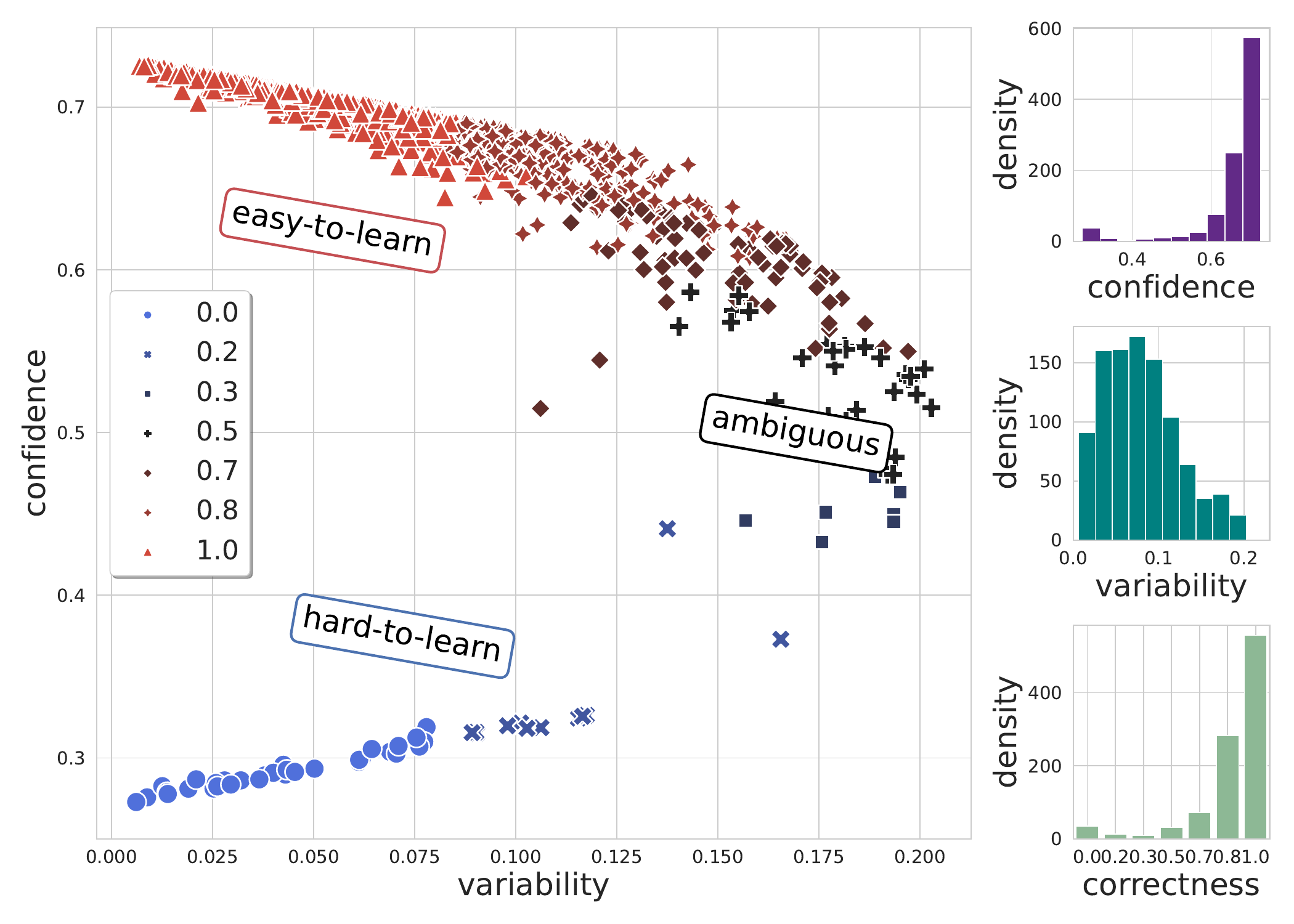}
         \vspace{-2em}
         \label{subfig:dataset_cartography_original_opt}
         \end{minipage}
     }
     \vspace{-0.3em}
     \subfigure[OPT ProGen K=1 (87.57)]{
         \begin{minipage}[t]{0.26\linewidth}
         \centering
         \includegraphics[width=1\linewidth]{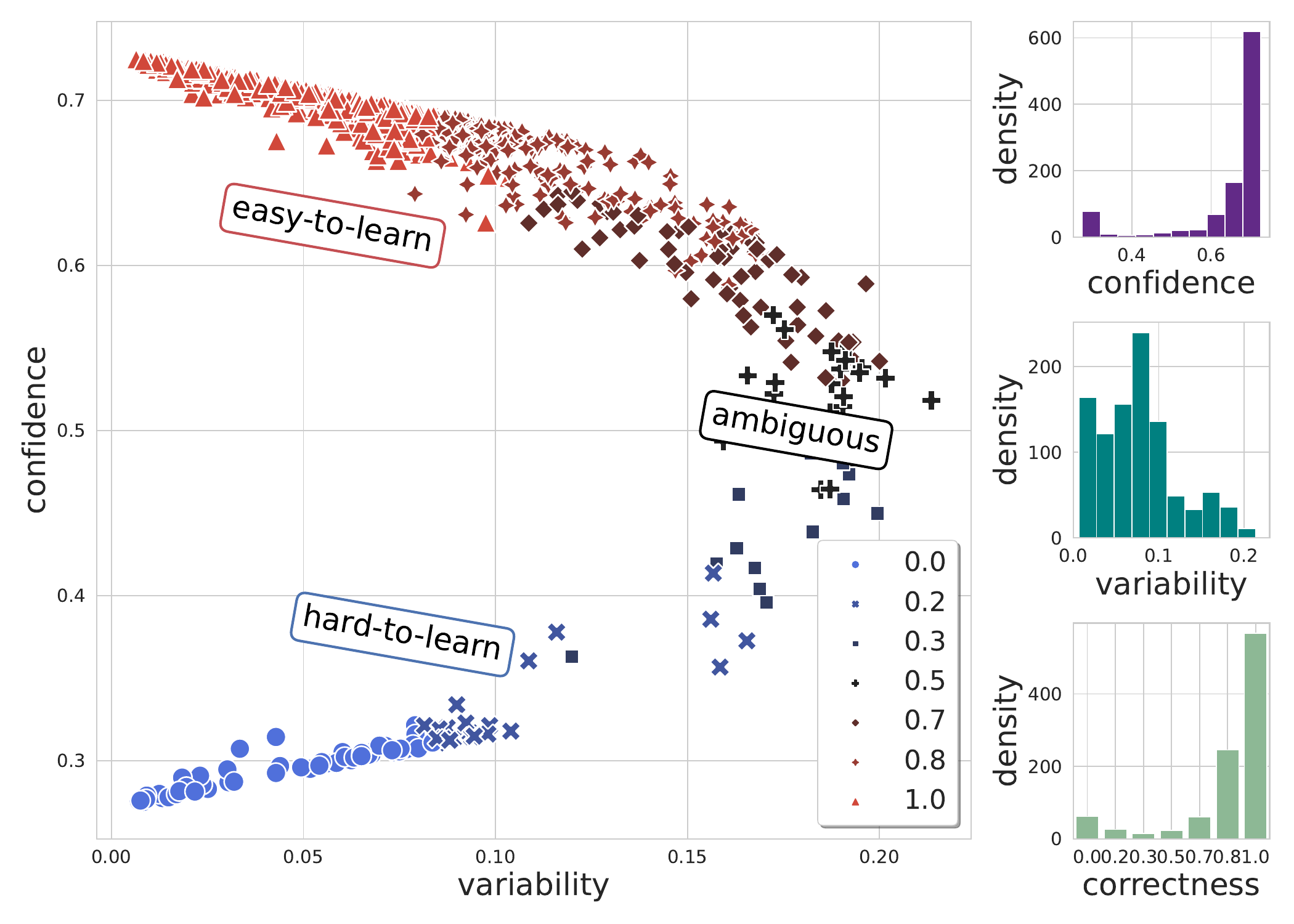}
         \vspace{-2em}
         \label{subfig:dataset_cartography_single_progen_opt}
         \end{minipage}
     }
     \vspace{-0.3em}
     \subfigure[OPT Ours K=6 (88.47)]{
         \begin{minipage}[t]{0.26\linewidth}
         \centering
         \includegraphics[width=1\linewidth]{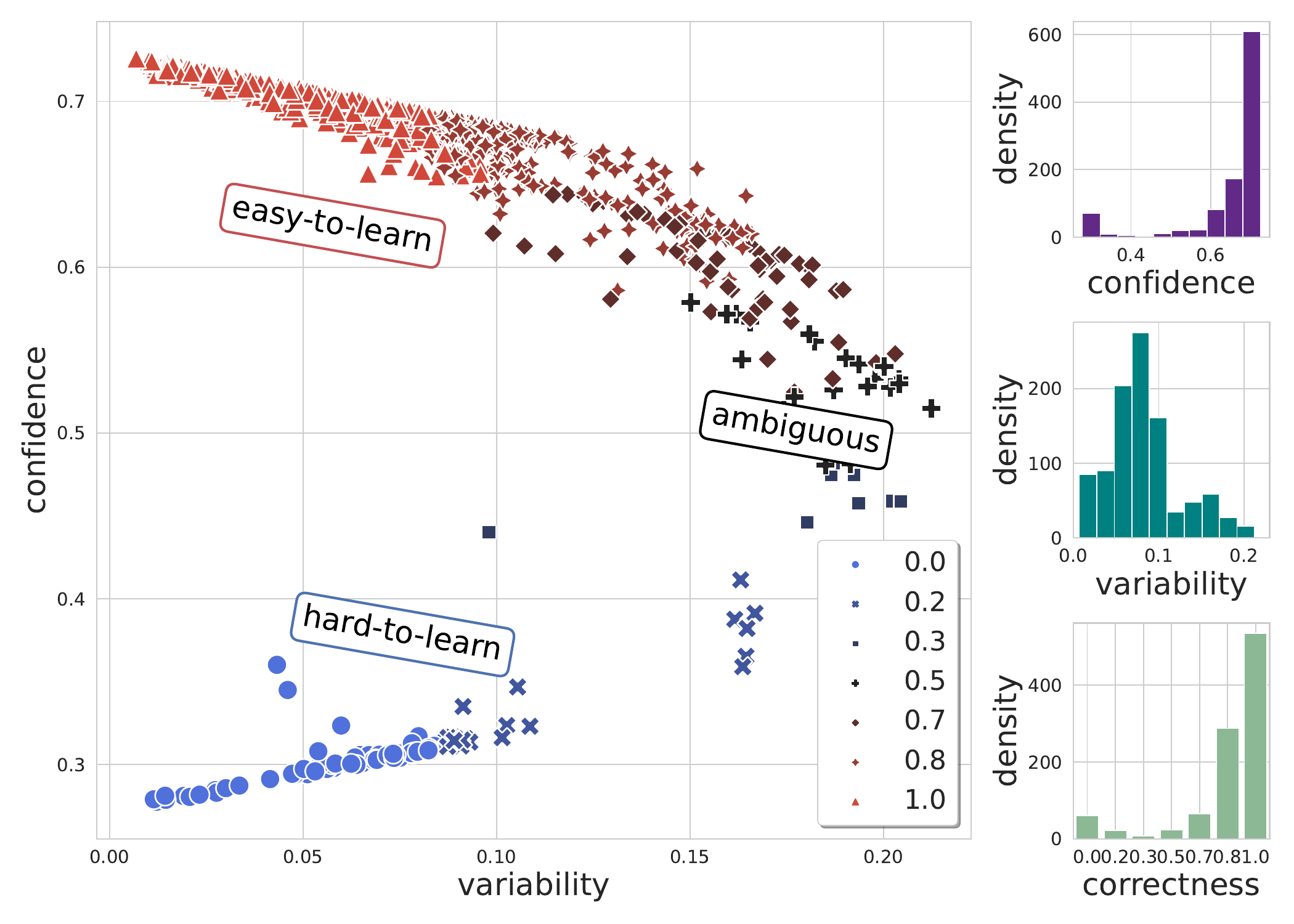}
         \vspace{-2em}
         \label{subfig:dataset_cartography_fusegen_opt}
         \end{minipage}
     }
     \vspace{-0.3em}
    \subfigure[ChatGLM3 ZeroGen K=1 (86.43)]{
         \begin{minipage}[t]{0.26\linewidth}
         \centering
         \includegraphics[width=1\linewidth]{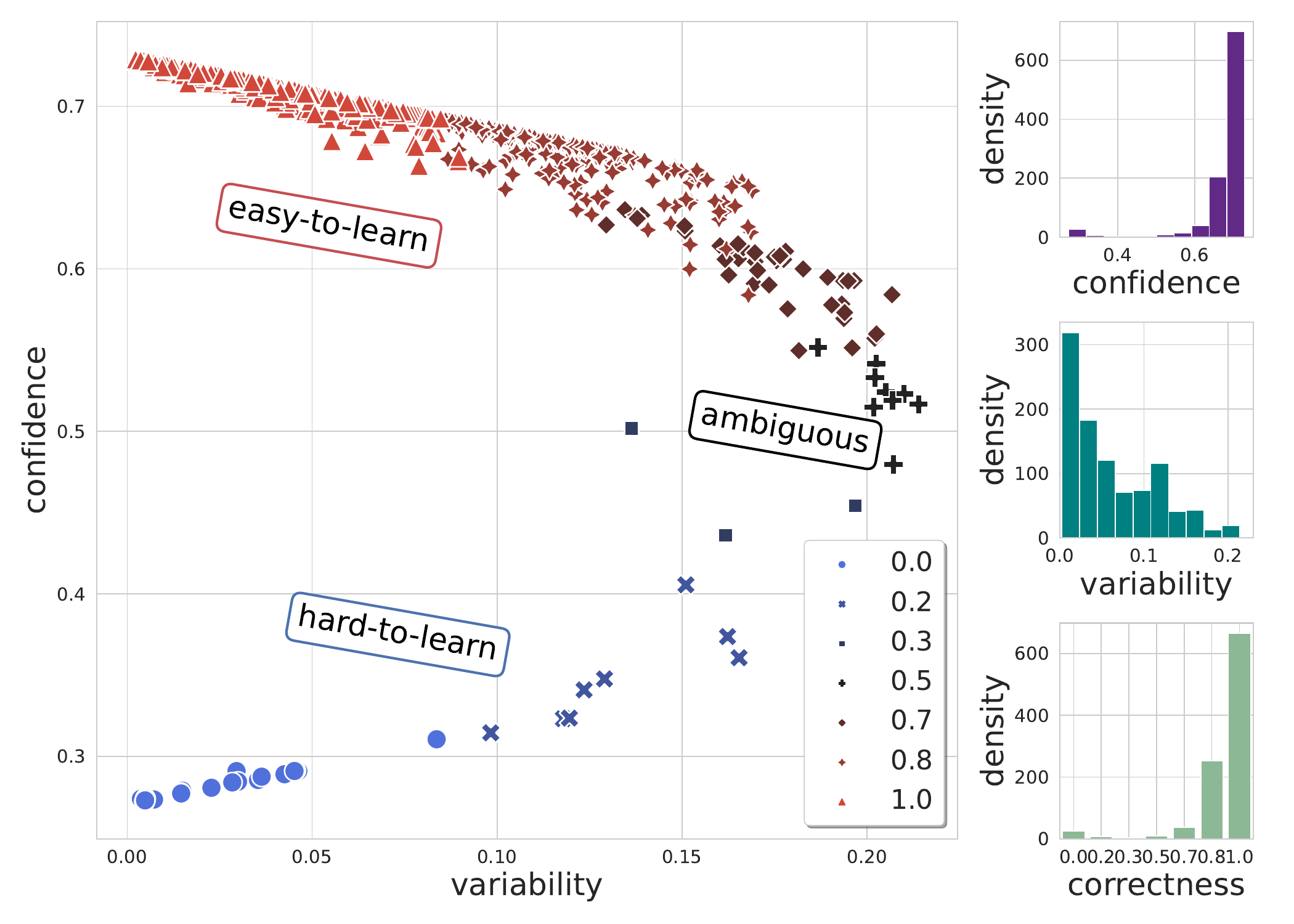}
         \vspace{-2em}
         \label{subfig:dataset_cartography_original_chatglm}
         \end{minipage}
     }
     \vspace{-0.3em}
     \subfigure[ChatGLM3 ProGen K=1 (87.07)]{
         \begin{minipage}[t]{0.26\linewidth}
         \centering
         \includegraphics[width=1\linewidth]{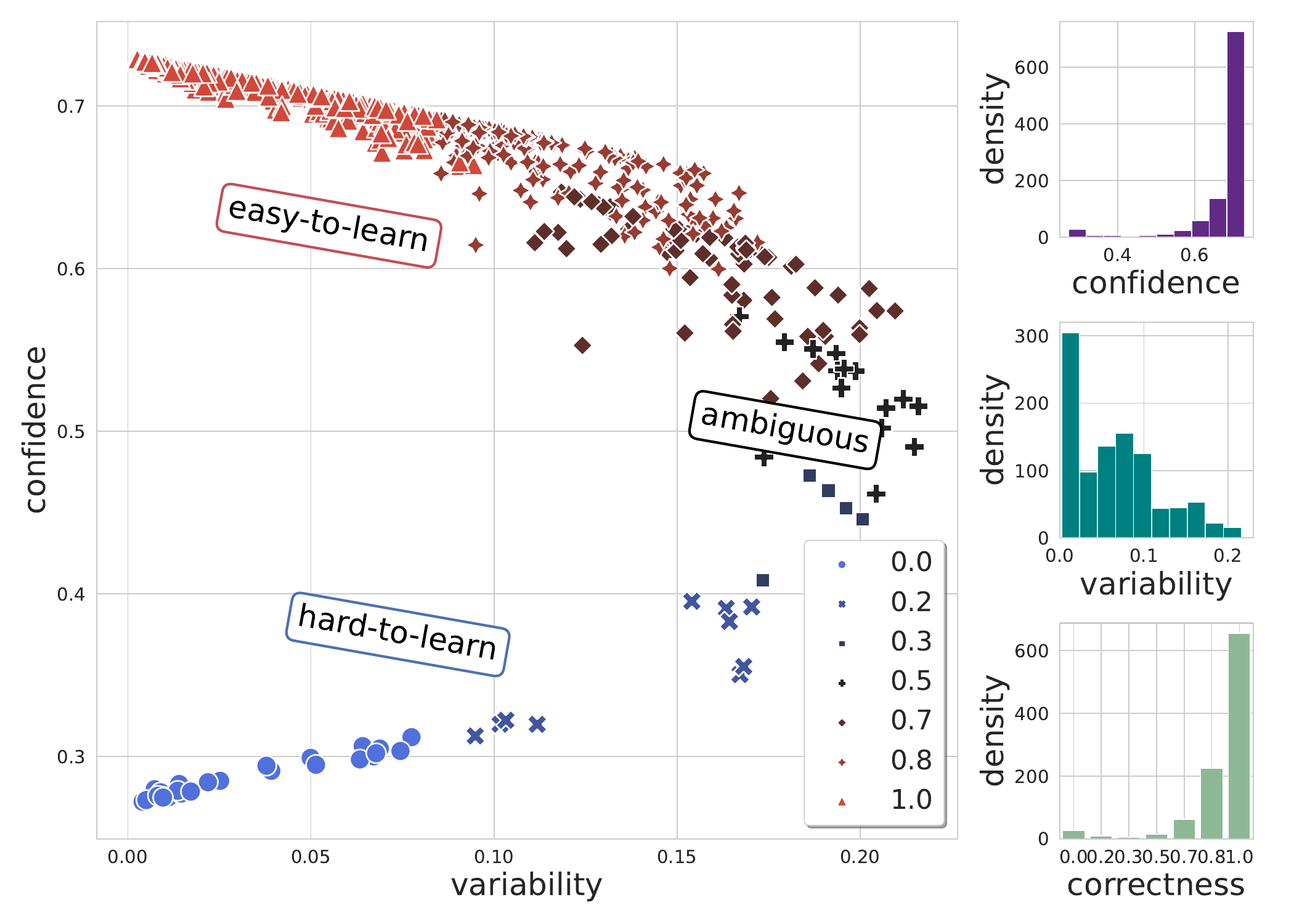}
         \vspace{-2em}
         \label{subfig:dataset_cartography_single_progen_chatglm}
         \end{minipage}
     }
     \vspace{-0.3em}
     \subfigure[ChatGLM3 Ours K=6 (88.56)]{
         \begin{minipage}[t]{0.26\linewidth}
         \centering
         \includegraphics[width=1\linewidth]{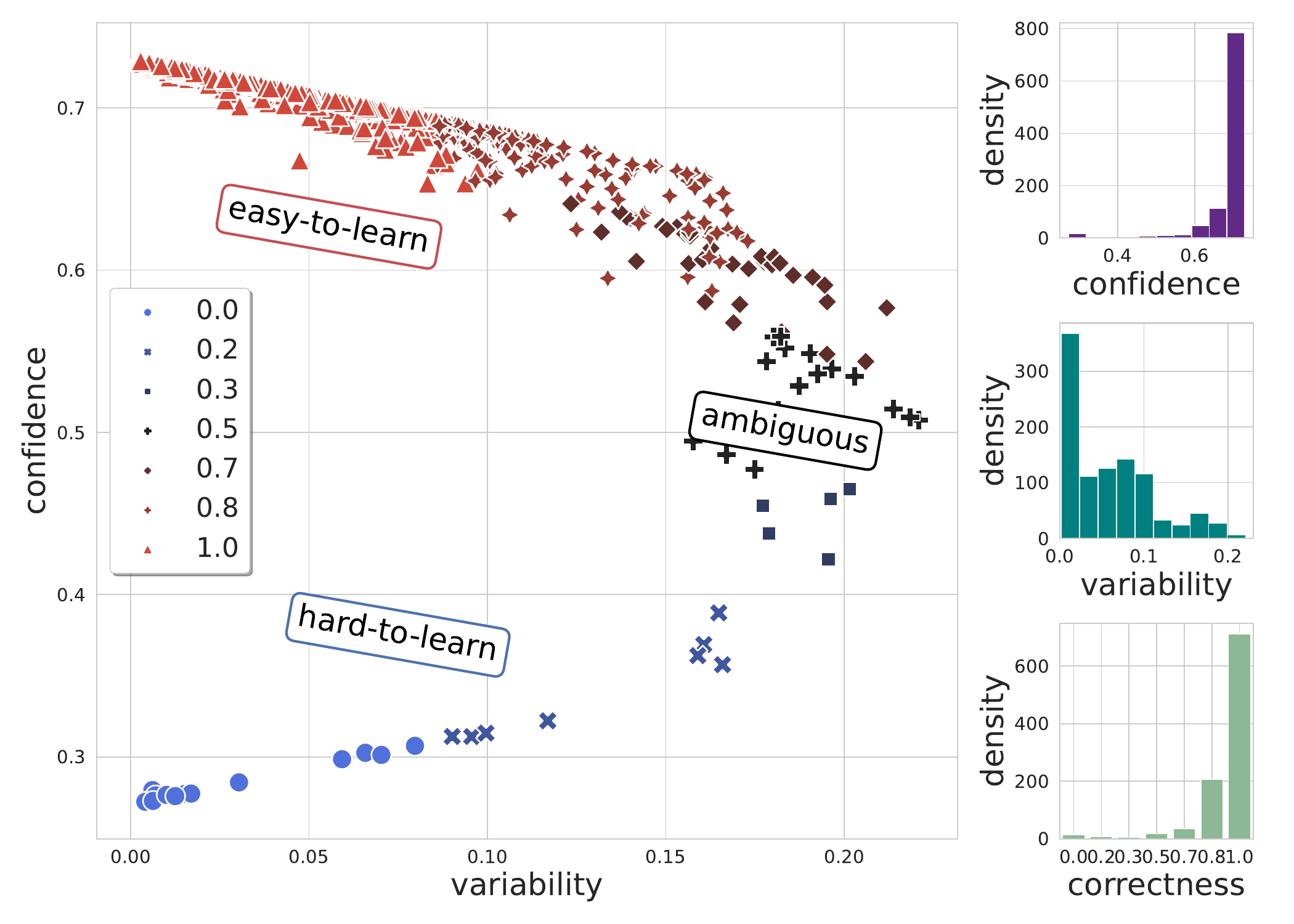}
         \vspace{-2em}
         \label{subfig:dataset_cartography_fusegen_chatglm}
         \end{minipage}
     }
     \vspace{-0.3em}
    \subfigure[Flan-T5 ZeroGen K=1 (88.18)]{
         \begin{minipage}[t]{0.26\linewidth}
         \centering
         \includegraphics[width=1\linewidth]{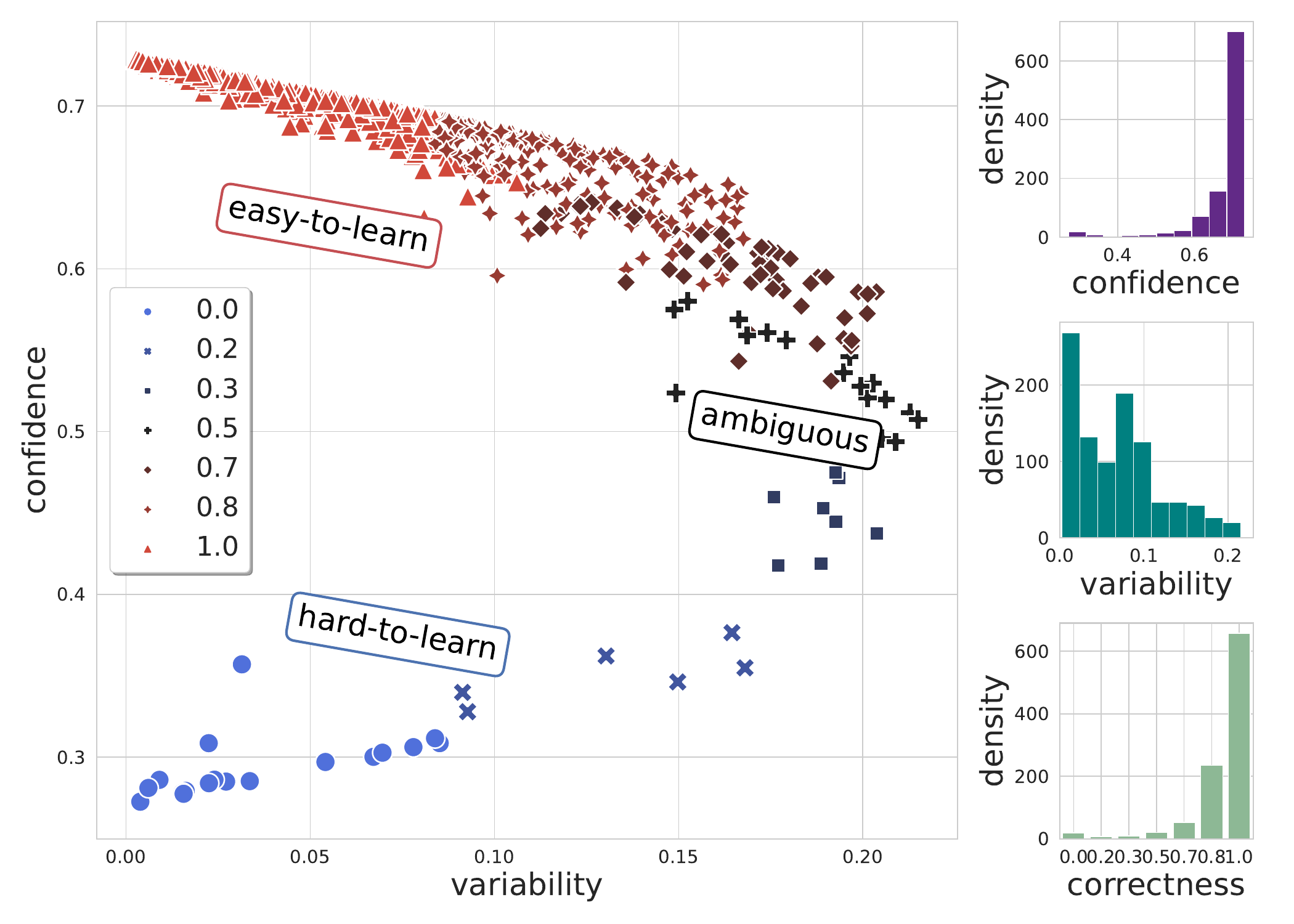}
         \vspace{-2em}
         \label{subfig:dataset_cartography_original_flant5}
         \end{minipage}
     }
     \subfigure[Flan-T5 ProGen K=1 (85.80)]{
         \begin{minipage}[t]{0.26\linewidth}
         \centering
         \includegraphics[width=1\linewidth]{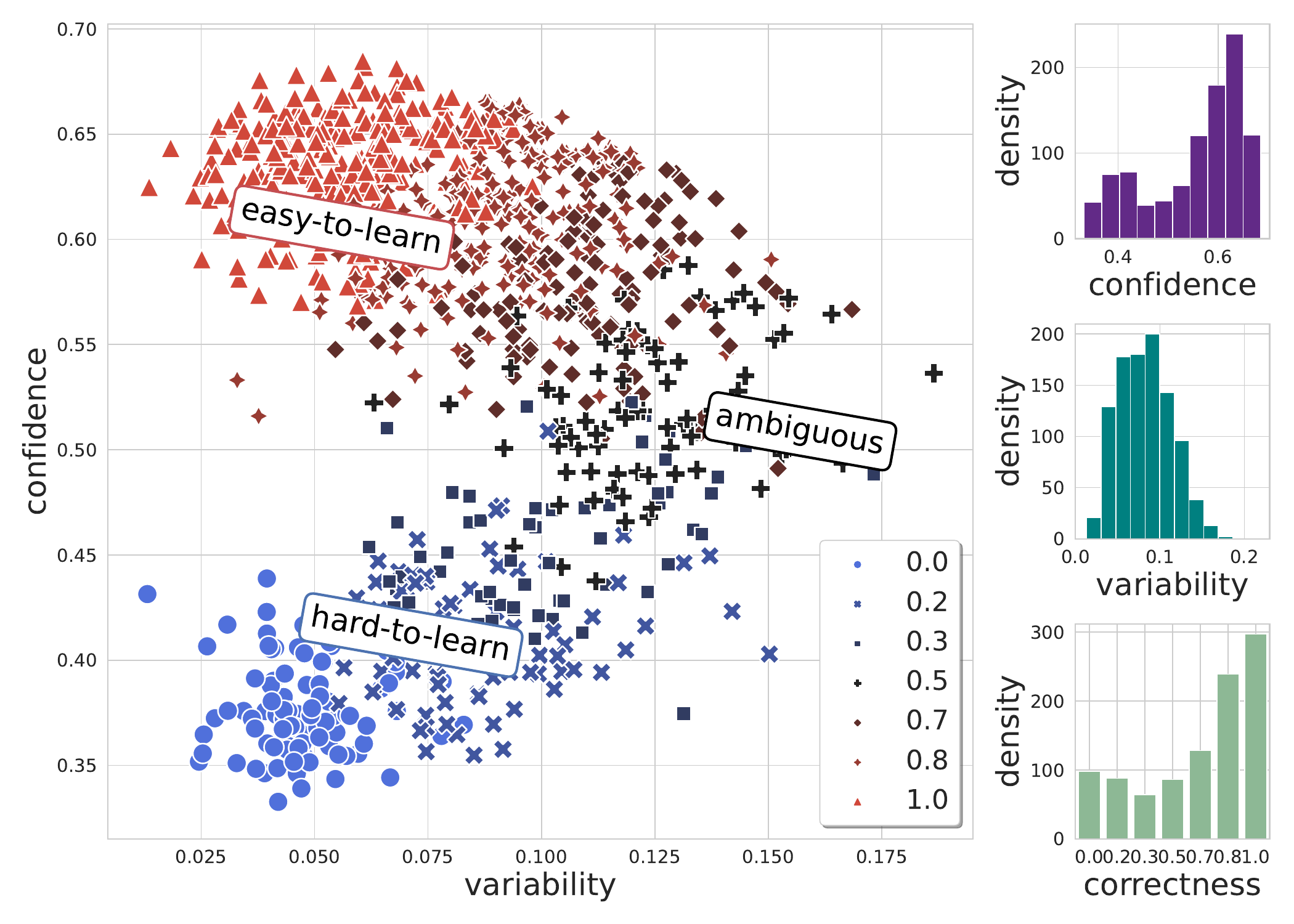}
         \vspace{-2em}
         \label{subfig:dataset_cartography_single_progen_flant5}
         \end{minipage}
     }
     \subfigure[Flan-T5 Ours K=6 (88.73)]{
         \begin{minipage}[t]{0.26\linewidth}
         \centering
         \includegraphics[width=1\linewidth]{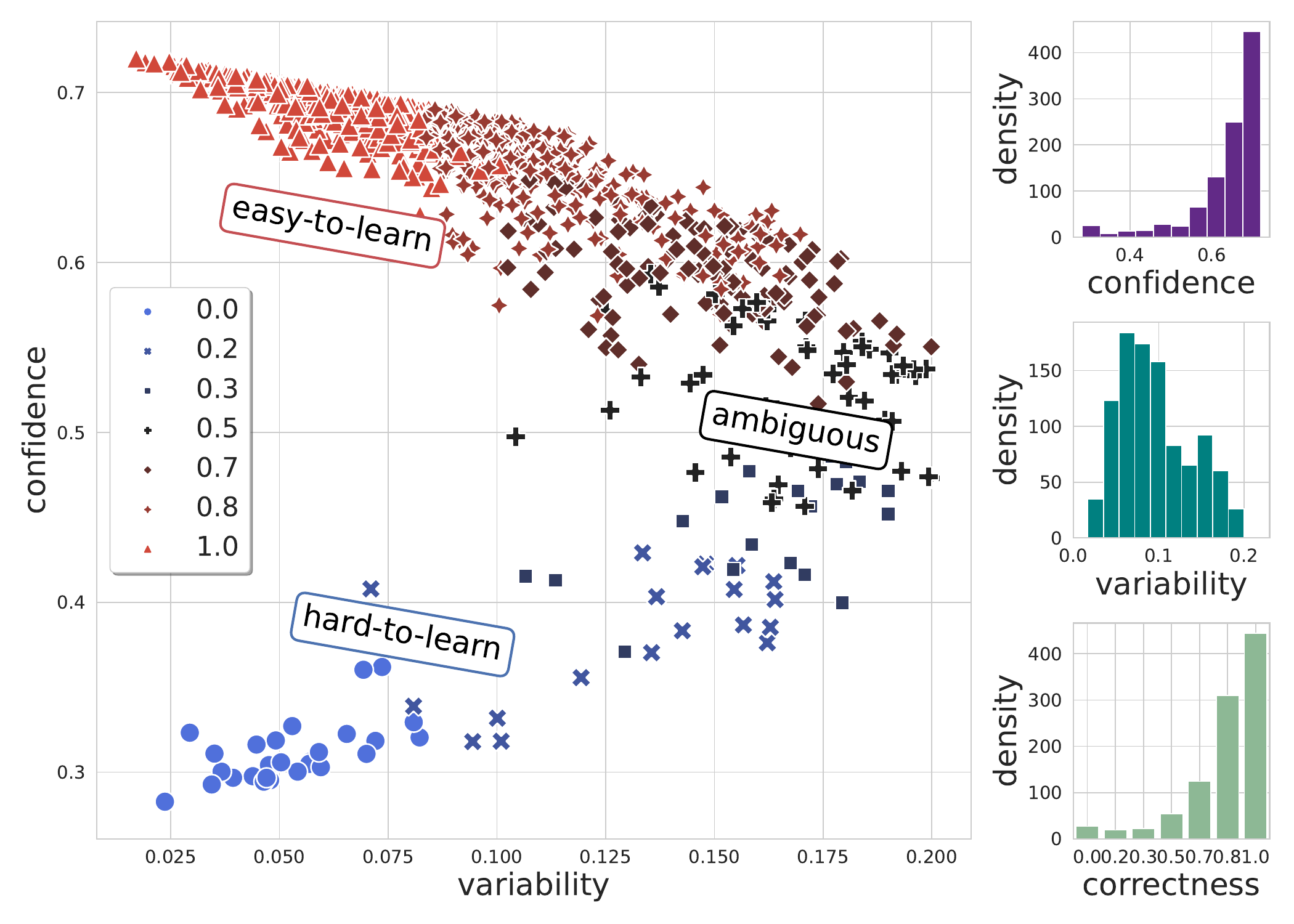}
         \vspace{-2em}
         \label{subfig:dataset_cartography_fusegen_flant5}
         \end{minipage}
     }
     \caption{Synthetic dataset cartography~\cite{swayamdipta2020dataset} using $1,000$ generated samples for movie review semantic analysis. ZeroGen uses zero-shot prompt for generation, while ProGen and FuseGen (Ours) use few-shot prompt with feedback but with different $K$, the number of PLMs involved. Numbers within parentheses are STM performance evaluated using IMDb after training on the generated dataset, with SWA applied during training.}
    \label{fig:appendix_dataset_cartography}
    \vspace{-1em}
\end{figure*}

We provide dataset cartography of synthetic datasets generated by $6$ different PLMs (GPT-2, Llama-2, Vicuna, OPT, ChatGLM3 and Flan-T5) in \cref{fig:appendix_dataset_cartography} . In left-subplot of each sub-figure in \cref{fig:appendix_dataset_cartography}, we display the variability (x-axis) and confidence (y-axis) of all samples. The right sub-plots depict histograms detailing the distributions of confidence, variability, and correctness. Notice that exactly $1,000$ samples are scattered onto each plot, although samples may overlap with each other, creating a visually sparser impression. 

Comparing dataset cartography generated by the same PLM, we can see that 
FuseGen helps to improve the dataset composition by introducing more ambiguous samples to balance the prevalence of the easy-to-learn samples, while ensuring hard-to-learn samples remain a minority. 


\subsection{T-SNE Visualization of Sample Distributions} \label{subsec:appendix_synthetic_dataset_distribution}

\begin{figure*}[!htb]
    \centering
    \subfigure[ZeroGen]{
         \begin{minipage}[t]{0.99\linewidth}
         \centering
         \includegraphics[width=1\linewidth]{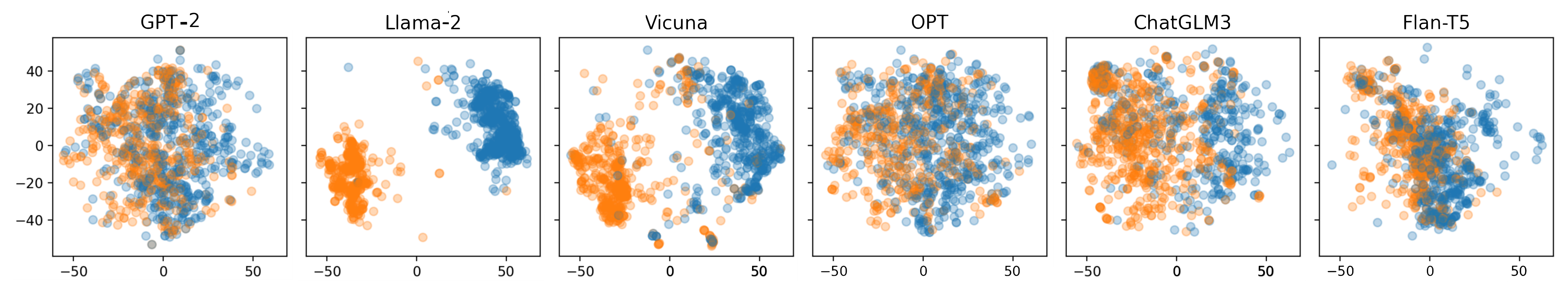}
         \label{subfig:dataset_distribution_original}
         \vspace{-1em}
         \end{minipage}
     }
     \subfigure[ProGen]{
         \begin{minipage}[t]{0.99\linewidth}
         \centering
         \includegraphics[width=1\linewidth]{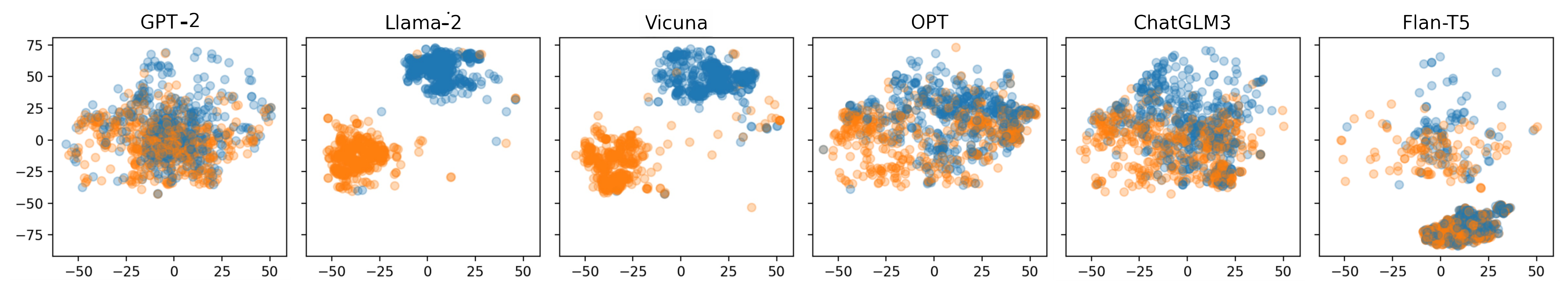}
         \label{subfig:dataset_distribution_single_progen}
         \vspace{-1em}
         \end{minipage}
     }
     \subfigure[FuseGen (Ours)]{
         \begin{minipage}[t]{0.99\linewidth}
         \centering
         \includegraphics[width=1\linewidth]{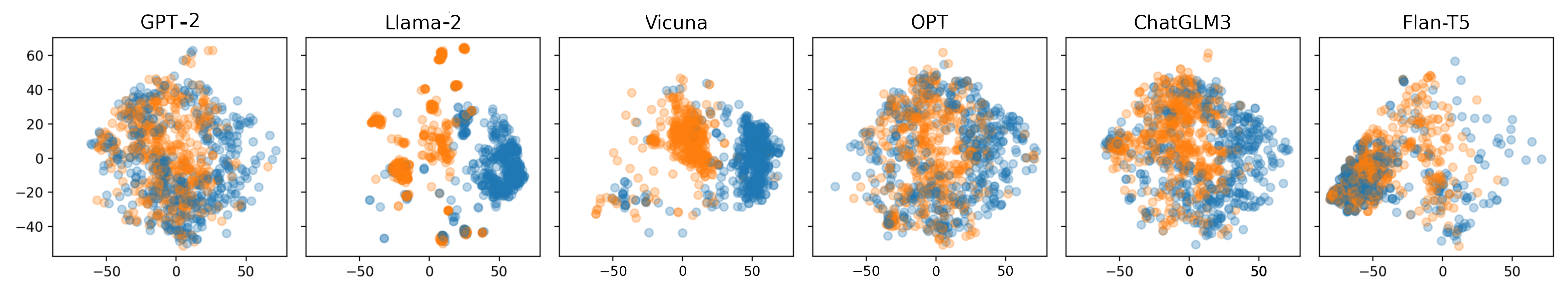}
         \label{subfig:dataset_distribution_fusegen}
         \vspace{-1em}
         \end{minipage}
     }
     \caption{t-SNE visualization of each synthetic sample generated by $6$ PLMs for movie review task. Different colors, blue and orange, represents embeddings from different class, positive and negative respectively.}
    \label{fig:imdb_total_tsne}
\end{figure*}

We also visualize the t-distributed Stochastic Neighbor Embedding (t-SNE) of synthetic samples ($N=1,000$) in \cref{fig:imdb_total_tsne}. All samples are embedded with a pre-trained bert-base-uncased encoder model. 

Consistent with the dataset cartography in \cref{fig:dataset_cartography_llama_flant5,fig:appendix_dataset_cartography}, FuseGen generates a higher proportion of ambiguous samples, which pulls the distribution of samples from different semantic classes closer to each other compared to ZeroGen and ProGen. This effect is particularly pronounced for synthetic datasets given by Llama-2 and Vicuna. 

\subsection{Low-quality Synthetic Dataset Samples} \label{subsec:appendix_dataset_noise}

In \cref{tab:low_quality_samples}, we show examples of low-quality samples, including samples that are ``mislabeled'', of ``low-relevancy'', and  of ``low-text-quality''. 
Samples are selected from synthetic datasets generated by individual PLMs using zero-shot prompt for the movie review semantic analysis task. This demonstrates the importance for improving the overall data quality of synthesic datasets. 
\begin{table*}[!htb]
\centering
    \resizebox{0.99\linewidth}{!}{
    \begin{tabular}{c|p{12.5cm}|c|c}
    \toprule
        PLM & Sample Text & Label & Type \\
    \midrule
        GPT-2 & A touching portrayal of friendship in the troubled 21st Century. & negative & mislabeled \\
        \hline
        \\[-1em]
        GPT-2 & We believe the movie <C> is decent but it's none too unexpected for someone like me who has fallen asleep watching movies and has seen this film before. & positive & mislabeled \\
        \hline
        \\[-1em]
        GPT-2 & The 'Godzilla' of the year thing is back. Say hello to The Amazing Spider-Man 2 & negative & low-relevancy \\
        \hline
        \\[-1em]
        GPT-2 & Broken Hearts Way was, on average, better than <T> was & positive & low-text-quality \\
        \hline
        \\[-1em]
        GPT-2 & Help! My name is Mark Hammer and I have always wanted to write a movie review. Now, after hours and hours of research, the truth is finally out. & negative & low-text-quality \\ 
        \hline
        \hline
        \\[-1em]
        Llama-2 & It was a complete waste of time and money, I would not recommend it to anyone. This movie was just plain bad, with uninteresting characters and a plot that was predictable from the start. The acting was also subpar and the dialogue was cringe-worthy. I would give it a 0/10 if I could. & positive & mislabeled \\
        \hline
        \\[-1em]
        Llama-2 & After waiting for what felt like an eternity, I finally watched & negative & low-text-quality \\
        \hline
        \hline
        \\[-1em]
        Vicuna & The film, directed by a renowned auteur, is a story of a group of people navigating through a dystopian society.**** It is a thought-provoking piece that leaves the audience pondering on the state of society and what it takes to survive in a world that is controlled by a powerful corporation. The cinematography is beautiful and the score is hauntingly good. & negative & mislabeled \\
        \hline
        \\[-1em]
        Vicuna & A great movie that will leave you thinking for days to come. No need to spend a lot of money to see it. It is a must watch for everyone. & negative & mislabeled \\
        \hline
        \\[-1em]
        Vicuna & C++ is a difficult and complex language, but with time and practice, it becomes easy and manageable.1234567890 & positive & low-relevancy \\
        \hline
        \hline
        \\[-1em]
        OPT & But the movie begins with this big, action-packed explosion, which carries right into the more thoughtful segments of the movie. But wait! There’s more! It continues through those thoughtful segments and the entire rest of the movie, not just a handful of sequences. & negative & mislabeled \\
        \hline
        \\[-1em]
        OPT & The movie is very disappointing. The story line is nothing but a shameless rip-off. Even the main cast of the movie is not worth any praise at all. This is a movie to just go and waste your money. Just don't! & positive & mislabeled \\
        \hline
        \\[-1em]
        OPT & When my teenage self in France was presented with this movie on VHS, I was in love with the soundtrack, with the French style of acting (something totally alien to my home in the States), and with the idea that there was a world like this. & negative & mislabeled \\
        \hline
        \\[-1em]
        OPT & The packaging is nothing more than an anonymous brown paper bag, and the theater provided stale popcorn. & positive & low-relevancy \\
        \hline
        \\[-1em]
        OPT & $\textbackslash$n$\textbackslash$n- a negative movie review$\textbackslash$n$\textbackslash$nThe movie review in negative sentiment for movie  & positive & low-text-quality \\
        \hline
        \hline
        \\[-1em]
        ChatGLM3 & Very disappointing. There was not one LOL moment. No wonder the movie was not a box office hit.  & positive & mislabeled \\
        \hline
        \\[-1em]
        ChatGLM3 & Perhaps a crime movie and is interesting to watch . & negative & mislabeled \\
        \hline
        \\[-1em]
        ChatGLM3 & i'm not the most romantic person and i'm not a chick. & positive & low-relevancy \\
        \hline
        \\[-1em]
        ChatGLM3 & even a bad magician should be able to catch the rabbit & positive & low-relevancy \\
        \hline
        \hline
        \\[-1em]
        Flan-T5 & He works in audio-visual technique and the end product is often flawed. & positive & mislabeled \\
        \hline
        \\[-1em]
        Flan-T5 & When a thing is a fantasy, it just become real, whether it was imagined or just played out. When they put on a performance in this movie, it has to be one of the best, most inspired moments. & negative & mislabeled \\
        \hline
        \\[-1em]
        Flan-T5 & if the time has come to say goodbye to Dick Van Patten. & positive & low-relevancy \\
        \hline
        \\[-1em]
        Flan-T5 & perverse creatures know they should be ashamed to exist. for human beings to walk around dressed like cannibals in a heavy jungle set up camp. & negative & low-relevancy \\
        \hline
        \\[-1em]
        Flan-T5 & And this is just another (incomplete) list of things that & negative & low-text-quality \\
    \bottomrule
    \end{tabular}
    }
\caption{Examples of low-quality samples in generated synthetic dataset for movie review.}
\label{tab:low_quality_samples}
\end{table*}

\subsection{Source of Selected In-context Samples} \label{subsec:appendix_sample_origination}

\begin{figure}[!htb]
    \vspace{-1em}
    \centering
     \subfigure[Samples in selected desirable subset of size $S=8$]{
         \begin{minipage}[t]{0.998\linewidth}
         \centering
         \includegraphics[width=1\linewidth]{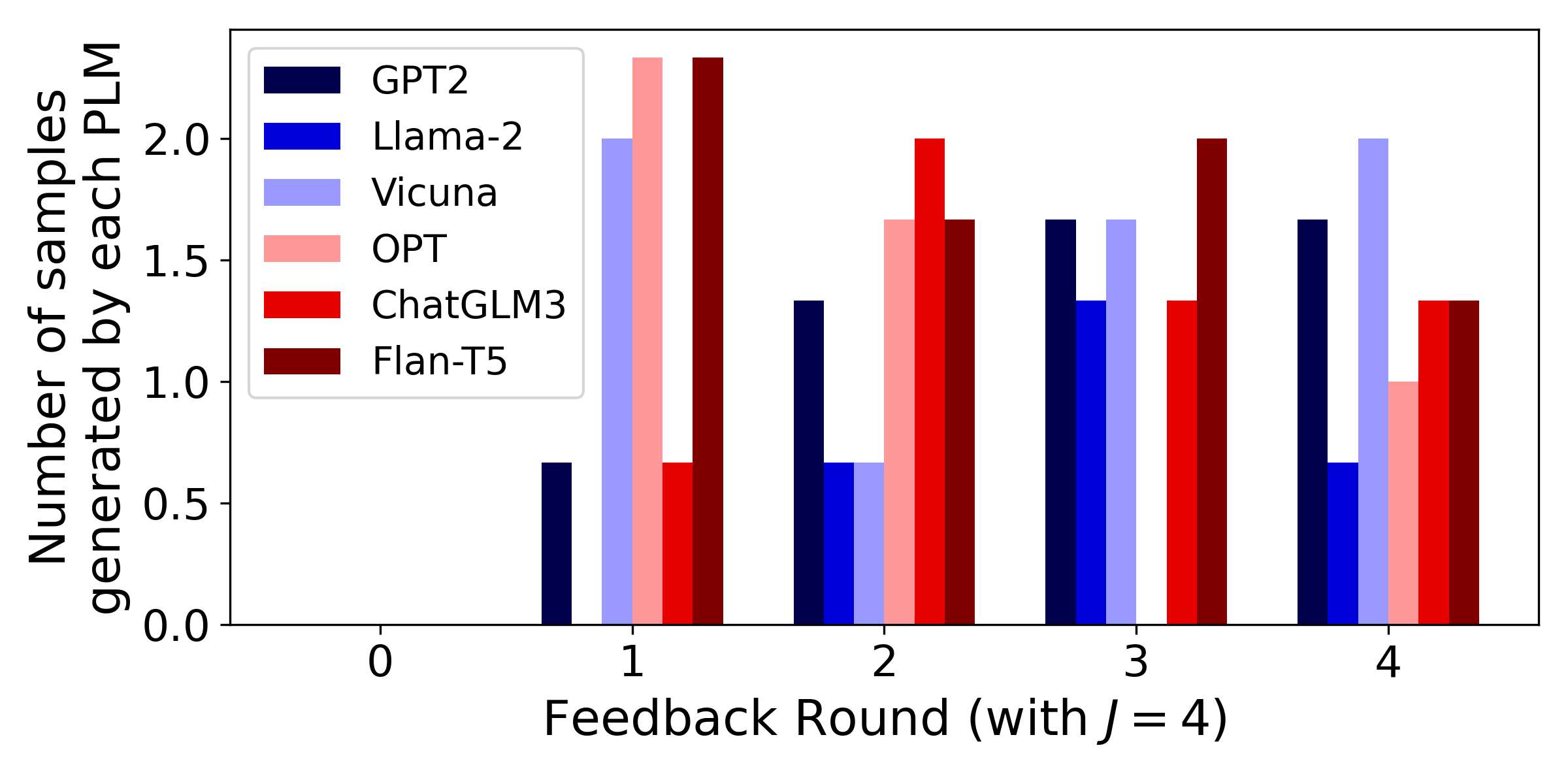}
         \label{subfig:sample_proportion_J_prompt}
         \end{minipage}
     }
     \subfigure[Selected candidates of size $R=40$]{
         \begin{minipage}[t]{0.998\linewidth}
         \centering
         \includegraphics[width=1\linewidth]{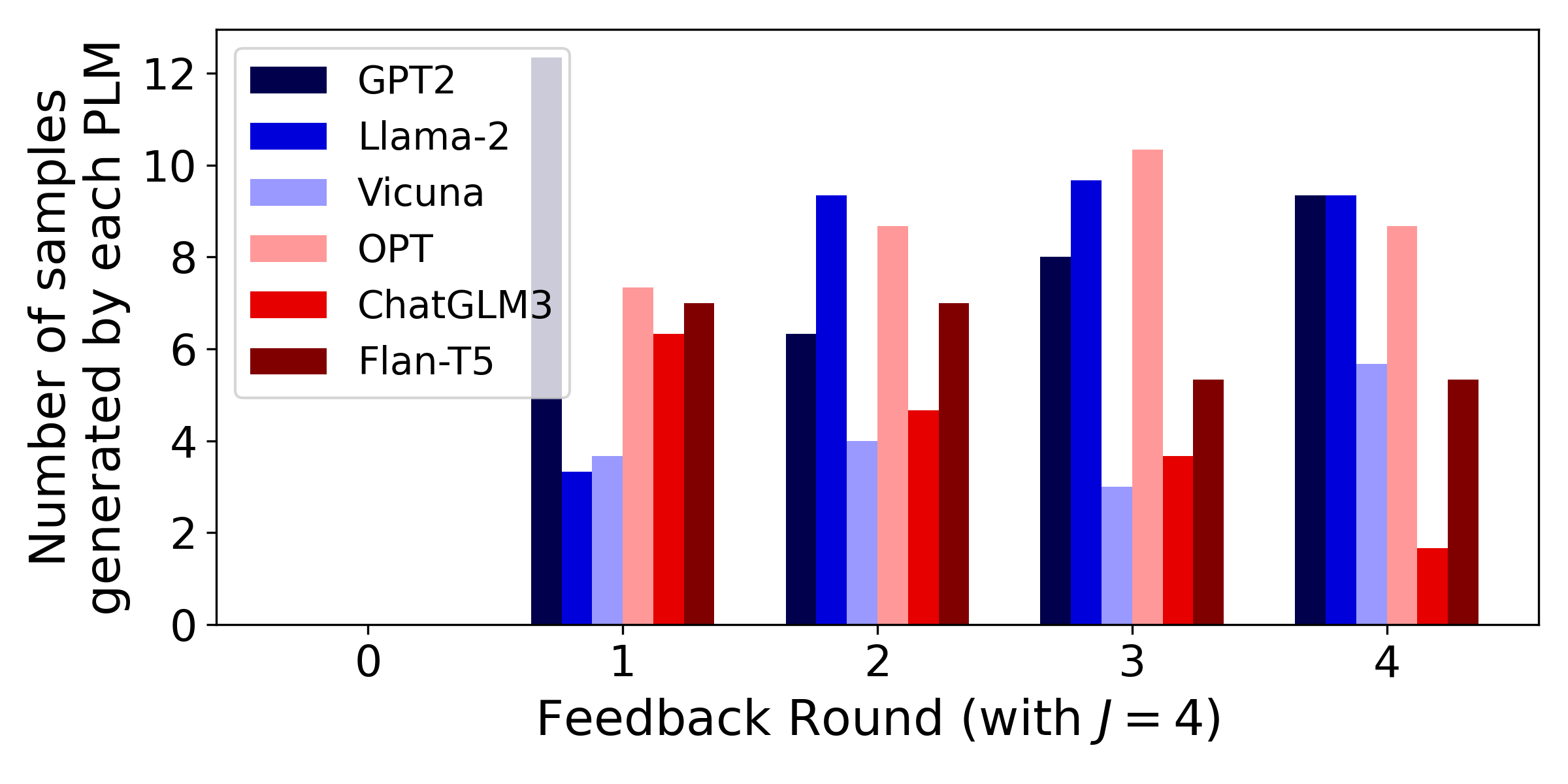}
         \label{subfig:sample_proportion_J_candidate}
         \end{minipage}
     }
     \caption{Proportion of samples in $S$ in-context samples and $R$ sample candidates that originate from each PLM at each feedback time ($J$) in FuseGen with $J=4,R=40,S=8,N=1,000,K=6$ for movie review sentiment analysis task. Results are averaged using $3$ different seeds. 
     }
    \label{fig:sample_proportion_J}
\end{figure}

We show in \cref{fig:sample_proportion_J} that, the selected in-context samples (desirable subset) and its candidates during CDG originate from various PLMs. However, the proportion of samples contributed by each PLM can fluctuate across iterations. This verifies that knowledge from different PLMs are fused and fed to each PLM through the feedback prompt, which further boosts the generation quality of each PLM.

\subsection{Ablations on More Tasks} \label{subsec:appendix_ablation_more_datasets}

\begin{table}[!tb]
\small
    \centering
    \resizebox{1\linewidth}{!}{
    \begin{tabular}{l||cccccc|c}
    \toprule
        ~ & \multicolumn{7}{c}{IMDb}\\
        \cmidrule(){2-8}
        ~ & $m_{G}$ & $m_{L}$ & $m_{V}$ & $m_{O}$ & $m_{C}$ & $m_{F}$ & $\tilde{m}$ \\
    \midrule
        FuseGen (Ours) & \textbf{87.85} & \textbf{86.60} & \textbf{87.50} & \textbf{88.47} & \textbf{88.56} & \textbf{88.73} & \textbf{90.19} \\
        \hline
        \\[-1em]
        w/o SWA & 82.90 & 78.98 & 74.34 & 85.17 & 85.77 & 85.43 & 89.07 \\
        \hline
        \\[-1em]
        w/o CDG \& SWA & 80.71 & 75.73 & 59.41 & 81.37 & 81.14 & 84.35 & 87.06 \\
        \hline
        \\[-1em]
        SDG+mixed & 80.72 & 76.18 & 65.05 & 84.19 & 84.56 & 81.19 & 87.41 \\
        \hline
        \hline
        \\[-1em]
         ~ & \multicolumn{7}{c}{SST-2} \\
        \cmidrule(){2-8}
        ~ & $m_{G}$ & $m_{L}$ & $m_{V}$ & $m_{O}$ & $m_{C}$ & $m_{F}$ & $\tilde{m}$ \\
    \midrule
        FuseGen (Ours) & \textbf{86.38} & \textbf{84.36} & \textbf{85.52} & \textbf{86.50} & \textbf{86.96} & \textbf{86.32} & \textbf{87.35} \\
        \hline
        \\[-1em]
        w/o SWA & 81.87 & 79.22 & 82.43 & 80.99 & 85.73 & 80.99 & 85.38 \\
        \hline
        \\[-1em]
        w/o CDG \& SWA & 80.68 & 76.42 & 76.46 & 80.80 & 84.58 & 78.44 & 85.01 \\
        \hline
        \\[-1em]
        SDG+mixed & 80.75 & 77.53 & 79.52 & 80.86 & 85.69 & 80.89 & 85.71 \\
        \hline
        \hline
        \\[-1em]
       ~ & \multicolumn{7}{c}{Yelp} \\
        \cmidrule(){2-8}
        ~ & $m_{G}$ & $m_{L}$ & $m_{V}$ & $m_{O}$ & $m_{C}$ & $m_{F}$ & $\tilde{m}$\\
    \midrule
        FuseGen (Ours) & \textbf{91.94} & \textbf{90.30} & \textbf{90.81} & \textbf{92.50} & \textbf{92.98} & \textbf{92.21} & \textbf{93.54} \\
        \hline
        \\[-1em]
        w/o SWA & 90.87 & 88.09 & 84.99 & 87.19 & 91.72 & 90.71 & 92.84 \\
        \hline
        \\[-1em]
        w/o CDG \& SWA & 89.13 & 79.17 & 81.97 & 86.78 & 81.50 & 89.48 & 92.16 \\
        \hline
        \\[-1em]
        SDG+mixed &  89.63 & 82.39 & 83.80 & 86.84 & 86.32 & 87.48 & 92.23 \\
        \hline
        \hline
        \\[-1em]
        ~ & \multicolumn{7}{c}{QNLI} \\
        \cmidrule(){2-8}
        ~ & $m_{G}$ & $m_{L}$ & $m_{V}$ & $m_{O}$ & $m_{C}$ & $m_{F}$ & $\tilde{m}$ \\
    \midrule
        FuseGen (Ours) & \textbf{60.55} & \textbf{72.48} & \textbf{74.10} & \textbf{57.39} & \textbf{69.89} & \textbf{72.13} & \textbf{74.95} \\
        \hline
        \\[-1em]
        w/o SWA & 56.72 & 69.99 & 70.94 & 51.98 & 56.39 & 68.65 & 73.41 \\
        \hline
        \\[-1em]
        w/o CDG \& SWA & 51.24 & 65.81 & 70.61 & 50.83 & 53.01 & 55.73 & 69.41 \\
        \hline
        \\[-1em]
        SDG+mixed & 52.13 & 69.22 & 70.11 & 51.79 & 54.87 & 68.58 & 70.20 \\
    \bottomrule
    \end{tabular}
    }
\caption{Comparison between FuseGen and its ablations with $K=6, N=1,000, J=4$. Each $m_{k}$ is trained on $\mathcal{D}_{k}$ of size $1,000$ while $\tilde{m}$ is trained on $\mathcal{D}$ of size $6,000$.} 
\label{tab:ablation_results_appendix}
\vspace{-1em}
\end{table}

We include the ablation results of ``w/o SWA'', ``w/o CDG \& SWA'' and ``SDG+mixed''(also w/o SWA) for more tasks and here due to space limitation. 
We also elaborate the explanation of ``SDG+mixed'' here. In ``SDG+mixed'', SWA is removed and CDG is replaced with self-based feedback, i.e. random selection is applied to select $R$ candidate samples from each $\mathcal{D}_{k}$. $K$ in-context samples subsets are than selected based on sample importance from the $K$ candidate sample sets of size $R$ and are further fed to respective PLM $\mathcal{P}_{k}$ to generate samples.

As illustrated in Table \ref{tab:ablation_results_appendix}, the application of SWA significantly improves the performance of all STMs, particularly for ${\{m_{k}\}}_{k=1}^K$. This improvement highlights the efficacy of SWA in enhancing the quality of synthetic datasets through the up-weighting of higher-quality samples and the down-weighting of lower-quality samples, thereby reducing the impact of the latter.
Furthermore, the application of CDG also significantly boosts the performance of all STMs to a greater extent than applying SDG. This underscores the superiority of cross-model feedback over the combination of self-guided feedback and highlights the efficacy of CDG in harnessing the capabilities of multiple PLMs.

\subsection{Multi-PLM v.s. single-PLM on More Tasks} \label{subsec:appendix_k_is_1_more_datasets}

\begin{table}[!tb]
    \centering
    \resizebox{0.999\linewidth}{!}{
    \begin{tabular}{l||c|cccccc}
    \toprule
        ~ & multi & \multicolumn{6}{c}{single} \\
        \cmidrule{2-8}
        ~ & $\tilde{m}$ & $\tilde{m}_{G}$ & $\tilde{m}_{L}$ & $\tilde{m}_{V}$ & $\tilde{m}_{O}$ & $\tilde{m}_{C}$ & $\tilde{m}_{F}$ \\
    \midrule
        IMDb & \textbf{89.96} & 87.60 & 86.14 & 85.42 & 87.59 & 88.84 & \underline{89.74} \\
        SST-2 & \textbf{87.51} & 84.81 & 84.39 & 85.22 & 85.88 & \underline{87.43} & 85.38 \\
        Yelp & \textbf{93.27} & \underline{93.03} & 91.07 & 91.69 & 92.72 & 92.08 & 92.07 \\
        QNLI & \textbf{74.92} & 64.52 & 73.22 & 73.34 & 59.03 & 64.93 & \underline{73.60} \\
        MNLI-m & \textbf{49.76} & 44.93 & \underline{49.61} & 49.11 & 37.40 & 32.82 & 49.34 \\
        MNLI-mm & \textbf{51.70} & 48.53 & \underline{51.62} & 50.76 & 42.32 & 33.05 & 51.47 \\
        AgNews & \textbf{86.89} & 82.21 & 85.34 & 85.36 & \underline{86.75} & 86.27 & 86.36 \\
        MarkedNews & \textbf{83.85} & 79.98 & 80.04 & 79.36 & 78.60 & \underline{83.54} & 80.86 \\
    \bottomrule
    \end{tabular}
    }
\caption{Comparison between FuseGen using multi-PLM ($K=6$) and single-PLM ($K=1$) with $4$ datasets. MNLI-m and MNLI-mm each stands for MNLI-matched and MNLI-mismatched. Best result is marked as \textbf{bold} with the second best marked with \underline{underline} for each dataset (each row).} 
\label{tab:singlePLM_and_multiPLM}
\end{table}

We provided additional results on the comparison of multi-PLM ($K=6$) and single-PLM ($K=1$) across $8$ datasets for various tasks in \cref{tab:singlePLM_and_multiPLM}. As multi-PLM ($K=6$) consistently outperforms all single-PLM under the each task, we conclude that multi-PLM collaboration is more effective than relying on a single PLM for enhancing STM performance.

\subsection{Detailed Results for Hyper-parameters $\alpha$, $N$ and $J$} \label{subsec:appendix_ablation_results}

Due to space limitation, we provide detailed results of hyper-parameters $\alpha$ (ratio of high-variability samples within the $R$ in-context sample candidates), $N$ (sample generation budget), and $J$ (feedback times) 
here in \cref{tab:varying_alpha,tab:varying_N,tab:varying_J}. We additionally include the performance of each $m_{k}$ 
as well (SWA applied). 
These results indicate that employing a more balanced mix of high-variability and low-variability samples ($\alpha=0.5$), a larger sample budget $N$ and more feedback times $J$ all help to achieve a better STM performance. This enhancement is observed not only for the final STM $\tilde{m}$, but also for each ${\{m_k\}}_{k=1}^{K}$.

\begin{table}[!tb]
    \centering
    \resizebox{1\linewidth}{!}{
    \begin{tabular}{c||cccccc|c}
    \toprule
        $\alpha$ & $m_{G}$ & $m_{L}$ & $m_{V}$ & $m_{O}$ & $m_{C}$ & $m_{F}$ & $\tilde{m}$ \\
    \midrule
        $0.0\hphantom{0}$ & 54.00 & 70.07 & 67.75 & 51.12 & 55.70 & 66.49 & 74.08 \\
        $0.25$ & \underline{56.12} & \underline{70.22} & \underline{70.45} & 52.10 & \underline{56.90} & \underline{71.12} & \underline{74.37} \\
        $0.5\hphantom{0}$ & \textbf{59.68} & \textbf{71.48} & \textbf{72.37} & \textbf{52.37} & \textbf{57.33} & \textbf{72.12} & \textbf{74.92} \\
        $0.75$ & 55.27 & 69.13 & 69.53 & \underline{52.19} & 56.59 & 70.91 & 74.23 \\
        $1.0\hphantom{0}$ & 54.85 & 66.47 & 64.46 & 50.08 & 56.50 & 70.50 & 74.16 \\
    \bottomrule
    \end{tabular}
    }
    \caption{Comparison of different $\alpha$ used for FuseGen with QNLI as test dataset. Best result is marked as \textbf{bold} with the second best marked with \underline{underline} for each STM (each column).} 
    \label{tab:varying_alpha}
\end{table}

\begin{table}[!tb]
    \centering
    \resizebox{1\linewidth}{!}{
    \begin{tabular}{c||cccccc|c}
    \toprule
        $N$ & $m_{G}$ & $m_{L}$ & $m_{V}$ & $m_{O}$ & $m_{C}$ & $m_{F}$ & $\tilde{m}$ \\
    \midrule
        \hphantom{00}$100$ & 51.33 & 53.16 & 53.79 & 50.62 & 51.20 & 51.11 & 56.27 \\
        \hphantom{00}$200$ & 52.23 & 60.42 & 60.06 & 50.71 & 53.07 & 59.09 & 65.11 \\
        \hphantom{00}$500$ & \underline{53.53} & \underline{67.36} & \underline{67.90} & \underline{51.67} & \underline{54.95} & \underline{64.72} & \underline{72.18} \\
        $1,000$ & \textbf{59.68} & \textbf{71.48} & \textbf{72.37} & \textbf{52.37} & \textbf{57.33} & \textbf{72.12} & \textbf{74.92} \\
    \bottomrule
    \end{tabular}
    }
    \caption{Comparison of different $N$ used for FuseGen with QNLI as test dataset. Best result is marked as \textbf{bold} with the second best marked with \underline{underline} for each STM (each column).} 
    \label{tab:varying_N}
\end{table}

\begin{table}[!tb]
    \centering
    \resizebox{1\linewidth}{!}{
    \begin{tabular}{c||cccccc|c}
    \toprule
        $J$ & $m_{G}$ & $m_{L}$ & $m_{V}$ & $m_{O}$ & $m_{C}$ & $m_{F}$ & $\tilde{m}$ \\
    \midrule
        $0$ & 56.95 & 71.13 & 72.21 & 51.96 & 55.12 & 58.43 & 74.44 \\
        $1$ & 57.11 & \underline{71.50} & 72.25 & 52.07 & 56.53 & 64.81 & 74.77 \\
        $4$ & \underline{59.68} & 71.48 & \textbf{72.37} & \textbf{52.37} & \underline{57.33} & \underline{72.12} & \underline{74.92} \\
        $9$ & \textbf{59.71} & \textbf{71.60} & \textbf{72.37} & \underline{52.34} & \textbf{57.70} & \textbf{72.14} & \textbf{75.07} \\
    \bottomrule
    \end{tabular}
    }
    \caption{Comparison of different $J$ used for FuseGen with QNLI as test dataset. Best result is marked as \textbf{bold} with the second best marked with \underline{underline} for each STM (each column).} 
    \label{tab:varying_J}
\end{table}

\end{document}